%% file: wacv2023.tex
\documentclass[10pt,twocolumn,letterpaper]{article}

\usepackage{wacv}
\usepackage{times}
\usepackage{epsfig}
\usepackage{graphicx}
\usepackage{amsmath}
\usepackage{amssymb}
\usepackage{booktabs}
\usepackage{comment}
\usepackage{color}
\usepackage{lipsum}  
\usepackage{nicefrac}
\usepackage{textcomp}
\usepackage{multirow}
\usepackage{bm}
\usepackage{subcaption}

\DeclareMathOperator*{\argmax}{arg\,max}
\DeclareMathOperator*{\argmin}{arg\,min}

\def\wacvPaperID{55} % Enter the WACV Paper ID here

\wacvfinalcopy % *** Uncomment this line for the final submission

\ifwacvfinal
\usepackage[pagebackref=true,breaklinks=true,colorlinks,bookmarks=false]{hyperref}
\else
\usepackage[pagebackref=true,breaklinks=true,colorlinks,bookmarks=false]{hyperref}
\fi

\begin{document}

%%%%%%%%% TITLE
\title{Anomaly Clustering: Grouping Images into Coherent Clusters of Anomaly Types}

\author{Kihyuk Sohn, 
Jinsung Yoon,
Chun-Liang Li,
Chen-Yu Lee,
Tomas Pfister\\
Google Cloud AI Research\\
{\tt\small \{kihyuks,jinsungyoon,chunliang,chenyulee,tpfister\}@google.com}
}

\maketitle
\thispagestyle{empty}

\begin{abstract}
We study anomaly clustering, grouping data into coherent clusters of anomaly types. This is different from anomaly detection that aims to divide anomalies from normal data.
Unlike object-centered image clustering, anomaly clustering is particularly challenging as anomalous patterns are subtle and local.
We present a simple yet effective clustering framework using a patch-based pretrained deep embeddings and off-the-shelf clustering methods. We define a distance function between images, each of which is represented as a bag of embeddings, by the Euclidean distance between weighted averaged embeddings. The weight defines the importance of instances (i.e., patch embeddings) in the bag, which may highlight defective regions. We compute weights in an unsupervised way or in a semi-supervised way when labeled normal data is available.
Extensive experimental studies show the effectiveness of the proposed clustering framework along with a novel distance function upon existing multiple instance or deep clustering frameworks.
Overall, our framework achieves 0.451 and 0.674 normalized mutual information scores on MVTec object and texture categories and further improve with a few labeled normal data (0.577, 0.669), far exceeding the baselines (0.244, 0.273) or state-of-the-art deep clustering methods (0.176, 0.277).
\end{abstract}

\vspace{-0.3in}
\section{Introduction}
\label{sec:intro}
%\vspace{-0.05in}
Anomaly detection aims to detect anomalous data when majority of the data is normal. To deal with the scarcity of the labeled anomalous data at train time, anomaly detection problems are often formulated as a one-class classification problem~\cite{scholkopf1999support,tax2004support,ruff2018deep}, where one builds a classifier that could separate anomalous data from normal ones at test time using only normal data at train time. 
As a result of anomaly detection, one would get a binary label of normalcy or anomaly.

% Progress on visual anomaly detection
% deep learning --> learning deep one-class classifier has demonstrated impressive performance on semantic anomaly detection~\cite{}, e.g., separating a dog from cats, as well as 

\begin{figure}[t]
    \centering
    \begin{subfigure}{0.48\linewidth}
    \centering
    \includegraphics[width=\linewidth]{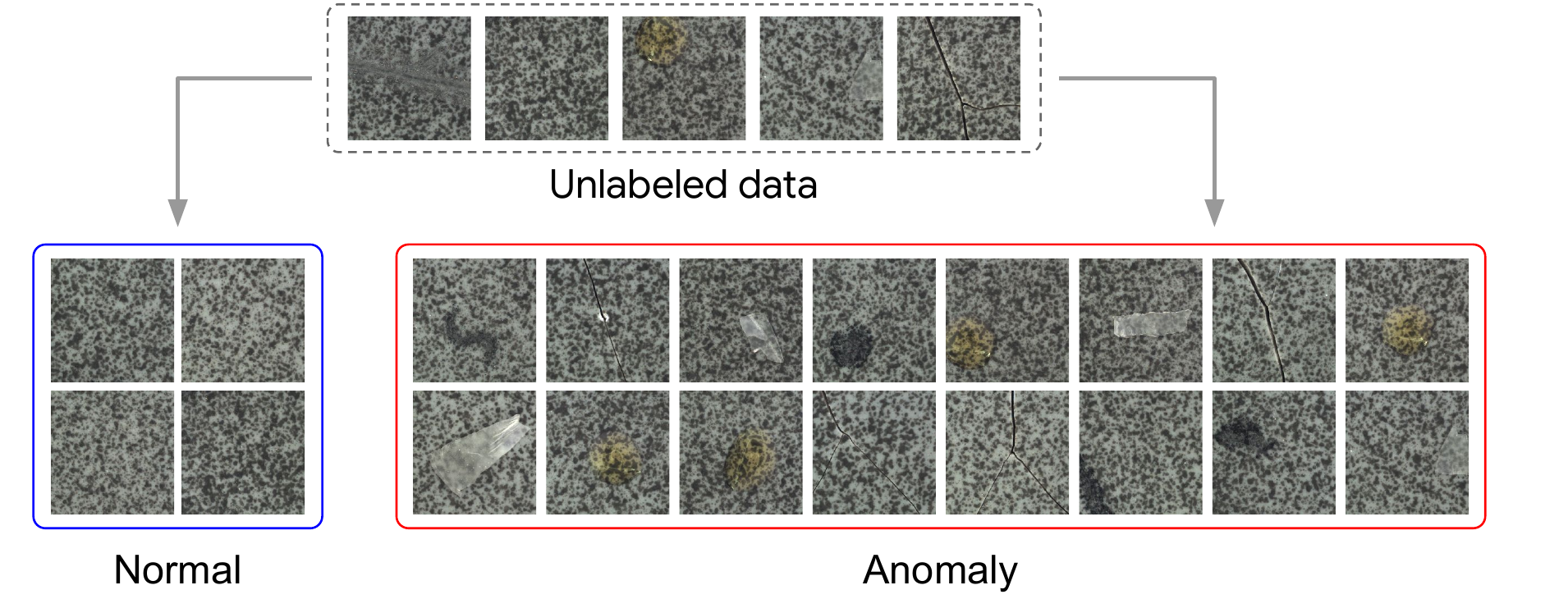}
    \caption{Anomaly Detection}
    \label{fig:teasar_anomaly_detection}
    \end{subfigure}
    \begin{subfigure}{0.48\linewidth}
    \centering
    \includegraphics[width=\linewidth]{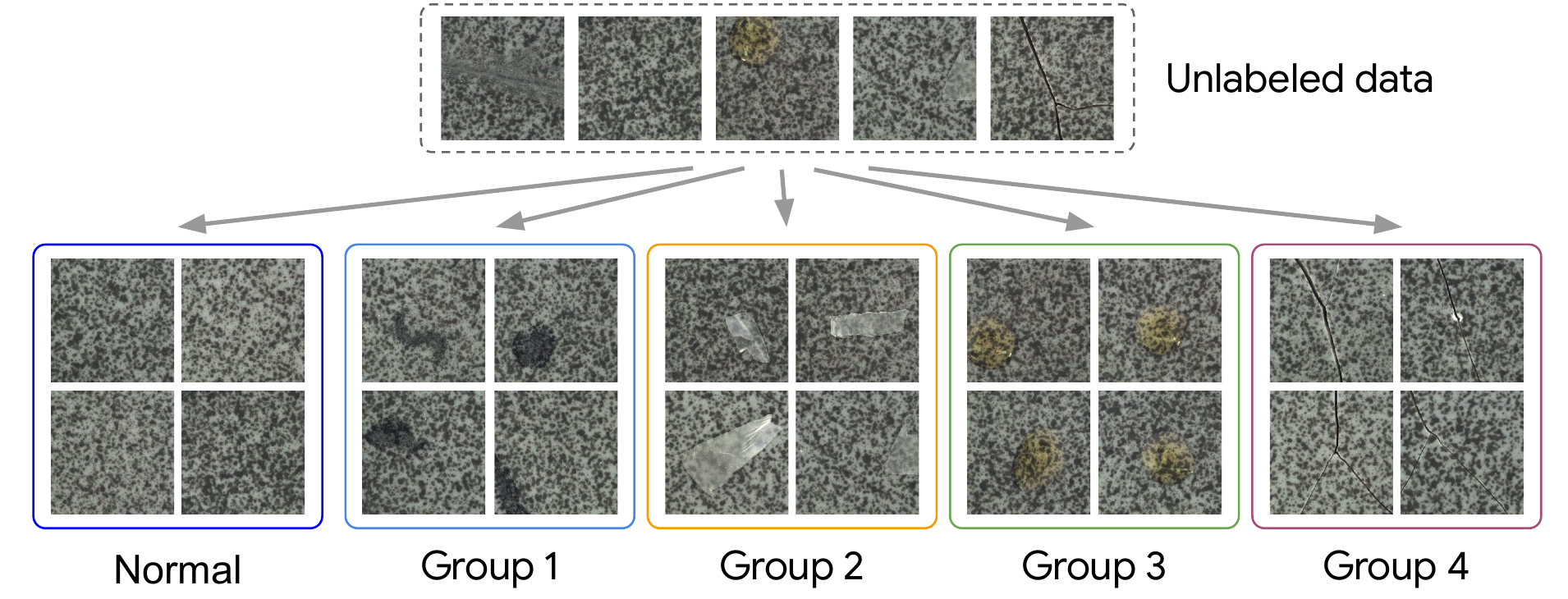}
    \caption{Anomaly Clustering (ours)}
    \label{fig:teasar_anomaly_clustering}
    \end{subfigure}
    \vspace{-0.1in}
    \caption{Different from existing works on (\ref{fig:teasar_anomaly_detection}) anomaly detection~\cite{bergmann2019mvtec,bergmann2020uninformed,li2021cutpaste,roth2021towards}, (\ref{fig:teasar_anomaly_clustering}) we study \textit{anomaly clustering} that groups unlabeled data automatically into multiple clusters, each of which may represent different types of anomaly. %By grouping unlabeled images, we could discover different anomaly types for root-cause analysis and building multi-class anomaly classifier via active learning.
    }
    \label{fig:teasar}
    \vspace{-0.2in}
\end{figure}

%
However, a binary label has limited expression as there could be many sources of anomalous behaviors as in Figure~\ref{fig:teasar_anomaly_detection}. 
On the other hand, grouping data into multiple, semantically coherent clusters, as in Figure~\ref{fig:teasar_anomaly_clustering}, would be valuable for some reasons.
For example, cluster assignments could be used to generate the query data for active learning, where diversity is important~\cite{nguyen2004active,mac2014hierarchical,hasan2015context,sener2018active}, to improve the performance of anomaly detector. 
Moreover, it would help data scientists analyze the root causes of various anomaly types, hoping to fix their manufacturing pipeline to reduce anomalous behaviors.
This paper deals with the problem of clustering images with anomalous patterns.

Clustering could be used for our problem.
%Clustering is an essential technique for machine learning and computer vision that has been extensively studied. 
%Clustering has been extensively studied over decades.
%
Classic methods like KMeans~\cite{macqueen1967some}, spectral clustering~\cite{ng2002spectral}, or hierarchical agglomerative clustering~\cite{ward1963hierarchical}, focus on grouping data given representations, while recent deep clustering methods~\cite{xie2016unsupervised,caron2018deep,ji2019invariant,van2020scan,niu2020gatcluster} aim to learn high-level representations and their grouping jointly.
They have shown impressive clustering accuracy on several vision datasets such as CIFAR-10~\cite{krizhevsky2009learning} or ImageNet~\cite{deng2009imagenet}.

% Challenge of anomaly clustering
In this paper, we introduce anomaly clustering, a problem of grouping images into different anomalous patterns. 
While there has been a substantial progress in image clustering research~\cite{wazarkar2018survey,van2020scan}, anomaly clustering poses unique challenges.
%
%Anomaly clustering, a problem of grouping images with different types of anomalous patterns, however, poses unique challenges. 
Firstly, unlike typical image clustering datasets, images for anomaly clustering may not be object-centered. 
Rather, images are mostly similar to each other but differs at local regions. 
To our knowledge, grouping images by capturing fine-grained details, as opposed to the coarse-grained object semantics, has not been studied in existing works.
Secondly, it is common that the data is limited in industrial applications, making state-of-the-art deep clustering methods, which are usually trained on large datasets, less applicable.
We highlight challenges of anomaly clustering via empirical comparisons to deep clustering in Section~\ref{sec:exp_deep_clustering}.

We present an anomaly clustering framework to tackle this important real-world problem. 
To resolve the limited-data issue, we employ the pretrained deep representation, similar to solutions for anomaly detection~\cite{rippel2021modeling,defard2021padim,roth2021towards}, followed by similarity-based clustering methods. 
To tackle the non-object-centric issue, we represent an image as a bag of patch embeddings, as in Figure~\ref{fig:framework}, instead of a holistic representation. 
This casts the problem naturally into a multiple instance clustering~\cite{zhang2009multi}. 
The question boils down to defining a distance function between bags of instances (i.e., patch embeddings), and we propose a weighted average distance, which aggregates patch embeddings with weights followed by the Euclidean distance. The weight indicates which instances to attend to, and could be derived in an unsupervised way or in a semi-supervised way using extra labeled normal images.
The proposed framework is described in Figure~\ref{fig:framework}.

We conduct comprehensive experiments on two anomaly detection datasets, MVTec anomaly detection~\cite{bergmann2019mvtec} and magnetic tile defect~\cite{huang2020surface}. We present a new experimental protocol for anomaly clustering, whose performance is evaluated using ground-truth defect type annotations. We test the proposed clustering framework using various distance functions, including variants of Hausdorff distances~\cite{huttenlocher1993comparing,dubuisson1994modified,zhang2009multi} and our weighted average distance. We also compare with state-of-the-art deep clustering methods~\cite{ji2019invariant,van2020scan,niu2020gatcluster} for anomaly clustering. 
While being conceptually and computationally simple, our results show that the proposed framework solves the anomaly clustering problem significantly better than existing deep clustering methods. Our weighted average distance also demonstrates the efficacy over existing distance functions for multiple instance clustering.

%We confirm that our proposed framework solves the anomaly clustering problem significantly better than existing deep clustering frameworks, while being simple both conceptually and computationally. The proposed weighted average distance demonstrates the efficacy over existing distance functions for multiple instance clustering.

%
Finally, we summarize our contributions as follows:
\begin{enumerate}
    \setlength{\itemsep}{0pt}
    \setlength{\parskip}{1pt}
    \item We introduce an anomaly clustering problem and cast it as a multiple instance clustering using patch-based deep embeddings for an image representation.
    \item We propose a weighted average distance that computes the distance by focusing on important instances in unsupervised or semi-supervised ways.
    \item We conduct experiments on industrial anomaly detection datasets, showing solid improvements over multiple instance and deep clustering baselines.
\end{enumerate}

\vspace{-0.15in}
\section{Related Work}
\label{sec:related}
%\vspace{-0.1in}

\noindent\textbf{Anomaly detection} has been extensively studied under various settings~\cite{chandola2009anomaly}, such as supervised with both labeled normal and anomalous data, semi-supervised with labeled normal data~\cite{scholkopf1999support,tax2004support}, or unsupervised with unlabeled data~\cite{breunig2000lof,liu2008isolation,yoon2021self}, to train classifiers. While anomaly detection divides data into two classes of normalcy and anomaly, our goal is to group them into many clusters, each of which represents various anomalous behaviors. %We study an anomaly clustering for unsupervised and semi-supervised settings. 

Thanks to deep learning there has been a solid progress in visual anomaly detection. Self-supervised representation learning methods~\cite{gidaris2018unsupervised,chen2020simple} have been adopted to build deep one-class classifiers~\cite{golan2018deep,hendrycks2019deep,bergman2020classification,tack2020csi,sohn2020learning}, showing improvement in anomaly detection~\cite{bergmann2020uninformed,yi2020patch,li2021cutpaste}. In addition, the deep image representations trained on large-scale object recognition datasets~\cite{donahue2014decaf} have shown to be a good feature for visual anomaly detection~\cite{bergmann2020uninformed,rippel2021modeling,reiss2021panda,defard2021padim,roth2021towards}.
While we follow the similar intuition as we represent an image as a bag of patch embeddings with pretrained networks, we propose a method for grouping images into multiple clusters instead of building one-class classifier for binary classification.

%to compute the distance between images for clustering instead of building one-class classifier for detection, i.e., binary classification.

\noindent\textbf{Image clustering} is an active research area, whose main concern is at image representation.
Typical approaches~\cite{leung2001representing,csurka2004visual} include bag-of-keypoints~\cite{csurka2004visual}, where one builds a histogram of local descriptors (e.g., SIFT~\cite{lowe2004distinctive} or Texton~\cite{julesz1981textons}), and spatial pooling~\cite{lazebnik2006beyond}, aggregating local descriptors by averaging, to obtain a holistic representation of an image. 
Some applications relevant to our work include texture and material classification~\cite{leung2001representing,varma2002classifying,lazebnik2005sparse} and description~\cite{ferrari2007learning,cimpoi2015deep}. While their goal is to classify images of different texture or material properties with supervision, our goal is to cluster images with subtle differences due to defects without or with a minimal supervision. Moreover, we use patch representations and cast the problem as multiple instance clustering by automatically identifying important instances.

On the other hand, deep clustering~\cite{xie2016unsupervised,caron2018deep,ji2019invariant,van2020scan,niu2020gatcluster} jointly learns image representations and group assignments using deep neural networks. 
%extract a holistic image representation (e.g., hand-crafted features such as SIFT~\cite{lowe2004distinctive}, HOG~\cite{dalal2005histograms}, or deep representations~\cite{donahue2014decaf}, followed by bag-of-keypoints~\cite{csurka2004visual} or spatial pyramid pooling~\cite{lazebnik2006beyond}) and apply clustering methods (e.g., KMeans~\cite{macqueen1967some}, spectral clustering~\cite{ng2002spectral}, hierarchical agglomerative clustering~\cite{ward1963hierarchical}). Deep clustering~\cite{xie2016unsupervised,caron2018deep,ji2019invariant,van2020scan,niu2020gatcluster}, on the other hand, jointly learns image representations and group assignments using deep neural networks. Many deep clustering methods adopt self-training, where one infers group assignments, and uses them to train deep network iteratively. Autoencoder~\cite{xie2016unsupervised} or self-supervised~\cite{van2020scan} pretraining have been adopted to provide a good initial group assignment. 
%
While there has been a huge progress in clustering natural and object-centered images, such as those from CIFAR-10~\cite{krizhevsky2009learning} or ImageNet~\cite{deng2009imagenet}, state-of-the-art deep clustering algorithms do not work well for anomaly clustering, which requires to capture subtle differences of various anomaly types, as in Section~\ref{sec:exp_deep_clustering}.
%

\iffalse
\begin{figure*}[t]
    \centering
    \begin{subfigure}{0.42\linewidth}
        \centering
        \includegraphics[width=0.93\linewidth]{fig/mi-cluster-framework.pdf}
        \caption{Anomaly clustering framework.}
        \label{fig:framework_framework}
    \end{subfigure}
    \hspace{0.1in}
    \begin{subfigure}{0.48\linewidth}
        \centering
        \includegraphics[width=0.93\linewidth]{fig/mi-cluster-feature.pdf}
        \caption{A method for aggregating embeddings to compute the distance between images.}
        \label{fig:framework_feature}
    \end{subfigure}
    \vspace{-0.05in}
    \caption{(\ref{fig:framework_framework}) The proposed anomaly clustering framework uses similarity-based clustering methods, such as spectral clustering or hierarchical clustering. (\ref{fig:framework_feature}) We compute the distance between images, each of which is represented as a bag of patch embeddings from a pretrained deep neural network, by the Euclidean distance of weighted averaged embeddings, whose weight represents the ``importance'' (e.g., defectiveness) of patch embeddings.}
    \label{fig:framework}
    \vspace{-0.1in}
\end{figure*}
\fi

\begin{figure*}[t]
    \centering
    \includegraphics[width=0.93\linewidth]{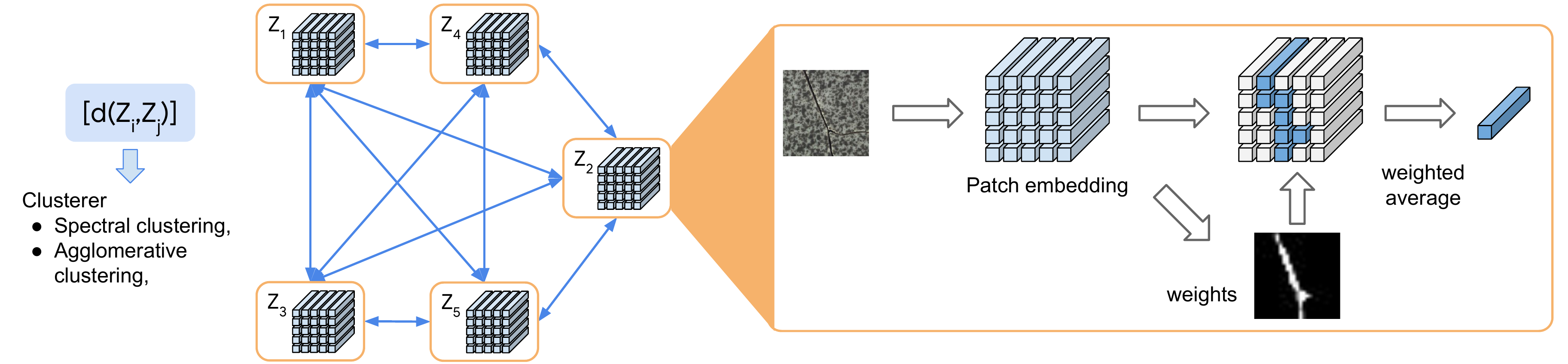}
    \caption{The proposed anomaly clustering framework uses similarity-based clustering methods, such as spectral clustering or hierarchical clustering. We compute the distance between images, each of which is represented as a bag of patch embeddings from a pretrained deep neural network, by the Euclidean distance of weighted averaged embeddings, whose weight represents the ``importance'' (e.g., defectiveness) of patch embeddings.}
    \label{fig:framework}
    \vspace{-0.1in}
\end{figure*}

%\noindent\textbf{Distance between sets}

%\vspace{-0.1in}
\section{Anomaly Clustering}
\label{sec:method}
%\vspace{-0.05in}

We introduce the proposed anomaly clustering, where in Section~\ref{sec:method_framework} we formulate it as a multiple instance clustering problem~\cite{zhang2009multi}. In Section~\ref{sec:method_unsup_ac}, we define a distance measure under unsupervised setting in Section~\ref{sec:method_unsup_alpha} and under semi-supervised setting with a few normal data in Section~\ref{sec:method_semisup_ac}.

%\vspace{-0.1in}
\subsection{Framework Overview}
\label{sec:method_framework}
%\vspace{-0.05in}

Let $\mathcal{X}\,{=}\,\{\mathbf{x}_{i}\,{\in}\,\mathbb{R}^{W{\times}H{\times}3}\}$ be a set of images to cluster into $K$ groups. Following deep clustering literature~\cite{caron2018deep,van2020scan}, for a given deep feature extractor, a straightforward way to formulate a framework is to extract a holistic deep representation of an image and apply off-the-shelf clustering methods on unlabeled images.
While plausible, this approach does not take into account that anomalous behaviors may happen happening locally (e.g., Figure~\ref{fig:attention}). As a result, it shows suboptimal clustering performance (see Section~\ref{sec:ablation_patch_vs_holistic}).

To account for the local nature of anomalous patterns in images, we propose to represent images with a bag of patch embeddings, similar to recent works on visual anomaly detection~\cite{yi2020patch,rippel2021modeling,defard2021padim,roth2021towards}. Let $Z_{i}\,{\triangleq}\,Z(\mathbf{x}_{i})\,{=}\,\{\mathbf{z}_{1}^{i},...,\mathbf{z}_{M}^{i}\}$, where $\mathbf{z}_{m}^{i}\,{\in}\,\mathbb{R}^{D}$ is a patch embedding from an image using pretrained deep neural networks. As we have bags of patch embeddings, each of which is from a single image, while wanting to assign a cluster membership to each bag, we can formulate it as a multiple instance clustering (MIC)~\cite{zhang2009multi} problem. 
In particular, we propose an anomaly clustering framework that follows steps below:
\begin{enumerate}
    \setlength{\itemsep}{0pt}
    \setlength{\parskip}{1pt}
    \item Extract patch embeddings and define embeddings from an image as a \textit{bag}.
    \item Compute the distance between bags.
    \item Apply similarity-based clustering methods.
\end{enumerate}
Figure~\ref{fig:framework} visualizes the proposed framework. 
We note that it is crucial to define a proper distance measure for clustering. In what follows, we discuss distance measures between bags computed in unsupervised (Section~\ref{sec:method_unsup_alpha}) or semi-supervised (Section~\ref{sec:method_semisup_ac}) ways.

\iffalse
\begin{table}[t]
    \small
    \centering
    \resizebox{0.99\linewidth}{!}{%
    \begin{tabular}{c|cll}
        \toprule
        Setting & \multicolumn{3}{c}{Distance measure} \\
        \midrule
        \multirow{2}{*}{Unsup.} & Average & $\Vert\mathbb{E}_{m}[\mathbf{z}_{m}^{i}]-\mathbb{E}_{n}[\mathbf{z}_{n}^{j}]\Vert^2$ & \eqref{eq:distance_average_identity_reform}\\
        & Maximum H.~\cite{zhang2009multi} & $\max_{m}\min_{n}\Vert\mathbf{z}_{m}^{i}-\mathbf{z}_{n}^{j}\Vert^2$ & \eqref{eq:distance_hausdorff_single}\\
        %& Average H.~\cite{zhang2009multi} & $\mathbb{E}_{m}\min_{n}\Vert\mathbf{z}_{m}^{i}-\mathbf{z}_{n}^{j}\Vert^2$ \\
        \midrule
        Unsup. & \multirow{2}{*}{Weighted Average} & \multirow{2}{*}{$\Vert\sum_{m}\alpha_{m}^{i}\mathbf{z}_{m}^{i}-\sum_{n}\alpha_{n}^{j}\mathbf{z}_{n}^{j}\Vert^2$} &   \eqref{eq:distance_weighted_average_identity_reform},\eqref{eq:weighted_distance_unsup}\\
        Semi-sup. & & & \eqref{eq:distance_weighted_average_identity_reform},\eqref{eq:weighted_distance_semisup}\\
        \bottomrule
    \end{tabular}
    }
    \caption{Distance measures for anomaly clustering. We have nothing but test data to cluster for unsupervised setting, while for semi-supervised setting, we are given a few labeled normal data to help compute weight vectors $\bm{\alpha}$.}
    \label{tab:my_label}
\end{table}
\fi

\input{table/figure_03.tex}

%\vspace{-0.1in}
\subsection{Weighted Average Distance between Bags}
\label{sec:method_unsup_ac}
%\vspace{-0.05in}

For the unsupervised setting, we are given a data $\{\mathbf{x}_{i}\}$, or equivalently, bags of instances $\{Z_{i}\}$, to cluster without any label information. 
We are interested in grouping these data using off-the-shelf similarity-based clustering methods, and we need to define the distance between bags $d(Z_{i}, Z_{j})$.

%\subsubsection{Weighted Average Distance}
%\label{sec:method_unsup_ac_weighted_average}
%
There are at least two ways to compute the distance between bags. First, we compute distances between pairs of instances from two bags then aggregate. On the other hand, we aggregate instances to have a single representation for each bag and compute the distance. We take the second approach as it reduces the distance computation significantly.

We note that not all instances should contribute equal to the distance between bags. For example, we do not expect a patch embedding corresponding to the background that are common across both normal and abnormal data to represent an anomaly. Instead, we may want instances for anomalous patterns to contribute more to the distance. To this end, we propose a distance between weighted average embeddings of two bags as follows:
\begin{align}
    d_{\mathrm{WA}}(Z_i, Z_j) = &\, \Big\Vert\Big(\sum_{m=1}^{M}\alpha_{m}^{i}\mathbf{z}_{m}^{i}\Big) - \Big(\sum_{n=1}^{M}\alpha_{n}^{j}\mathbf{z}_{n}^{j}\Big)\Big\Vert\label{eq:distance_weighted_average_identity_reform}
\end{align}
where $\bm{\alpha}\,{\in}\,\Delta^{M}$ is a weight vector specifying which instance to attend to. %$d_{\mathrm{WA}}$ is equivalent to $d_{\mathrm{avg}}$ in Equation~\eqref{eq:distance_average_identity_reform} if both weight vectors $\bm{\alpha}^{i}$ and $\bm{\alpha}^{j}$ are $\frac{\bm{1}}{M}$. By formulation, we also note that weighted average distance measure satisfies both identity and symmetry properties. 

\subsubsection{Defining ${\alpha}$ Without Supervision.}
\label{sec:method_unsup_alpha}
The remaining question is how to define the weight $\alpha$. Intuitively speaking, we expect $\alpha$ to attend to discriminative instances, e.g., patch embeddings of defective regions, rather than those of normal regions.
In MIC literature~\cite{zhang2009multi,cheplygina2015multiple}, the maximum Hausdorff distance~\cite{huttenlocher1993comparing,dubuisson1994modified,edgar2007measure} has been a popular choice, whose distance metric is written as follows:
\begin{gather}
    d_{\mathrm{maxH}}(Z_{i},Z_{j}) = \max\big\{d(Z_{i},Z_{j}),d(Z_{j},Z_{i})\big\},\label{eq:distance_hausdorff}\\
    d(Z_{i},Z_{j}) = \max_{m=1,...,M}\min_{n=1,...,M} \big\{\Vert\mathbf{z}_{m}^{i} - \mathbf{z}_{n}^{j}\Vert\big\}\label{eq:distance_hausdorff_single}
\end{gather}
Eq.~\eqref{eq:distance_hausdorff_single} returns the maximum over instances in $Z_{i}$ of minimum distances to instances in $Z_{j}$, and is likely the distance between inhomogeneous instances of two bags, as shown in Figure~\ref{fig:attention}.
Moreover, when $\alpha_{m}^{i}$ and $\alpha_{n}^{j}$ are determined as below (with a bit abuse of notation), Eq.~\eqref{eq:distance_weighted_average_identity_reform} recovers Eq.~\eqref{eq:distance_hausdorff_single}.
\begin{equation}
    \begin{array}{c}
    \alpha_{m^{*}}^{i}\,{=}\,1, \; m^{*}\,{=}\,\argmax_{m}\min_{n}\big\{\Vert\mathbf{z}_{m}^{i}-\mathbf{z}_{n}^{j}\Vert\big\}\\
    \alpha_{n^{*}}^{j}\,{=}\,1, \; n^{*}\,{=}\,\argmin_{n}\big\{\Vert\mathbf{z}_{m^{*}}^{i}-\mathbf{z}_{n}^{j}\Vert\big\}
    \end{array}\label{eq:weighted_distance_to_maxh}
\end{equation}
One downside of maximum Hausdorff distance is that, as is clear from Eq.~\eqref{eq:weighted_distance_to_maxh}, it only focuses on the distance between a single instance from two bags. \cite{zhang2009multi} has proposed an average Hausdorff distance to account for such cases by taking an average instead of maximum of minimum distances, but we find them less suitable for anomaly clustering problem, as shown empirically in Section~\ref{sec:exp_unsup}.

Alternatively, we propose the soft weight as follows:
\begin{equation}
    \alpha_{m}^{i}\propto\exp\Big(\frac{1}{\tau}\mathbb{E}_{j\neq i}\big\{\min_{n}\Vert\mathbf{z}_{m}^{i} - \mathbf{z}_{n}^{j}\Vert\big\}\Big)\label{eq:weighted_distance_unsup}
\end{equation}
%\begin{equation}
%    \alpha_{m}^{i}\propto\exp\Big(\frac{1}{\tau}\min_{\mathbf{z} \in %Z_{-i}}\Vert\mathbf{z}_{m}^{i} - \mathbf{z}\Vert^2\Big)\label{eq:weighted_distance_unsup}
%\end{equation}
%where $Z_{-i} = \{ \mathbf{z}^j_n | j \neq i, n = 1,...,M \}$
%\begin{align*}
%    \alpha_{m}^{i}[1] & \propto\exp\Big(\frac{1}{\tau}\mathbb{E}_{j\neq i}\big\{\min_{n}\Vert\mathbf{z}_{m}^{i} - \mathbf{z}_{n}^{j}\Vert^2\big\}\Big)  \\
%    \alpha_{m}^{i}[2] & \propto\exp\Big(\frac{1}{\tau}\mathbb{E}_{j\neq i}\big\{\min_{n} \frac{1}{(1-\alpha_{n}^{j}[1])} \Vert\mathbf{z}_{m}^{i} - \mathbf{z}_{n}^{j}\Vert^2\big\}\Big)   \\
%    \alpha_{m}^{i}[K] & \propto\exp\Big(\frac{1}{\tau}\mathbb{E}_{j\neq i}\big\{\min_{n} \frac{1}{(1-\alpha_{n}^{j}[K-1])} \Vert\mathbf{z}_{m}^{i} - \mathbf{z}_{n}^{j}\Vert^2\big\}\Big)  
%\end{align*}
%
where $\tau$ controls the smoothness of $\alpha$. For example, when $\tau\,{\rightarrow}\,0$, we tend to focus on a single instance of $m$ with the maximum average minimum distance, while $\tau\,{\rightarrow}\,\infty$ evenly distributes weights across instances. 
%
%We use expectation ($\mathbb{E}_{j\neq i}$) over other aggregation operators such as max ($\max_{j\neq i}$) or min ($\min_{j\neq i}$) 
%
The inner $\min$ operator finds the most similar instance of other bags, and it allows instances that commonly occur across images in the dataset, e.g., patches relevant to non-defective regions, to be down-weighted when computing $\alpha$. 
While we choose to aggregate minimum distances with expectation ($\mathbb{E}_{j\neq i}$), there are alternatives, such as $\max$ ($\max_{j\neq i}$) or $\min$ ($\min_{j\neq i}$). However, $\max$ operator would suffer for aligned objects as some normal instances may not be found from an anomalous images. $\min$ operator would suffer if there are duplicates.
More explanations on these insights are in Appendix~\ref{sec:app_ablation_distance_measure}.
Finally, we ablate with combinations of operators in Section~\ref{sec:ablation_distance_measure}.

%
%We find that expectation ($\mathbb{E}_{j\neq i}$) works more robustly than max ($\max_{j\neq i}$) or min ($\min_{j\neq i}$) operators across categories, as in Section~\ref{sec:ablation_distance_measure}. 

%The computational complexities of maximum Hausdorff and the weighted average distances are similar in their current forms as they both require pairwise distances among instances. On the other hand, the computation for weighted average distance can be reduced by sampling a subset for $\mathbb{E}$.

%
We show ${\alpha}$ chosen by maximum Hausdorff distance criteria of Eq.~\eqref{eq:weighted_distance_to_maxh} in the second column, and those based on Eq.~\eqref{eq:weighted_distance_unsup} in the third column of Figure~\ref{fig:attention}. We observe that defective areas are highlighted for most cases, with an extra granularity for soft weights.

%\vspace{-0.05in}
\subsubsection{Defining $\alpha$ with Labeled Normal Data.}
\label{sec:method_semisup_ac}

As in Figure~\ref{fig:attention}, the highlighted regions from unsupervised distance measures are around defective areas. This motivates us to directly derive weight vectors that are designed to attend the defective areas.
Motivated by the recent success in semi-supervised defect localization~\cite{bergmann2020uninformed,yi2020patch,li2021cutpaste,roth2021towards}, we propose a semi-supervised anomaly clustering, where a few normal data are given to help compute weight vectors.
Specifically, we extend Eq.~\eqref{eq:weighted_distance_unsup} as follows:
\begin{equation}
    \alpha_{m}^{i}\propto\exp\Big(\frac{1}{\tau}\min_{\mathbf{z}{\in}Z^{\mathrm{tr}}}\Vert\mathbf{z}_{m}^{i} - \mathbf{z}\Vert\Big)\label{eq:weighted_distance_semisup}
\end{equation}
where $Z_{\mathrm{tr}}\,{=}\,\bigcup_{\mathbf{x}{\in}\mathcal{X}_{\mathrm{tr}}} Z(\mathbf{x})$ is a union of bags of normal data $\mathbf{x}\,{\in}\,\mathcal{X}_{\mathrm{tr}}$. Since we put all instances from bags of normal data we do not need expectation. 
We visualize weight vectors by Eq.~\eqref{eq:weighted_distance_semisup} in the fourth column of Figure~\ref{fig:attention}, followed by the ground-truth segmentation mask based weights.
%

%\vspace{-0.1in}
\subsection{Comparison to BAMIC~\cite{zhang2009multi}}
\label{sec:method_comp_to_bamic}
%\vspace{-0.05in}
BAMIC~\cite{zhang2009multi} is a multiple-instance clustering framework that requires a pairwise distance measure and the similarity-based clustering methods.
An instance in \cite{zhang2009multi} uses variants of Hausdorff measure to compute distances and k-medoids for clustering.
However, the method following \cite{zhang2009multi} (maxH and k-medoids) performs poorly on anomaly clustering as in Table~\ref{tab:main_clusterer_comparison}.
Besides improved clustering accuracy, our proposal has a few other advantages as we discuss below.

\vspace{0.04in}
\noindent\textbf{Time complexity.}
Let $N$ be the number of data. The time complexity of distance measures are written as follows:
\begin{eqnarray*}
    \text{Maximum Hausdorff : } & O(N^2 M^2 D) \\
    \text{Weighted Average : } & O(\underbrace{N^2 M^2 D}_{\text{Weights in Eq.~\eqref{eq:weighted_distance_unsup}}} + \underbrace{N^2 D}_{\text{Distance in Eq.~\eqref{eq:distance_weighted_average_identity_reform}}}) 
    %\text{WA: } & O(N^2 B^2 D + N^2 D) \\
    %\text{WA-semi: } & O(NL B^2 D + N^2 D) 
\end{eqnarray*}
While WA appears to be slightly more expensive due to the second term, it is negligible for large bag sizes ($M$). Importantly, it can be substantially reduced by subsampling the data when computing weights in Eq.~\eqref{eq:weighted_distance_unsup}:
\begin{equation}
    \alpha_{m}^{i}\propto \exp\Big(\frac{1}{\tau}\mathbb{E}_{Z_{\mathrm{sub}}\backslash\{i\}}\big\{\min_{n}\Vert \mathbf{z}_{m}^{i}-\mathbf{z}_{n}^{j}\Vert\big\}\Big),
\end{equation}
resulting in $O(N |Z_{\mathrm{sub}}| M^2 D + {N^2 D})$, which would be beneficial when $N\,{\gg}\, |Z_{\mathrm{sub}}|$.

\vspace{0.04in}
\noindent\textbf{Use of labeled normal data.}
In Section~\ref{sec:exp_semisup} we show that the semi-supervised WA distance measure could drastically lift the clustering performance using a small amount of labeled normal data (see Section~\ref{sec:ablation_labeled_data} for an ablation study). This is a unique feature of WA measure and such an extension has not been discussed in \cite{zhang2009multi}. Notably, the time complexity of semi-supervised WA measure is $O(N|Z_{\mathrm{tr}}| B^2 D + N^2 D)$, making our method scalable with large $N$.

%\vspace{-0.1in}
\section{Experiments}
\label{sec:exp}
%\vspace{-0.05in}

%We evaluate the anomaly clustering performance on industrial anomaly detection benchmarks, including MVTec dataset~\cite{bergmann2019mvtec}\footnote{\url{https://www.mvtec.com/company/research/datasets/mvtec-ad}} and magnetic tile defect (MTD) dataset~\cite{huang2020surface}.\footnote{\url{https://github.com/abin24/Magnetic-tile-defect-datasets.}}
We test the anomaly clustering using anomaly detection benchmarks, including MVTec dataset~\cite{bergmann2019mvtec} and magnetic tile defect (MTD) dataset~\cite{huang2020surface}.
%
%Below we provide a brief description for each dataset:
MVTec dataset has 10 object and 5 texture categories. The training set of each category includes normal (non-defective) images, whose number varies from 60 to 391. The test set contains both normal and anomalous (defective) images and anomalous images are grouped into 2$\sim$9 sub-categories by defect types. See Figure~\ref{fig:teasar_anomaly_clustering} for an example of anomaly sub-categories.
Magnetic Tile Defect dataset has 952 non-defective and 392 defective images. Defective images are grouped into 5 sub-categories, such as blowhole, break, crack, fray or uneven. %For anomaly detection, 80\% of normal images are used as a train set and the rest, along with defective images, are used as a test set~\cite{roth2021towards}.

\iffalse
\begin{itemize}
    \setlength{\itemsep}{0pt}
    \setlength{\parskip}{1pt}
    \item {\bf MVTec}~\cite{bergmann2019mvtec} dataset contains 10 object and 5 texture categories. The training set of each category consists of normal (non-defective) images, whose number varies from 60 to 391. The test set contains both normal and anomalous (defective) images and anomalous images are grouped into 2$\sim$9 sub-categories by their types. See Figure~\ref{fig:teasar_anomaly_clustering} for an example of anomaly sub-categories.
    \item {\bf Magnetic Tile Defect}~\cite{huang2020surface} dataset has 952 non-defective and 392 defective images. Defective images are grouped into 5 sub-categories, such as blowhole, break, crack, fray or uneven. %For anomaly detection, 80\% of normal images are used as a train set and the rest, along with defective images, are used as a test set~\cite{roth2021towards}.
\end{itemize}
\fi
%

\vspace{0.04in}
\noindent\textbf{Protocol.} 
We test unsupervised and semi-supervised clustering. For unsupervised case, no labeled data is provided for clustering, whereas for semi-supervised case, labeled normal data in the anomaly detection train set is given to compute ${\alpha}$ in Eq.~\eqref{eq:weighted_distance_semisup}. We emphasize that \emph{no labeled defective images are provided} until we compute evaluation metrics to report. All clustering experiments are under a transductive setting~\cite{gammerman1998learning,chapelle2006discussion} and no training is involved other than deep clustering experiments.

Note that we exclude \texttt{combined} sub-categories from evaluation metric computation as images of \texttt{combined} category may contain multiple defect types in a single image or their ground-truth labels may not be accurate~\cite{bergmann2019mvtec}.

\vspace{0.04in}
\noindent\textbf{Methods.}
We evaluate combinations of distance measures and clustering methods. For distance measure, we consider (uniform) average, variants of Hausdorff (Table~\ref{tab:exp_variants_dist_measure})~\cite{zhang2009multi}, and the proposed weighted average in unsupervised and semi-supervised ways. For clustering, k-means, GMM, spectral clustering, and hierarchical clustering with various linkage options are tested (Table~\ref{tab:main_clusterer_comparison}). BAMIC~\cite{zhang2009multi} is a special case, which combines variants of Hausdorff distance and k-medoids~\cite{park2009simple} for clustering.
Lastly, we make a comparison with state-of-the-art deep clustering methods, including IIC~\cite{ji2019invariant}, GATCluster~\cite{niu2020gatcluster}, and SCAN~\cite{van2020scan}. A comprehensive comparison to existing methods is in Section~\ref{sec:exp_deep_clustering}.

\vspace{0.04in}
\noindent\textbf{Metric.}
Normalized Mutual Information (NMI)~\cite{schutze2008introduction} and Adjusted Rand Index (ARI)~\cite{rand1971objective} are two popular metrics for clustering quality analysis when ground-truth cluster assignments are given for test set.
We also report the F1 score to account for the label imbalance. %\footnote{Concretely, we set an argument \texttt{average} to \texttt{weighted} to compute f1 score using \texttt{sklearn.metrics} library.}
The optimal matching between ground-truth and predicted cluster assignments are computed efficiently using Hungarian algorithm~\cite{kuhn1955hungarian}.
The maximum values of these metrics are 1.0 and higher values indicate a better clustering quality.

\iffalse
\paragraph{Model Selection.}
MVTec or MTD datasets do not contain a hold-out validation data, making the model selection difficult. Rather than optimizing hyperparameters on the test set, we conduct the model selection as follows:

\begin{enumerate}
    \setlength{\itemsep}{0pt}
    \setlength{\parskip}{1pt}
    \item Find a set of hyperparameter that works the best on the rest of the category.
    \item Evaluate the performance for the category of interest using hyperparameters found from step 1.
\end{enumerate}
%
We apply the procedure to all object and texture categories of MVTec dataset using an average of NMI, ARI and F1 scores as a performance metric. For MTD, texture categories of MVTec dataset is used for model selection.
\fi

\vspace{0.04in}
\noindent\textbf{Implementation.}
We use PyTorch~\cite{NEURIPS2019_9015} for neural network implementations and vision models~\cite{rw2019timm} and scikit-learn~\cite{scikit-learn} for clustering methods.

ImageNet pretrained WideResNet-50 (WRN-50) is used by default to extract patch embeddings, similarly to \cite{roth2021towards}. Specifically, we use the output of the second residual block followed by 3$\times$3 average pooling. Each patch embedding is then normalized to have unit L2 norm before fed into clustering method.
In addition, we conduct an extensive study with diverse pretrained networks (e.g., EfficientNet~\cite{tan2019efficientnet} and Vision Transformer (ViT)~\cite{dosovitskiy2020image}) in Appendix~\ref{sec:ablation_feature_extractor}.

\input{table/table_01.tex}

%\vspace{-0.1in}
\subsection{Unsupervised Clustering Experiments}
\label{sec:exp_unsup}
%\vspace{-0.05in}

In Table~\ref{tab:main_mvtec}, we report NMI, ARI, and F1 scores of unsupervised clustering methods on MVTec object, texture and MTD datasets. We test with diverse distance measures, including average (i.e., $\alpha\,{=}\,\frac{\bm{1}}{M}$ in Eq.~\eqref{eq:distance_weighted_average_identity_reform}), maximum Hausdorff in Eq.~\eqref{eq:distance_hausdorff}, and the proposed weighted average in Eq.~\eqref{eq:weighted_distance_unsup}. Finally, hierarchical clustering with Ward linkage~\cite{ward1963hierarchical} is used for clustering. A study using different clustering methods is in Section~\ref{sec:exp_deep_clustering}.

We confirm that the distance measure using average embeddings performs poorly,
%\footnote{We also evaluate the performance based on Equation~\eqref{eq:distance_average}, the pairwise average distance, which results in 0.196, 0.232, 0.071 NMIs for MVTec object, texture, MTD, respectively. The worse performance of the pairwise average distance might be because it does not satisfy the identity property.} 
%
whereas that based on discriminative instances chosen by max-min criteria of maximum Hausdorff distance significantly improves the performance. The proposed weighted average distance further improves the clustering NMI score by $\mathbf{0.041}$ on average. As shown in Figure~\ref{fig:attention}, generated weights attend to multiple discriminative instances instead of a single pair of instances, resulting in improved clustering accuracy.

%as shown in Figure~\ref{fig:attention}, our proposed weighted average attends to multiple discriminative instances instead of a single pair of instances of maximum Hausdorff distance, improving clustering NMI score by $\mathbf{0.041}$ on average.

\input{table/table_03.tex}

%\vspace{-0.1in}
\subsection{Comparison to Other Clustering Methods}
\label{sec:exp_deep_clustering}
%\vspace{-0.05in}

In this section, we report the clustering performance with various clustering methods under unsupervised setting. 
We test spectral clustering and hierarchical clustering with single, complete, and average linkages.
In addition, as the bag can be represented as a single aggregated embedding for weighted average distance, we test KMeans and Gaussian Mixture Model (GMM) with full covariance.
%Note that one advantage of weighted average distance is that the bag can be represented as a single aggregated embedding, making it compatible with popular clustering methods such as Kmeans or Gaussian Mixture Model (GMM).
%

%
Moreover, we test state-of-the-art deep clustering methods that learn deep representations and cluster assignments jointly. It has been studied extensively in recent years~\cite{xie2016unsupervised,yang2016joint,jiang2017variational,ji2019invariant,van2020scan,niu2020gatcluster} and demonstrated a strong performance over shallow counterparts in clustering object-centered images. 
%
%We test a few methods, including IIC~\cite{ji2019invariant},\footnote{\url{https://github.com/xu-ji/IIC}} GATCluster~\cite{niu2020gatcluster},\footnote{\url{https://github.com/niuchuangnn/GATCluster}} and SCAN~\cite{van2020scan}.\footnote{\url{https://github.com/wvangansbeke/Unsupervised-Classification}}
We study a few state-of-the-art methods, including IIC~\cite{ji2019invariant}, GATCluster~\cite{niu2020gatcluster}, and SCAN~\cite{van2020scan}.
Since we only have a few images per category, methods like SCAN that require a self-supervised pretraining may be suboptimal. In that case, we use the ImageNet pretrained model. Implementation details are in the Appendix~\ref{sec:app_deep_cluster_detail}.

The results are in Table~\ref{tab:main_clusterer_comparison}. We find that hierarchical clustering with Ward linkage is particularly effective, followed by the complete linkage. Linkages such as single or average that take into account distances between nearest neighbors between clusters do not perform well for anomaly clustering.
Spectral clustering appears to be moderately effective. As mentioned before, the weighted average distance is compatible with more scalable, center based clustering methods such as KMeans or GMM, though they perform a bit worse than hierarchical Ward clustering.
Finally, we note that the proposed weighted average distance shows higher NMIs for most cases regardless of clustering methods.

We find that state-of-the-art deep clustering methods do not work well on anomaly clustering. Even if we report the best performance chosen via early stopping based on the test set performance (numbers in the parentheses of Table~\ref{tab:main_clusterer_comparison}), the performance is not as good as our method. 
The suboptimal performance of deep clustering methods might be due to a lack of data, but requirement for a large amount of data could be their own limitation for industrial applications.

%\vspace{-0.1in}
\subsection{Semi-supervised Clustering Experiments}
\label{sec:exp_semisup}
%\vspace{-0.05in}
%
We test the semi-supervised clustering described in Section~\ref{sec:method_semisup_ac}. In this setting we are given labeled normal data from train set to compute instance weights of Eq.~\eqref{eq:weighted_distance_semisup}. Similarly, we use the hierarchical Ward clustering.
The results are described in Table~\ref{tab:main_mvtec}. We observe a significant boost in performance over unsupervised clustering methods. For example, we improve upon the best unsupervised clustering method by $\mathbf{0.050}$ in NMI on average.

Where is the improvement from? We hypothesize that weights derived in a semi-supervised way localize defective instances better than the unsupervised counterpart and make distance more meaningful, leading to an improved clustering accuracy. 
To answer this question, we visualize semi-supervised weights in Figure~\ref{fig:attention}. While the proposed unsupervised weights are already good at localizing defective areas, we find that it also has a few false positives (e.g., third row of Figure~\ref{fig:attention_tile}, fourth row of Figure~\ref{fig:attention_cable}). Whereas, semi-supervised weights effectively remove those false positives.
Moreover, we evaluate the pixel-level anomaly localization AUC, achieving 0.973 AUC with semi-supervised weights, improving upon 0.912 AUC of unsupervised weights. This suggests that the lift in clustering accuracy is from better localization of defective patches. We believe that more advanced defect localization and segmentation methods~\cite{lin2021cam} could improve the performance of anomaly clustering.

From this finding, we test using weights derived from the ground-truth segmentation masks,\footnote{We compute weights by resizing the ground-truth binary segmentation masks with 1 for anomalous and 0 for normal pixels into the same spatial dimension of patch embeddings and normalize their values to sum up to 1.} achieving 0.724, 0.685, and 0.467 NMIs for MVTec object, texture and MTD, respectively, further improving upon unsupervised and semi-supervised clustering performance.

\begin{figure*}[t]
    \centering
    \begin{subfigure}{0.27\textwidth}
        \includegraphics[width=\linewidth]{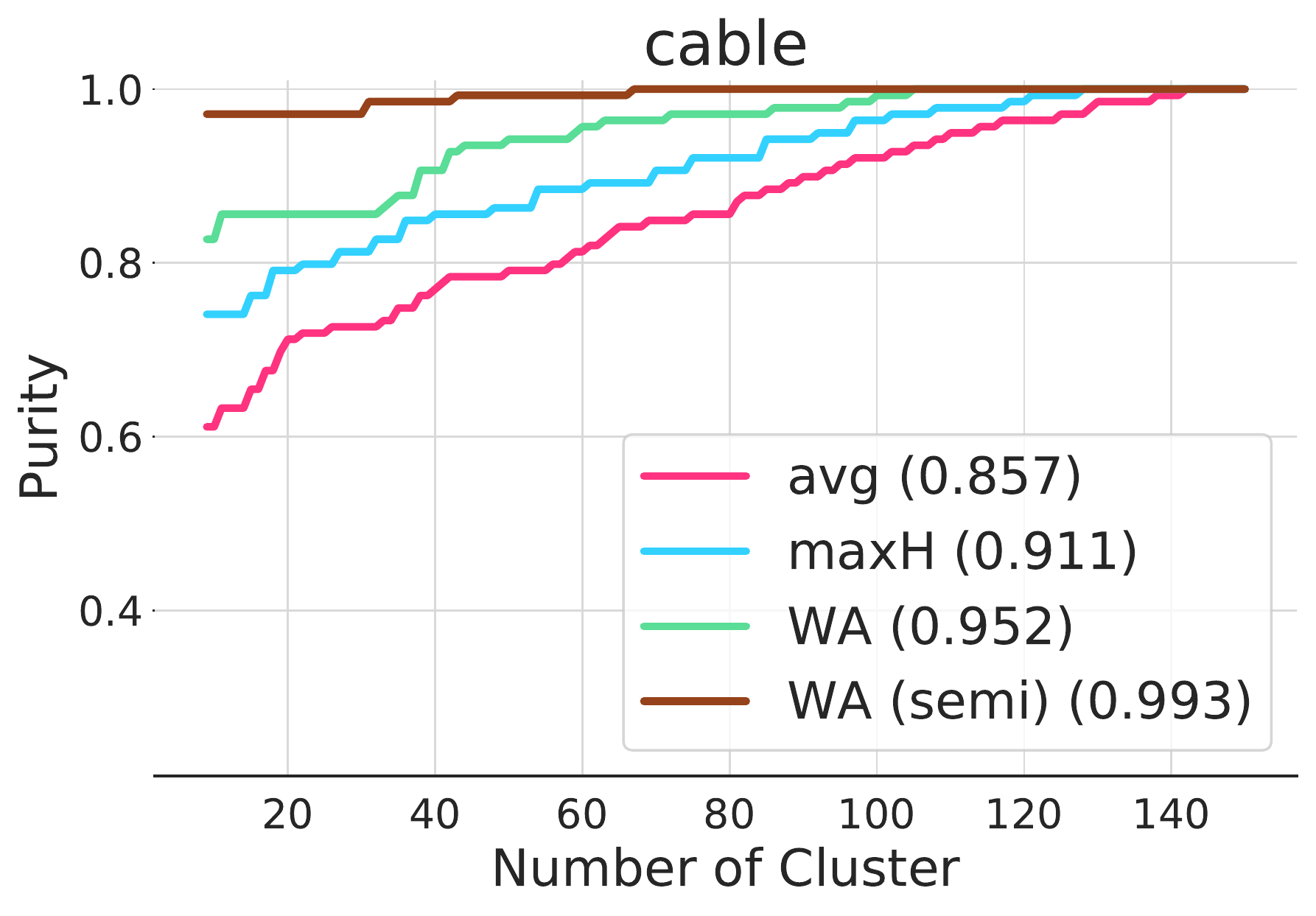}
    \end{subfigure}
    \hspace{0.1in}
    \begin{subfigure}{0.27\textwidth}
        \includegraphics[width=\linewidth]{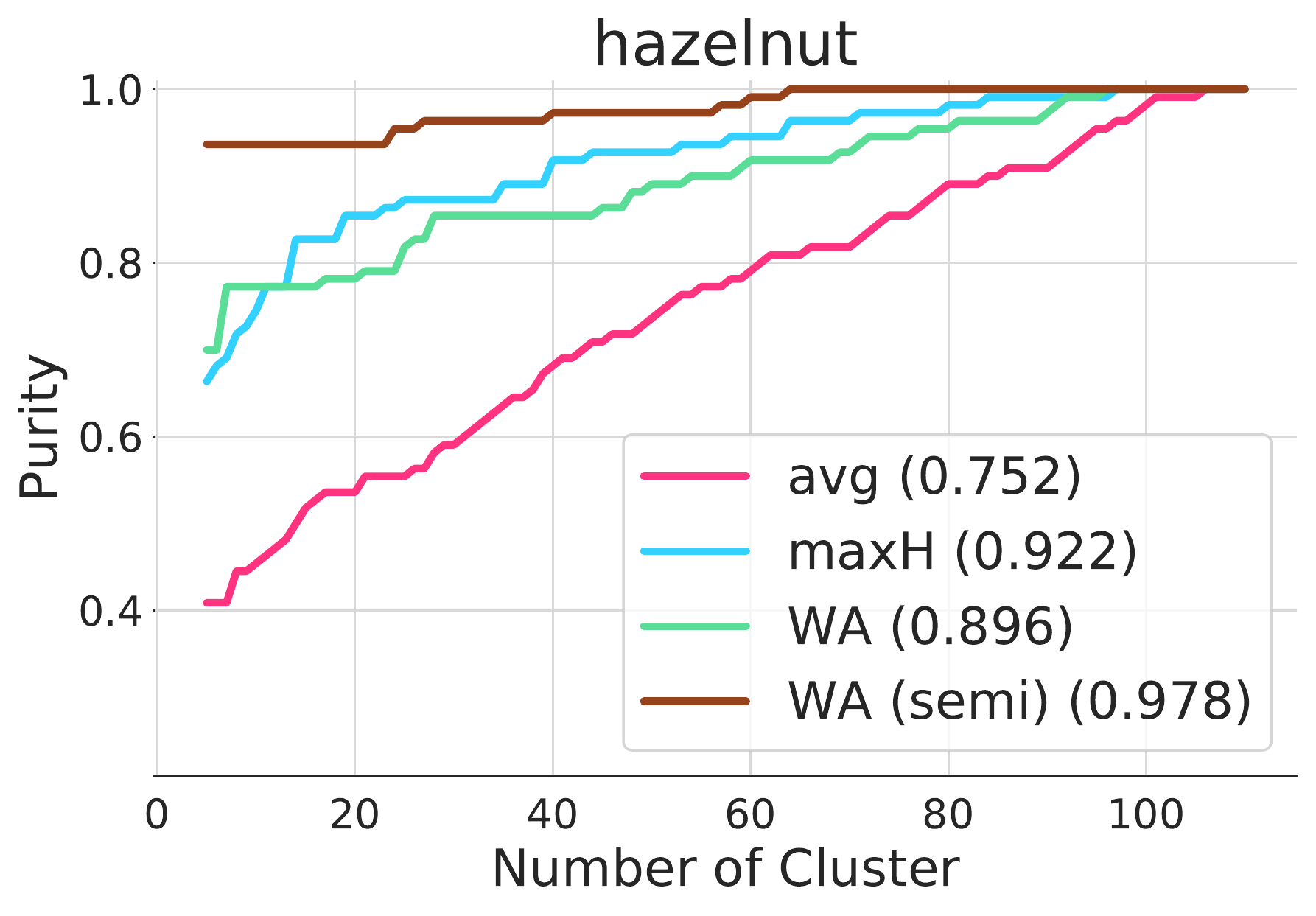}
    \end{subfigure}
    \hspace{0.1in}
    \begin{subfigure}{0.27\textwidth}
        \includegraphics[width=\linewidth]{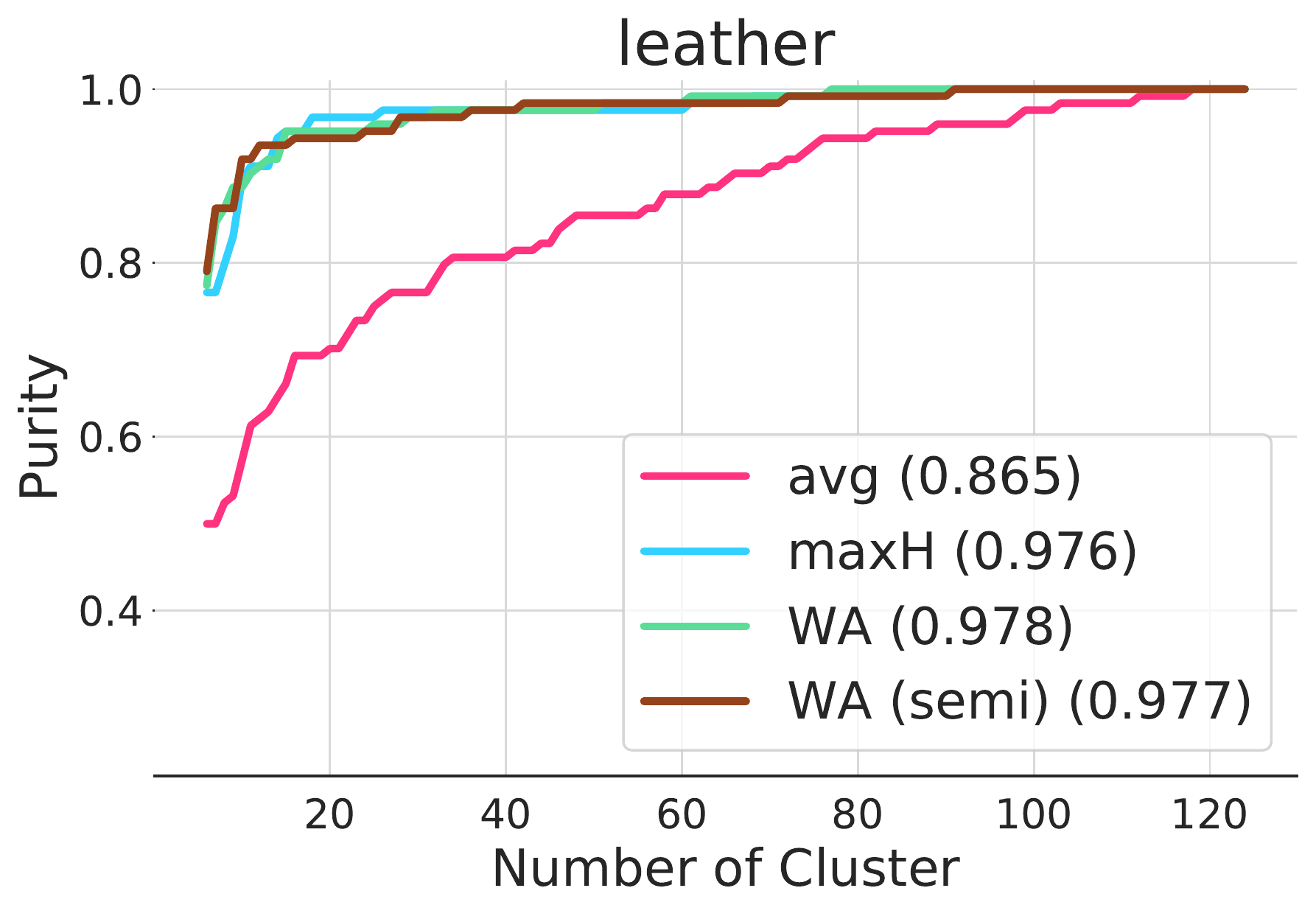}
    \end{subfigure}
    \caption{Purity of clusters on a few MVTec categories. Ward clustering is used for clustering method. Numbers in the bracket represent the area under the curve divided by the total number of examples (mAUC). Complete results are in Appendix~\ref{sec:app_exp_purity}.}
    \label{fig:purity}
    \vspace{-0.1in}
\end{figure*}

\begin{figure*}[t]
    \centering
    \begin{subfigure}{0.42\textwidth}
        \centering
        \includegraphics[width=\linewidth]{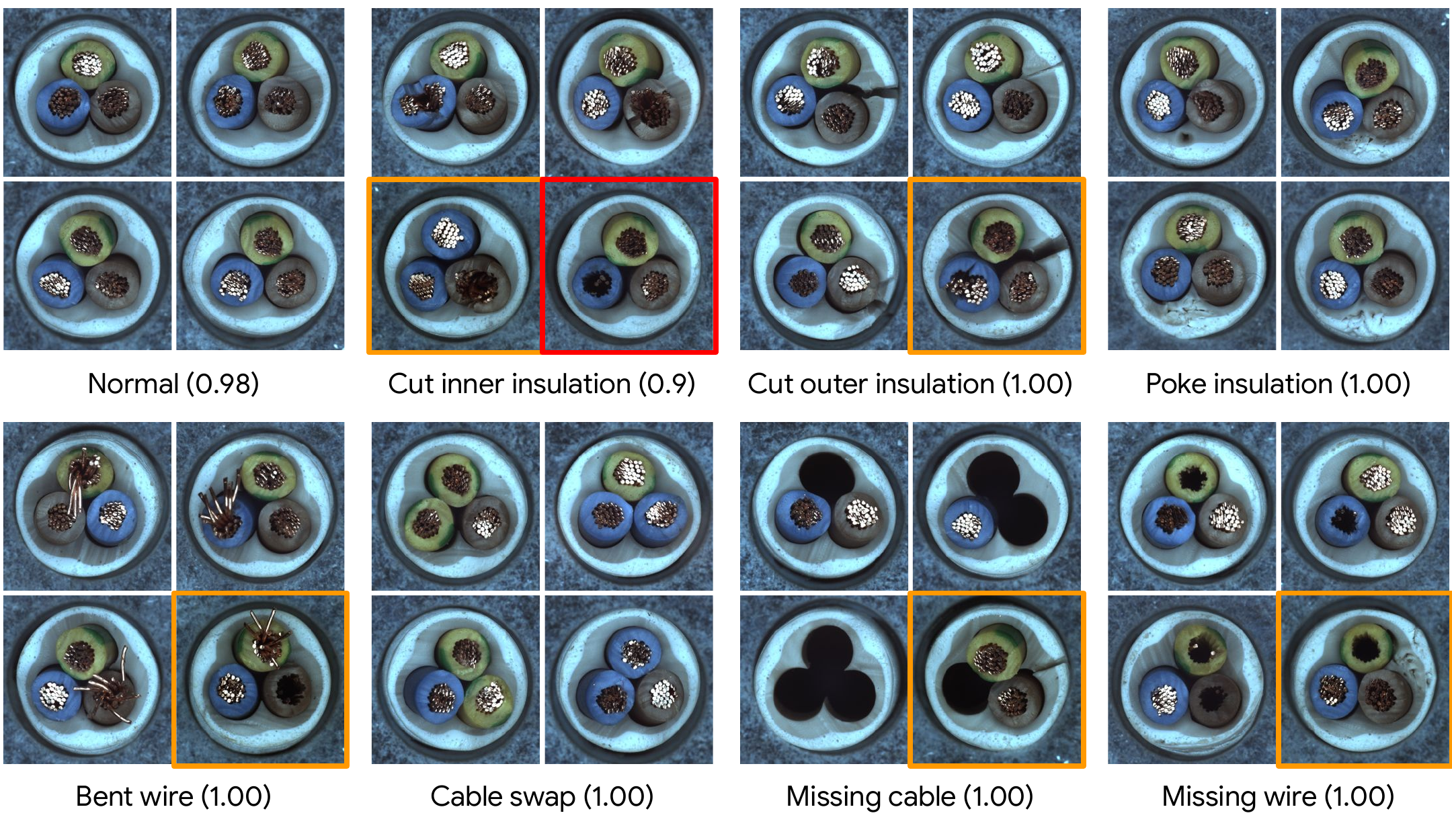}
        \caption{MVTec cable (purity: 0.97, \# sub-categories: 8)}
        \label{fig:cluster_visualize_cable}
    \end{subfigure}
    \hspace{0.25in}
    \begin{subfigure}{0.42\textwidth}
        \centering
        \includegraphics[width=\linewidth]{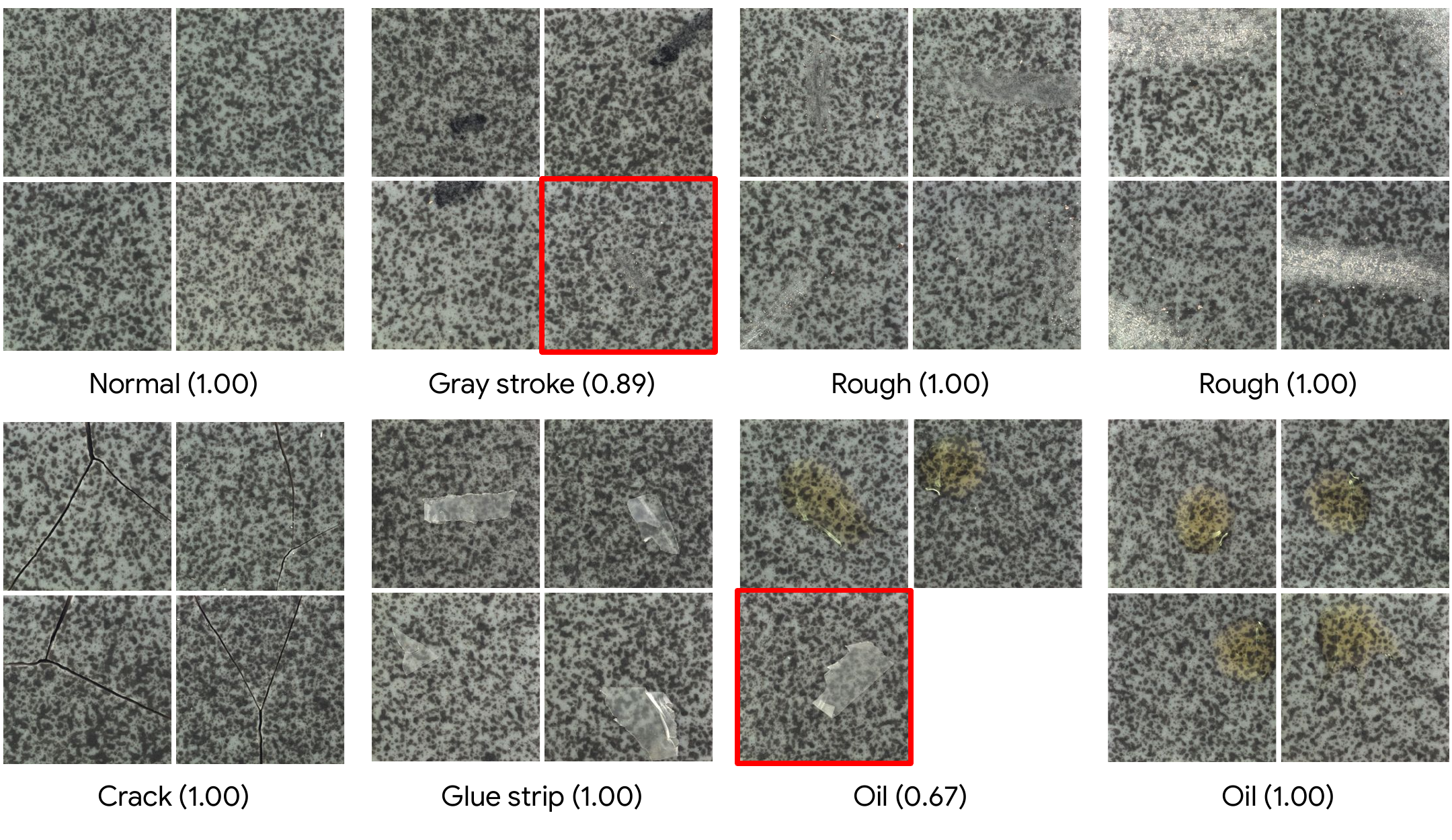}
        \caption{MVTec tile (purity: 0.97, \# sub-categories: 6)}
        \label{fig:cluster_visualize_tile}
    \end{subfigure}
    \vspace{-0.05in}
    \caption{Visualization of images in each cluster using semi-supervised WA distance and Ward clustering with 16 target clusters. We annotate the name of the major sub-category and the purity in parenthesis to each cluster. We highlight images with red if they do not belong to the major sub-category, and with orange when they contain multiple defect types.}
    \label{fig:cluster_visualize}
    \vspace{-0.2in}
\end{figure*}

%\vspace{-0.1in}
\subsection{Cluster Purity}
\label{sec:ablation_purity}
%\vspace{-0.05in}

While we report clustering accuracy with known number of clusters in Section~\ref{sec:exp_unsup} and \ref{sec:exp_semisup}, the number of cluster may not be available in practice. What could be important is the purity of clusters when data is over-clustered. For example, the labeling effort could be reduced from the number of data to the number of clusters if we can achieve a high purity.

Figure~\ref{fig:purity} shows the cluster purity with different number of target clusters for Ward clustering on a few MVTec categories. We see a clear gain in purity with the proposed clustering framework (brown, green, light blue) over the baseline (pink).
Moreover, we report purity metrics in Table~\ref{tab:purity}, including mAUC, the area under the curve divided by the total number of examples, and R@P, the reduction in the number of clusters at a given purity.\footnote{$\mbox{R@P}=1 - \nicefrac{\mbox{(\# clusters required to reach purity P)}}{(\mbox{\# data})}$.}
We confirm that the proposed framework improves the purity. For example, we improve R@0.95 on object categories from $0.231$ to $0.527$, meaning that we can reduce the number of clusters to label by 53\% (compared to exhaustively labeling all images) while retaining 95\% cluster purity. %Among distance measures, we find maximum Hausdorff distance shows higher purity than weighted average distance, though its purity could be improved with a few labeled normal data in semi-supervised setting.

\input{table/table_02.tex}

%\vspace{-0.1in}
\subsection{Cluster Visualization}
\label{sec:exp_visualization}
%\vspace{-0.05in}

In Figure~\ref{fig:cluster_visualize}, we show images of discovered clusters. We over-cluster with 16 clusters using semi-supervised WA distance and hierarchical Ward clustering. We annotate the major defect types to each cluster and the purity in parenthesis. %We provide more visualizations with different distance measures in the Appendix.

We verify from Figure~\ref{fig:cluster_visualize} that clusters are fairly pure and images with the same or similar type of defects are grouped together. This is because our proposed distance measure is able to attend to discriminative defective areas to compute the distance between images. While some images are clustered incorrectly (highlighted in red), they do not seem too different to other images in the same cluster. 
Another interesting observation is that two ``rough'' clusters in Figure~\ref{fig:cluster_visualize_tile} indeed show somewhat distinctive textures and our method is able to pick such a fine-grained difference to cluster them separately.
Finally, there are some images that contain more than one defect type highlighted in orange. For example, in Figure~\ref{fig:cluster_visualize_cable}, the one in ``bent wire'' cluster not only has bent wires but also a missing wire. It is promising that our method at least groups it into one of two correct candidate clusters. We leave a multi-label anomaly clustering, which could assign multiple cluster labels to an image with multiple defect types, as a future work.

%\vspace{-0.15in}
\section{Ablation study}
\label{sec:ablation}
%\vspace{-0.1in}

In this section we conduct in-depth study of the proposed anomaly clustering framework. %In Section~\ref{sec:ablation_patch_vs_holistic} we highlight the importance of patch-based MIC framework over a holistic representation. In Section~\ref{sec:ablation_soft_vs_hard} we ablate $\tau$ that controls the sparseness of weights. In Section~\ref{sec:ablation_distance_measure} we study variants of unsupervised distance measures. 
Due to space constraint, we provide extra study, such as an impact of number of labeled normal data (Section~\ref{sec:ablation_labeled_data}) or diverse feature extractors (Section~\ref{sec:ablation_feature_extractor}), in the Appendix.

%In this section we conduct in-depth study of the proposed anomaly clustering framework. In Section~\ref{sec:ablation_soft_vs_hard} we introduce a variant of weights based on the hard instance selection and ablate their hyperparameters that control the sparseness of weights. In Section~\ref{sec:ablation_distance_measure} we study with variants of unsupervised distance measures, and with diverse feature extractors in Section~\ref{sec:ablation_feature_extractor}. Finally, we study the impact of extra unlabeled normal data on unsupervised clustering in Section~\ref{sec:ablation_extra_normal}.

%\vspace{-0.1in}
\subsection{Patch vs Holistic Representation}
\label{sec:ablation_patch_vs_holistic}
%\vspace{-0.05in}
%
We highlight the importance of patch embeddings with multiple instance clustering over the holistic representation. 
For the holistic representation, we use the last hidden layer of WRN-50 after average pooling, resulting in $2048$ dimensional vector, as is commonly used in deep clustering literature~\cite{van2020scan}. For fair comparison, we use the same hidden layer for patch embeddings but without average pooling. For example, we obtain $8{\times}8$ $2048$ dimensional patch embeddings for input of size $256{\times}256$.

In Table~\ref{tab:abl_patch_vs_holistic}, we find that holistic representations, though better than learning-based deep clustering methods in Section~\ref{sec:exp_deep_clustering}, perform worse than our proposed patch-based multiple instance clustering methods. We observe similar trends using various ResNet~\cite{he2016deep,zagoruyko2016wide} and EfficientNet~\cite{tan2019efficientnet} models, whose results are in Appendix~\ref{sec:app_exp_patch_vs_holistic}.

\input{table/table_10.tex}

%\vspace{-0.1in}
\subsection{Sensitivity Analysis on $\tau$}
\label{sec:ablation_soft_vs_hard}
%\vspace{-0.05in}

Weights in Eq.~\eqref{eq:weighted_distance_unsup} and \eqref{eq:weighted_distance_semisup} play an important role in anomaly clustering. Specifically, both formulations involve the hyperparameter $\tau$ that controls the smoothness of the distribution of $\alpha$, which we ablate in this section.
Moreover, we study the variant of weights, called hard weights, where we select $k$ most discriminative instances (instead of softly weighing them) in Appendix~\ref{sec:ablation_hard_selection}.

Figure~\ref{fig:ablation_tau} presents the sensitivity analysis of $\tau$. It shows a trend that intermediate values of $\tau$ are preferred and the performance deteriorates as we increase their values as the model converges to uniform weights. Texture classes still shows outstanding performances even with small $\tau$ as they can focus on the smaller regions, which is consistent with our observation that some texture anomalies are tiny in size.
%
%Between soft and hard weights, we find that the soft version performs better as they assign different weights to instances, while hard one assigns a uniform weight to top-$k$ instances. 

%\vspace{-0.1in}
\subsection{Variants of Distance Measure}
\label{sec:ablation_distance_measure}
%\vspace{-0.05in}

Variants of Hausdorff distance metrics are proposed to compute similarities between bags. \cite{cheplygina2015multiple} present variants by replacing $\max$ or $\min$ operators of Eq.~\eqref{eq:distance_hausdorff} and \eqref{eq:distance_hausdorff_single}. For example, one can replace $\max$ operator in Eq.~\eqref{eq:distance_hausdorff} into $\mathrm{mean}$ as suggested in \cite{dubuisson1994modified}. %Replacing $\max$ and $\min$ of Eq.~\eqref{eq:distance_hausdorff_single} into $\mathrm{mean}$ recovers average distance of Eq.~\eqref{eq:distance_average}.
Exact formulations are in Appendix~\ref{sec:app_hausdorff_formulations}.

We report results using variants of Hausdorff distance in Table~\ref{tab:exp_variants_dist_measure} (top). For asymmetric distance measure such as $\max\min$ or $\mathrm{mean}\min$ of Eq.~\eqref{eq:distance_hausdorff_single}, aggregating them by $\max$ for Eq.~\eqref{eq:distance_hausdorff} shows better performance. 
Replacing the first $\max$ in Eq.~\eqref{eq:distance_hausdorff_single} into $\min$ or $\mathrm{mean}$ degrades the performance, as it deludes the attention to non-discriminative instances, which is critical for clustering data based on anomaly types. 

We study variants of unsupervised weight by replacing $\mathbb{E}$ in Eq.~\eqref{eq:weighted_distance_unsup} into $\max$ or $\min$. The results are in Table~\ref{tab:exp_variants_dist_measure} (bottom). We find that $\mathbb{E}$ works the robustly across datasets. We provide more qualitative analysis in Appendix~\ref{sec:app_ablation_distance_measure}.

\input{table/table_05-2.tex}

%\vspace{-0.1in}
\section{Conclusion}
\label{sec:conclusion}
%\vspace{-0.05in}

We introduce anomaly clustering, a challenging problem that existing approaches like deep clustering do not work well on. 
We propose to frame it as a multiple instance clustering problem by taking into account certain characteristics of industrial defects and present a novel distance function that focuses on the defective regions when exist. Experimental results show our proposed framework is promising.
Future directions include an extension to multiple instance deep clustering and active anomaly classification. 
We believe that the proposed framework is not only effective for anomaly clustering, but could also be useful for clustering images of fine-grained deformable objects.

\iffalse
\section{Limitations}
\label{sec:limitation}

The proposed work relies on the assumption that images are largely similar and defects are local. If anomalous patterns are global (e.g., global color contamination) a holistic representation may work as well. The reliance on deep image representations could be another limitation. Though it is shown to transfer well to some industrial applications, there is no principled way to measure the transfer-ability of these representations to novel problems. Data-driven approaches (e.g., deep clustering) can be explored in the future.
\fi

%\paragraph{Acknowledgment.}
%We thank Sercan Arik for the proofread of our manuscript.

%%%%%%%%% REFERENCES
\clearpage

\bibliographystyle{ieee_fullname}
\bibliography{egbib}

\newpage

\onecolumn

\appendix

\section{Additional Experimental Results}
\label{sec:app_exp}

\subsection{Variants of Weight with Hard Instance Selection}
\label{sec:ablation_hard_selection}

\begin{figure*}[t]
    \centering
    \begin{subfigure}{0.49\linewidth}
        \centering
        \includegraphics[width=\linewidth]{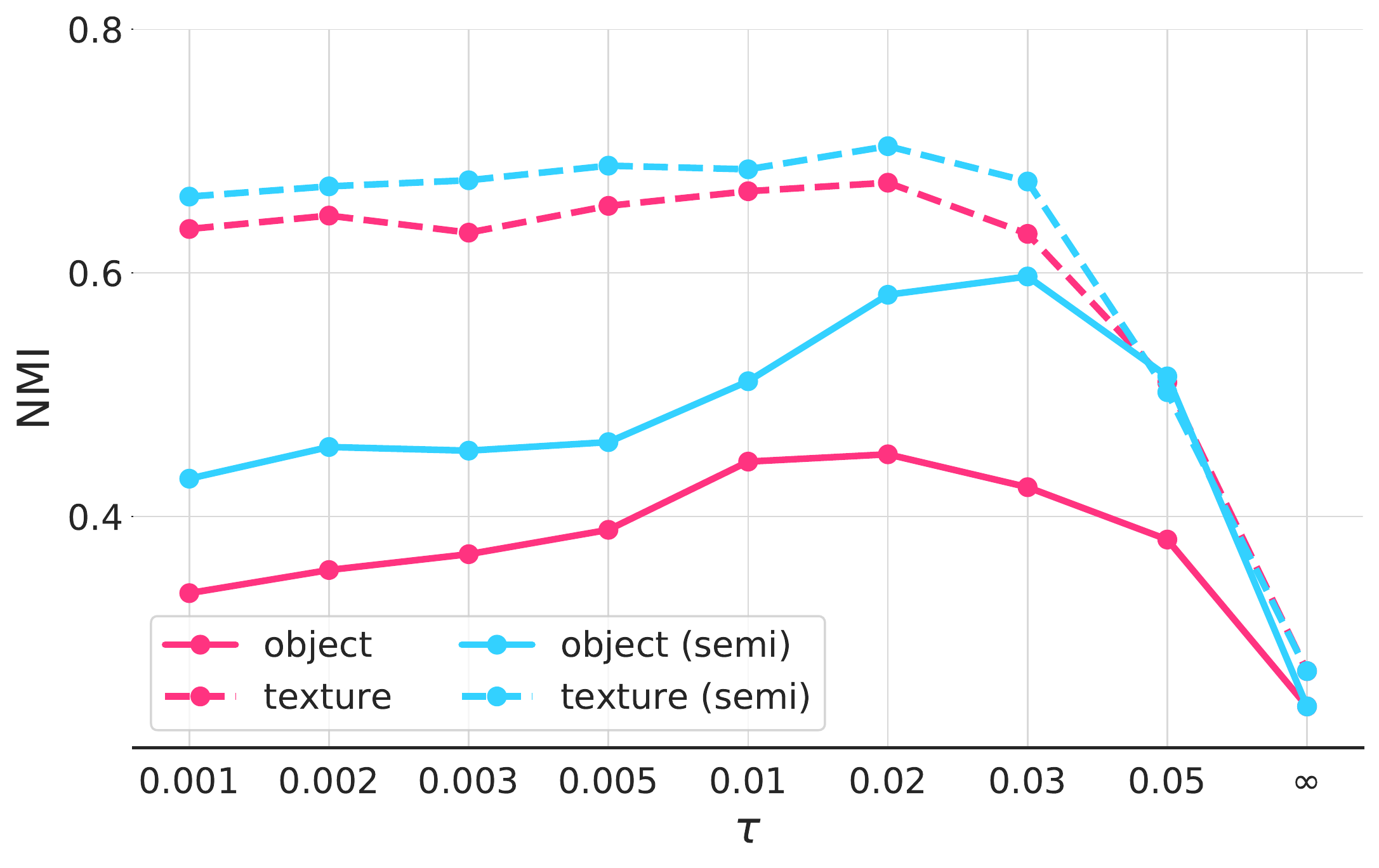}
        \caption{NMI-vs-$\tau$}\label{fig:supp_ablation_tau}
    \end{subfigure}
    \begin{subfigure}{0.49\linewidth}
        \centering
        \includegraphics[width=\linewidth]{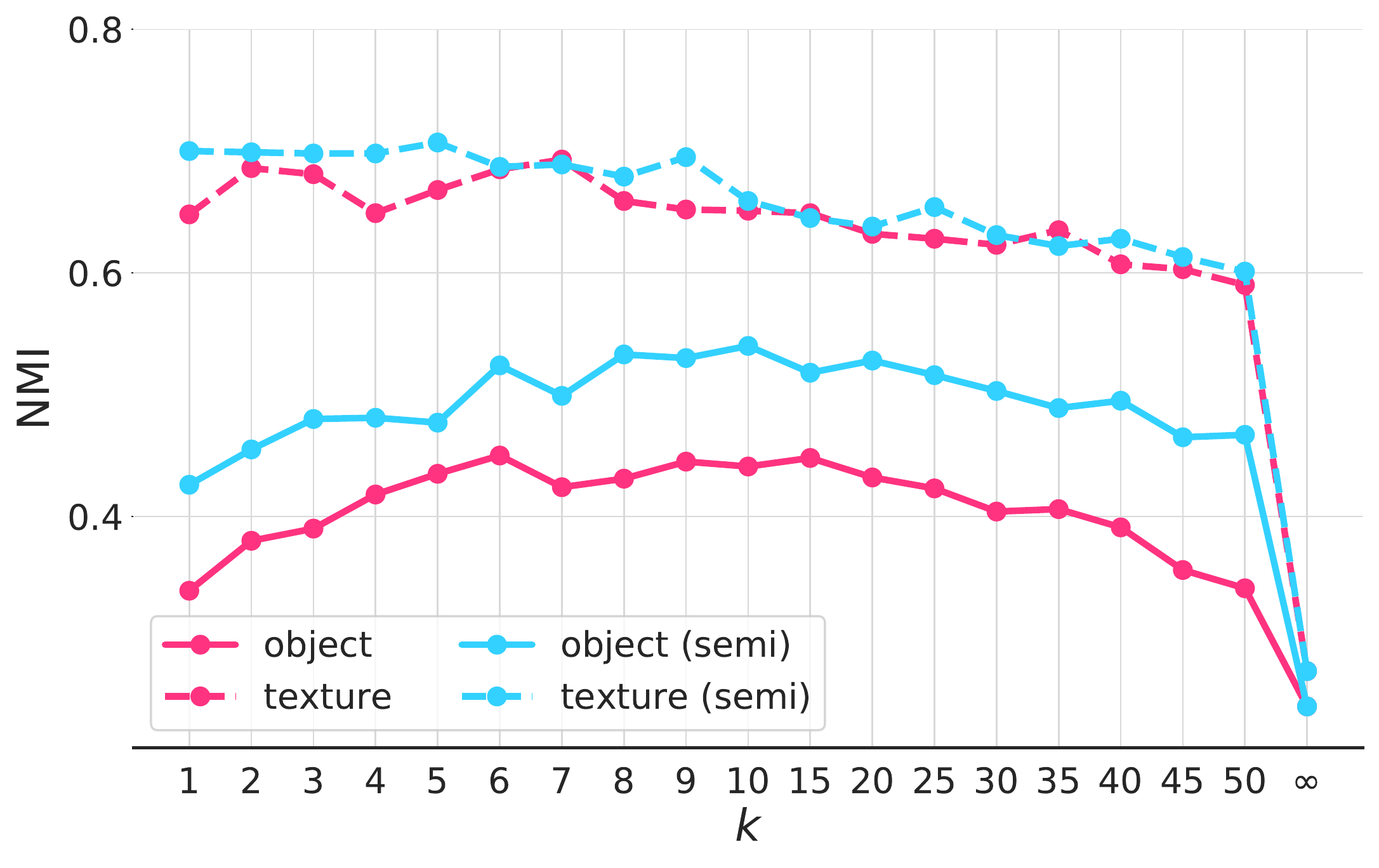}
        \caption{NMI-vs-$k$}\label{fig:supp_ablation_k}
    \end{subfigure}
    \caption{Sensitivity analysis of $\tau$ and $k$ on MVTec dataset.}
    \label{fig:supp_ablation_k_tau}
\end{figure*}

In Eq.~\eqref{eq:weighted_distance_unsup} we propose a soft weight. Alternatively, we consider a hard weight, where top-$k$ instances receive entire weight while weights for the rest of instances are set to 0. Specifically, let $I\,{=}\,\mathrm{argsort}\Big[\mathbb{E}_{j\neq i}\big\{\min_{n}\Vert\mathbf{z}_{m}^{i} - \mathbf{z}_{n}^{j}\Vert\big\}\Big]$, a sorted list in descending order. Let $I_{k}$ is a set of indices including the first $k$ items of the list $I$. The hard weight is defined as:
\begin{equation}
    \alpha_{m}^{i} = \frac{1}{k}\bm{1}\{m\,{\in}\, I_{k}\}
\end{equation}
We present results in Figure~\ref{fig:supp_ablation_k}. Overall we observe a similar trend with the soft weight in Figure~\ref{fig:supp_ablation_tau}. For example, both $\tau$ and $k$ work robustly when their values are small for texture categories. For object categories we generally require a bit larger $\tau$ or $k$ to obtain an optimal performance, as defective regions could be sometimes larger and even global. Between hard and soft weights, we find that soft weights are slightly better as it still assigns different weights to instances while hard weight assigns uniform weights to top-$k$ instances. 
One could develop to take the best of both worlds as follows:
\begin{equation}
    \alpha_{m}^{i} \propto \exp\Big(\frac{1}{\tau}\mathbb{E}_{j\neq i}\big\{\min_{n}\Vert\mathbf{z}_{m}^{i} - \mathbf{z}_{n}^{j}\Vert\big\}\Big), \;m\,{\in}\,I_{k}\; \mbox{ or 0 otherwise.}
\end{equation}

\subsection{Analysis on Labeled Normal Data Size}
\label{sec:ablation_labeled_data}

In this section we study the impact of the number of labeled normal data on the clustering performance of semi-supervised weighted average distance. Specifically, we vary the number of labeled normal data used to compute the weight of Eq.~\eqref{eq:weighted_distance_semisup}.

\begin{figure}[t]
    \centering
    \includegraphics[width=0.5\linewidth]{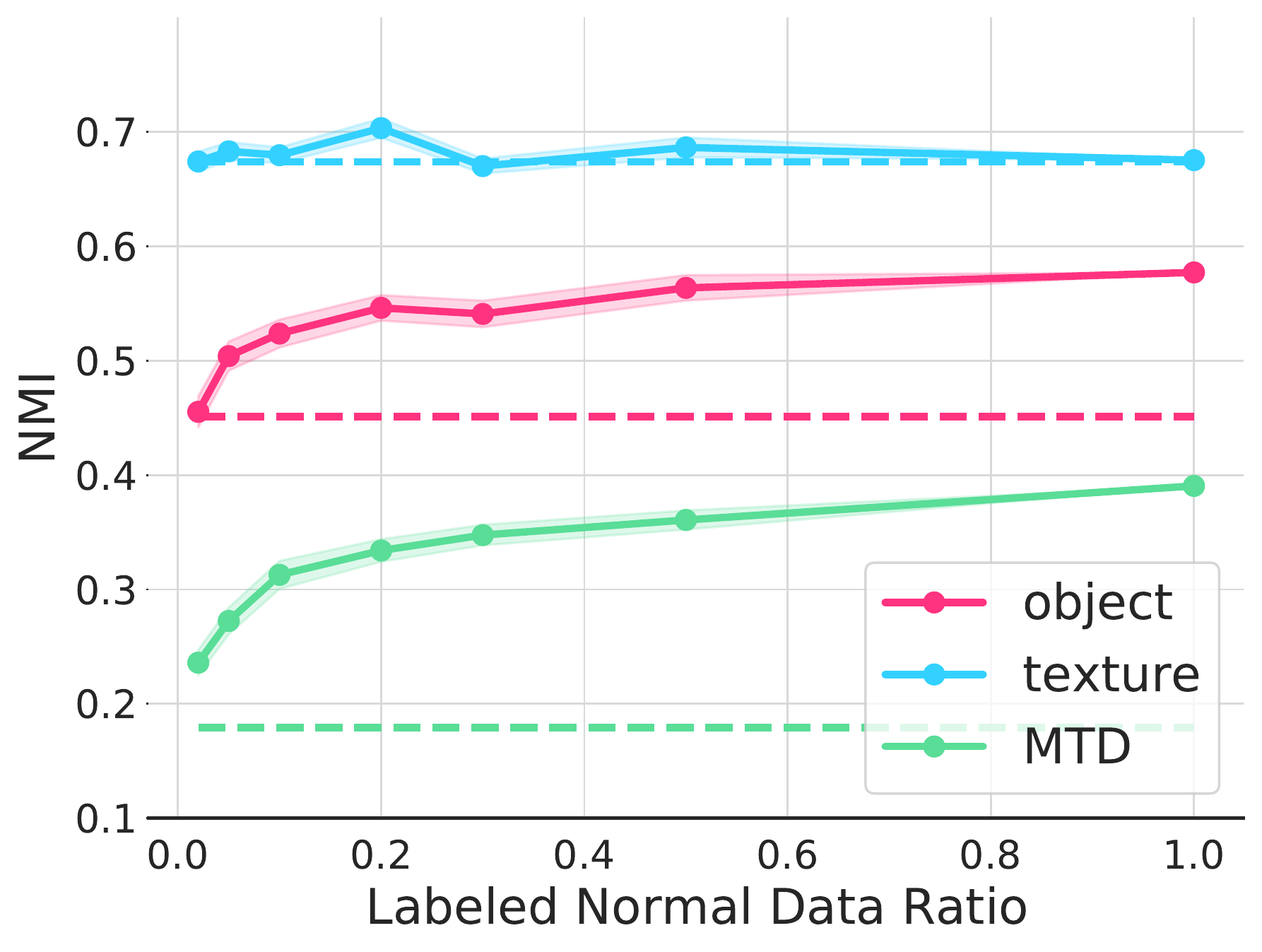}
    \caption{NMI scores of semi-supervised anomaly clustering with varying ratios of labeled normal data. Plots with dotted line represent unsupervised anomaly clustering results with the proposed weighted distance.}
    \label{fig:abl_nmi_vs_labeled_normal}
\end{figure}

The summary results are in Figure~\ref{fig:abl_nmi_vs_labeled_normal}. We also plot the performance of unsupervised version of Eq.~\eqref{eq:weighted_distance_unsup}. For object and MTD we find a clear trend of performance improvement as we increase the number of labeled normal data, while for texture the performance does not change much. Since acquiring labeled normal data is a lot cheaper than acquiring labeled anomaly data of multiple types, our results suggest a relatively inexpensive way to improve the clustering performance with a minimal supervision. For example, 20\% of labeled normal data for object categories of MVTec dataset corresponds to around 50 images.

\subsection{Patch vs Holistic Representation}
\label{sec:app_exp_patch_vs_holistic}

We provide results comparing the clustering performance of holistic and patch-based representations using ResNet~\cite{he2016deep,zagoruyko2016wide} and EfficientNet~\cite{tan2019efficientnet} models. Specifically, we conduct experiments using ResNet with various depths (18, 50, 101, 152) and EfficientNet with various sizes (B0, B4, B7).
Summary results are in Figure~\ref{fig:app_patch_vs_holistic}. For all accumulated bar plots over three datasets, we observe consistent trend of improved anomaly clustering performance using patch-based representations (second, third and fourth columns, with maximum Hausdorff, weighted average and semi-supervised version of that, respectively) over a holistic representation (first column).

\input{table/figure_08.tex}
\input{table/figure_06.tex}

\subsection{Feature Extractor}
\label{sec:ablation_feature_extractor}

We study the performance of anomaly clustering for various feature extractors, including ResNet~\cite{he2016deep,zagoruyko2016wide}, EfficientNet~\cite{tan2019efficientnet}, and Vision Transformer (ViT)~\cite{dosovitskiy2020image}. All aforementioned models are trained on ImageNet~\cite{deng2009imagenet}. We provide which layer and average pooling kernel size have been used for each network in Table~\ref{tab:app_network_config}.

\begin{table}[ht]
    \centering
    \caption{Implementation details on the layer and average pooling kernel size used for each network architecture.}
    \label{tab:app_network_config}
    \vspace{0.05in}
    \small
    \begin{tabular}{c|c|c|c|c|c|c}
        \toprule
        Network & ResNet & EfficientNet & ViT-T & ViT-S & ViT-B & ViT-L \\
        \midrule
        Layer used & ResBlock 2 & Reduction 3 & \multicolumn{3}{c|}{Block 7} & Block 13\\
        Kernel size & \multicolumn{2}{c|}{3${\times}$3} & \multicolumn{4}{c}{1${\times}$1}\\
        \bottomrule
    \end{tabular}
\end{table}

The results are in Figure~\ref{fig:abl_feature_extractor_resnet} and \ref{fig:abl_feature_extractor_others}. We plot accumulated NMI scores of average distance (i.e., $\alpha\,{=}\,\frac{\mathbf{1}}{M}$), maximum Hausdorff, and weighted distance without and with labeled normal data. It is clear that the proposed multiple instance clustering framework outperforms a single instance clustering via average distance. We observe weighted average improves upon maximum Hausdorff for many cases. Moreover, semi-supervised version of weighted average distance significantly improves the performance.

\iffalse
\subsection{Unsupervised Clustering with Extra Normal Data}
\label{sec:ablation_extra_normal}
%
In Section~\ref{sec:exp_unsup}, we use images in the test split for clustering, while leaving non-defective images in the train split unused. Here, we study an impact of extra \emph{unlabeled} normal data for unsupervised clustering. Note that the setting is different from semi-supervised setting of Section~\ref{sec:exp_semisup}, where we assume \emph{labeled} normal data is given.

%
Specifically, we study the impact of extra normal data on the quality of weights $\alpha$. While clustering analysis remains the same (e.g., clustering and evaluation are done on the test split only), we compute $\alpha$ by comparing distances against all data including extra normal data as follows:
%
\begin{equation}
    \alpha_{m}^{i}\propto\exp\Big(\frac{1}{\tau}\mathbb{E}_{j\neq i,\mathbf{x}_j\in\mathcal{X}\cup\mathcal{X_{\mathrm{tr}}}}\big\{\min_{n}\Vert\mathbf{z}_{m}^{i} - \mathbf{z}_{n}^{j}\Vert^2\big\}\Big)\label{eq:weighted_distance_unsup_extra}
\end{equation}
%
Note the change from Eq.~\eqref{eq:weighted_distance_unsup} that the expectation is over all data in $\mathcal{X}$ and $\mathcal{X}_{\mathrm{tr}}$. The results are in Table~\ref{tab:abl_extra_train_for_weights}.

\input{table/table_06.tex}
\fi

\subsection{Results on Purity with Overclustering}
\label{sec:app_exp_purity}

We provide additional results on the purity of clusters with overclustering in Figure~\ref{fig:overclustering_auc} on MVTec dataset. We also present the area under the curve divided by the total number of examples (mAUC) in the bracket of each legend. As we see in Figure~\ref{fig:overclustering_auc}, we observe significantly higher purity with our proposed clustering framework (brown, green, light blue) over the baseline (pink) for most cases.

\section{Additional Analysis with Variants of Distance Measure}
\label{sec:app_ablation_distance_measure}

We provide additional qualitative reasons on why $\max$ or $\min$ operators perform less robust than $\mathbb{E}$ when computing unsupervised weights of Eq.~\eqref{eq:weighted_distance_unsup}. 
Firstly, the downside of $\min$ operator is clear from the formulation. To be clear, we write the formulation as follows:
\begin{equation}
    \alpha_{m}^{i}\propto\exp\Big(\frac{1}{\tau}\min_{j\neq i}\big\{\min_{n}\Vert\mathbf{z}_{m}^{i} - \mathbf{z}_{n}^{j}\Vert\big\}\Big)\label{eq:weighted_distance_unsup_min}
\end{equation}
Let an image $\mathbf{x}_{i}$ is a duplicate of $\mathbf{x}_{j}$, i.e., $\mathbf{x}_{i}\,{=}\,\mathbf{x}_{j}$. Then, for any $\mathbf{z}_{m}^{i}$, we can always find $\mathbf{z}_{n}^{j}$ whose distance is $0$. In other words, $\min_{n}\Vert\mathbf{z}_{m}^{i} - \mathbf{z}_{n}^{j}\Vert\,{=}\,0$ for all $m$, and we get an uniform weight $\alpha_{m}^{i}\propto\exp(0)$. This is problematic if $\mathbf{x}_{i}$ is indeed an anomalous image as $\alpha$ does not provide any meaningful signal to attend to the defective area.

\begin{figure}[ht]
    \centering
    \includegraphics[width=0.5\linewidth]{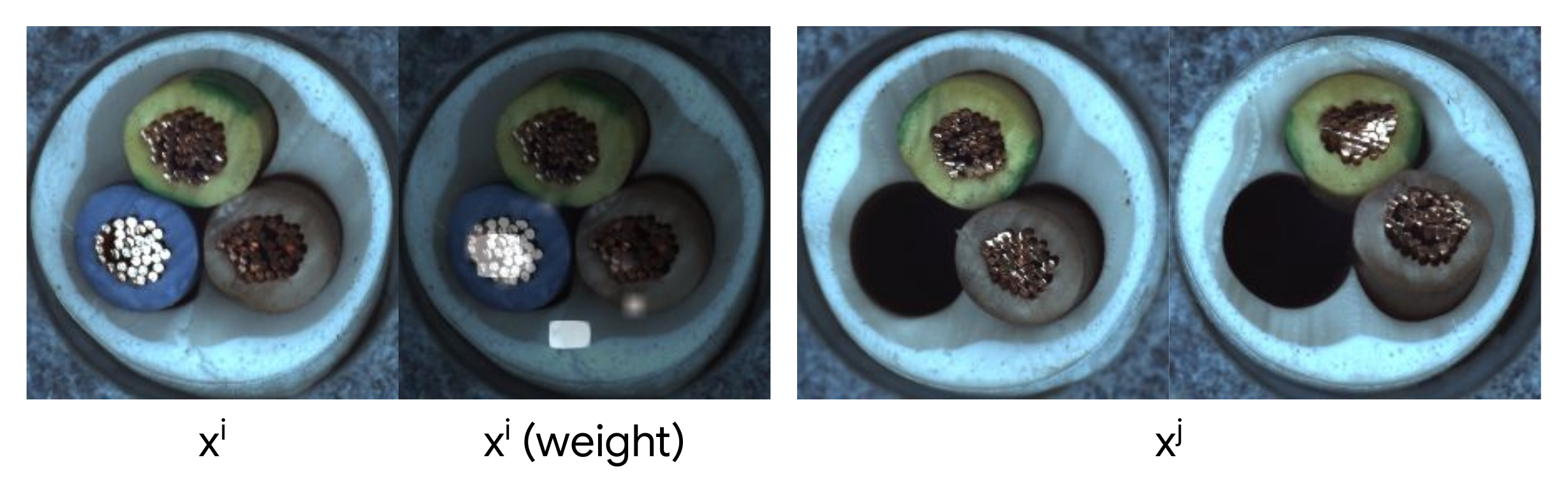}
    \caption{An input image $\mathbf{x}^{i}$ and that with weight overlaid when computed via $\max$ operators against $\mathbf{x}^{j}$'s on the right.}
    \label{fig:issue_max_op}
\end{figure}

Secondly, as in Figure~\ref{fig:issue_max_op}, the $\max$ operator would highlight the blue cable as it does not exists for some images in the dataset, even though it is a normal pattern. 

\section{Formulations for Variant of Hausdorff Distance}
\label{sec:app_hausdorff_formulations}

In this section we provide exact formulations that we use for experiments in Section~\ref{sec:ablation_distance_measure}.

\begin{enumerate}
    \item Eq.~\eqref{eq:distance_hausdorff_single}: $\mathrm{mean}$ $\mathrm{mean}$, Eq.~\eqref{eq:distance_hausdorff}: -- :
    \begin{gather*}
        d_{\mathrm{avgavg}}(Z_{i},Z_{j}) = \frac{1}{MN}\sum_{m=1,...,M}\sum_{n=1,...,M} \big\{\Vert\mathbf{z}_{m}^{i} - \mathbf{z}_{n}^{j}\Vert\big\}
    \end{gather*}
    \item Eq.~\eqref{eq:distance_hausdorff_single}: $\max$ $\min$, Eq.~\eqref{eq:distance_hausdorff}: $\max$ :
    \begin{gather*}
        d_{\mathrm{maxH}}(Z_{i},Z_{j}) = \max\big\{d(Z_{i},Z_{j}),d(Z_{j},Z_{i})\big\},\\
        d(Z_{i},Z_{j}) = \max_{m=1,...,M}\min_{n=1,...,M} \big\{\Vert\mathbf{z}_{m}^{i} - \mathbf{z}_{n}^{j}\Vert\big\}
    \end{gather*}
    \item Eq.~\eqref{eq:distance_hausdorff_single}: $\max$ $\min$, Eq.~\eqref{eq:distance_hausdorff}: $\mathrm{mean}$ :
    \begin{gather*}
        d_{\mathrm{maxH-avg}}(Z_{i},Z_{j}) = \frac{1}{2}\big(d(Z_{i},Z_{j}) + d(Z_{j},Z_{i})\big),\\
        d(Z_{i},Z_{j}) = \max_{m=1,...,M}\min_{n=1,...,M} \big\{\Vert\mathbf{z}_{m}^{i} - \mathbf{z}_{n}^{j}\Vert\big\}
    \end{gather*}
    \item Eq.~\eqref{eq:distance_hausdorff_single}: $\min$ $\min$, Eq.~\eqref{eq:distance_hausdorff}: -- :
    \begin{gather*}
        d_{\mathrm{minmin}}(Z_{i},Z_{j}) = \min_{m=1,...,M}\min_{n=1,...,M} \big\{\Vert\mathbf{z}_{m}^{i} - \mathbf{z}_{n}^{j}\Vert\big\}
    \end{gather*}
    \item Eq.~\eqref{eq:distance_hausdorff_single}: $\mathrm{mean}$ $\min$, Eq.~\eqref{eq:distance_hausdorff}: $\max$ :
    \begin{gather*}
        d_{\mathrm{avgmin}}(Z_{i},Z_{j}) = \max\big\{d(Z_{i},Z_{j}),d(Z_{j},Z_{i})\big\},\\
        d(Z_{i},Z_{j}) = \frac{1}{M}\sum_{m=1,...,M}\min_{n=1,...,M} \big\{\Vert\mathbf{z}_{m}^{i} - \mathbf{z}_{n}^{j}\Vert\big\}
    \end{gather*}
    \item Eq.~\eqref{eq:distance_hausdorff_single}: $\mathrm{mean}$ $\min$, Eq.~\eqref{eq:distance_hausdorff}: $\mathrm{mean}$ :
    \begin{gather*}
        d_{\mathrm{avgmin}}(Z_{i},Z_{j}) = \frac{1}{2}\big(d(Z_{i},Z_{j}) + d(Z_{j},Z_{i})\big),\\
        d(Z_{i},Z_{j}) = \frac{1}{M}\sum_{m=1,...,M}\min_{n=1,...,M} \big\{\Vert\mathbf{z}_{m}^{i} - \mathbf{z}_{n}^{j}\Vert\big\}
    \end{gather*}
\end{enumerate}

\subsection{Implementation Details for Deep Clustering}
\label{sec:app_deep_cluster_detail}

We follow general guidelines provided by the authors of IIC~\cite{ji2019invariant},\footnote{\url{https://github.com/xu-ji/IIC}} GATCluster~\cite{niu2020gatcluster},\footnote{\url{https://github.com/niuchuangnn/GATCluster}} and SCAN~\cite{van2020scan},\footnote{\url{https://github.com/wvangansbeke/Unsupervised-Classification}} for experiments with deep clustering methods.
For IIC and SCAN, we use a ResNet-50 backbone. We replace the first step of the SCAN, which is the self-supervised pretraining, with an ImageNet pretraining as the number of images for each dataset we consider in the paper is relatively small (e.g., 100$\sim$1000, as opposed to 50k for CIFAR-10 or more than a million for ImageNet). For GATCluster, we use the custom CNN architecture suggested by the author for ImageNet experiments.

For hyperparameters, we simply use the ones suggested by the authors. While these hyperparameters may not be optimal for anomaly detection datasets, we believe this is fair treatment as we do not conduct serious hyperparameter tuning for our methods.

\input{table/table_07.tex}

\input{table/table_08.tex}

\include{table/table_01_supp}

\include{table/table_03_supp}

\input{table/figure_09.tex}

\end{document}

%% file: table/figure_03.tex
\begin{figure*}[t]
    \centering
    \begin{subfigure}{0.3\linewidth}
        \centering
        \includegraphics[width=0.98\linewidth]{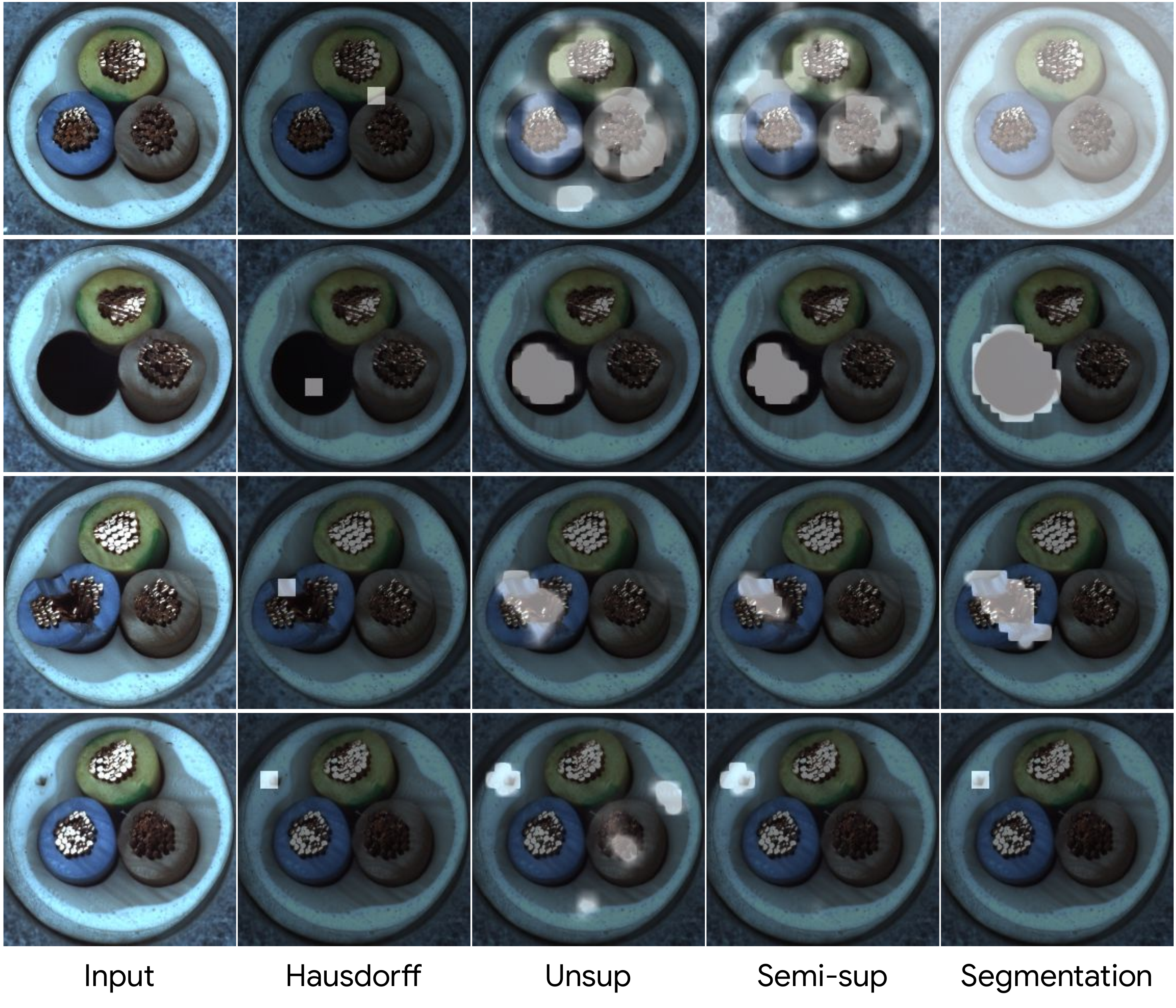}
        \caption{MVTec (Cable)}\label{fig:attention_cable}
    \end{subfigure}
    \hspace{0.05in}
    \begin{subfigure}{0.3\linewidth}
        \centering
        \includegraphics[width=0.98\linewidth]{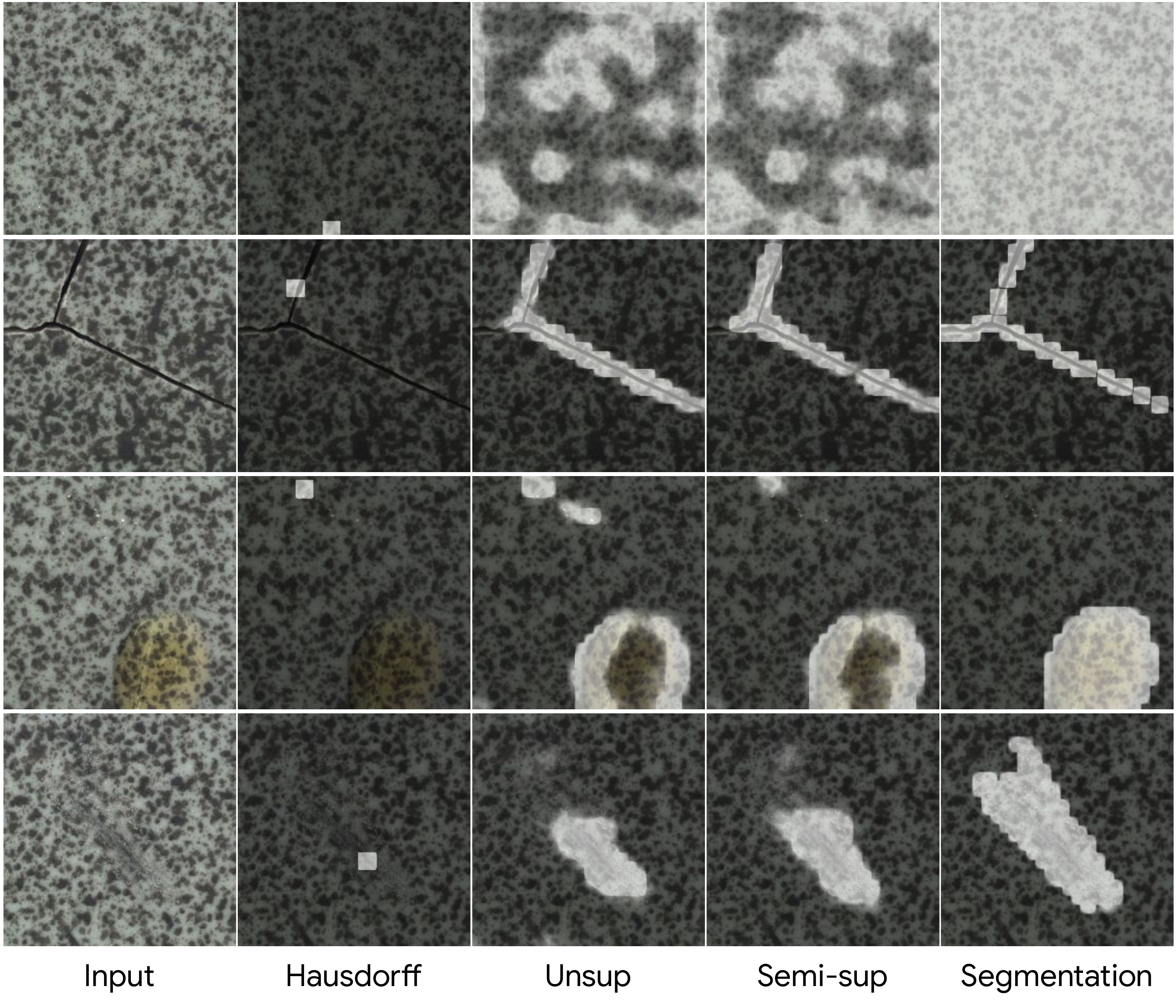}
        \caption{MVTec (Tile)}\label{fig:attention_tile}
    \end{subfigure}
    \hspace{0.05in}
    \begin{subfigure}{0.3\linewidth}
        \centering
        \includegraphics[width=0.98\linewidth]{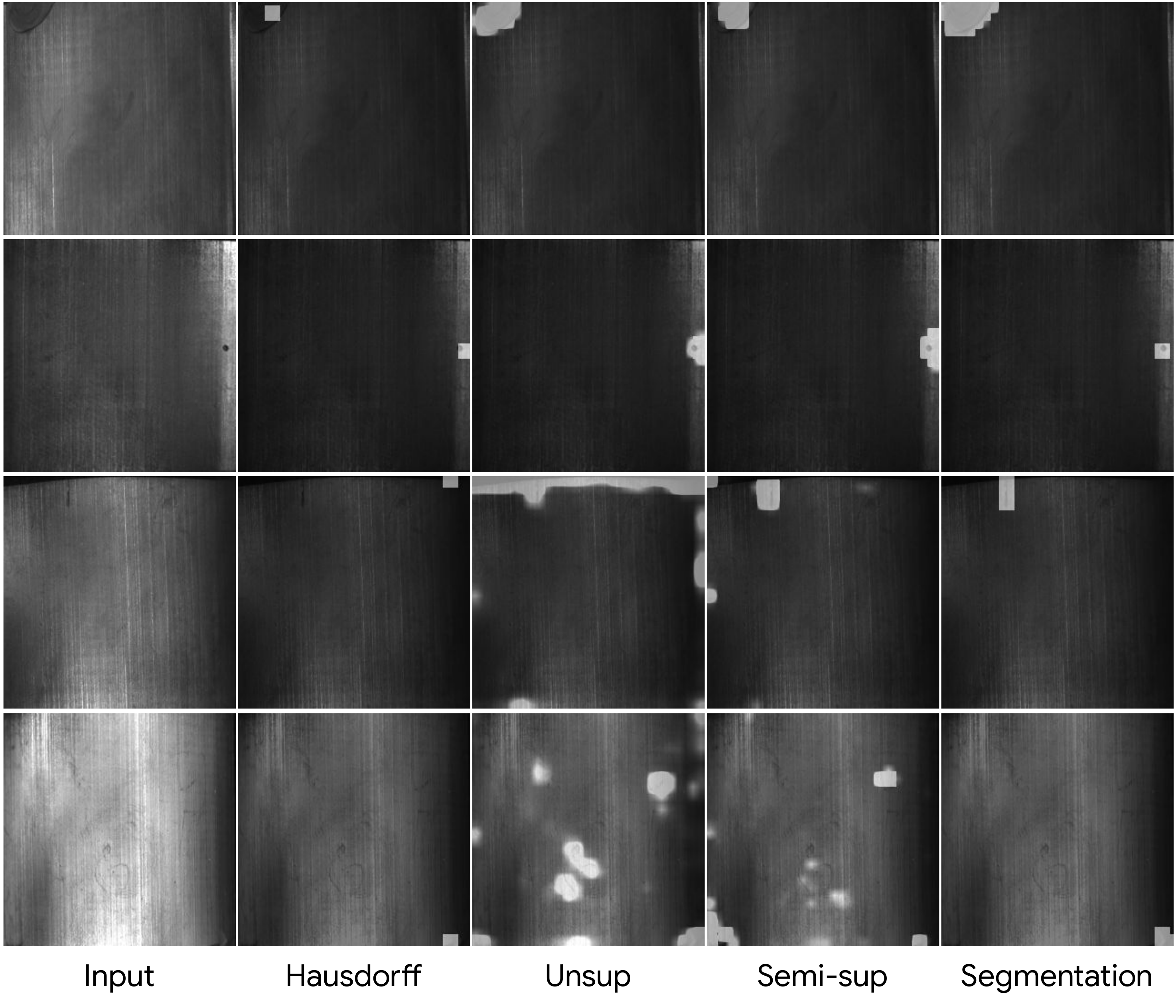}
        \caption{Magnetic Tile}\label{fig:attention_mtd}
    \end{subfigure}
    \vspace{-0.1in}
    \caption{Visualization of weights (${\alpha}$) overlaid on images. From the second column, we show instances (e.g., patches) chosen to compute maximum Hausdorff distances in Equation~\eqref{eq:weighted_distance_to_maxh}, unsupervised in Equation~\eqref{eq:weighted_distance_unsup}, semi-supervised in Equation~\eqref{eq:weighted_distance_semisup}, and ground-truth segmentation label-based weights.}
    \label{fig:attention}
    \vspace{-0.2in}
\end{figure*}

%% file: table/table_01.tex
\begin{table*}[t]
    \centering
    \caption{NMI, ARI and F1 scores of unsupervised and semi-supervised clustering methods on MVTec (object, texture) and MTD datasets. Compared to the baseline method (``average'') that uses a holistic representation via average pooling of patch embeddings, the multiple instance clustering framework with various distance measures, such as maximum Hausdorff or the proposed weighted average distances, show huge improvement. We also report the performance of weighted average distance whose weights are computed using labeled normal data (``Semi''). 
    %
    %Various distance measures, such as average, maximum Hausdorff, or the proposed weighted average, are evaluated with the hierarchical Ward clustering. We also report the performance of weighted average distance whose weights are generated from the ground-truth segmentation masks. 
    We provide per-category results in Table~\ref{tab:main_mvtec_per_cat} of Appendix.
    }
    \label{tab:main_mvtec}
    \vspace{0.05in}
    \resizebox{0.9\textwidth}{!}{%
    \begin{tabular}{c||c|c|c|c|c|c|c|c|c||c|c|c}
        \toprule
        Supervision & \multicolumn{9}{c||}{Unsupervised} & \multicolumn{3}{c}{Semi} \\
        \midrule
        Distance & \multicolumn{3}{c|}{Average} & \multicolumn{3}{c|}{{\,\,}Maximum Hausdorff{\,\,}} & \multicolumn{3}{c||}{{\,\,}Weighted Average{\,\,}} & \multicolumn{3}{c}{{\,\,}Weighted Average{\,\,}} \\
        \midrule
        Metric & {\,\,}NMI{\,\,} & {\,\,}ARI{\,\,} & {\,\,}F1{\,\,} & {\,\,}NMI{\,\,} & {\,\,}ARI{\,\,} & {\,\,}F1{\,\,} & {\,\,}NMI{\,\,} & {\,\,}ARI{\,\,} & {\,\,}F1{\,\,} & {\,\,}NMI{\,\,} & {\,\,}ARI{\,\,} & {\,\,}F1{\,\,} \\
        \midrule
        %MVTec (object) & 0.244 & 0.109 & 0.399 & 0.415 & 0.275 & 0.526 & 0.451 & 0.346 & 0.553 & \textbf{0.585} & 0.460 & 0.635 & \textbf{\color{red}0.724} & 0.652 & 0.725 \\
        %MVTec (texture) & 0.273 & 0.123 & 0.402 & 0.625 & 0.534 & 0.708 & 0.649 & 0.570 & 0.689 & \textbf{0.659} & 0.577 & 0.704 & \textbf{\color{red}0.685} & 0.609 & 0.703 \\
        %MTD & 0.065 & 0.024 & 0.289 & 0.193 & 0.112 & 0.381 & 0.179 & 0.120 & 0.346 & \textbf{0.390} & 0.314 & 0.490 & \textbf{\color{red}0.467} & 0.359 & 0.482 \\
        %MVTec (object) & 0.244 & 0.109 & 0.399 & 0.415 & 0.275 & 0.526 & 0.451 & 0.346 & 0.553 & \textbf{0.585} & 0.460 & 0.635 & \textbf{\color{red}0.724} & 0.652 & 0.725 \\
        %MVTec (texture) & 0.273 & 0.123 & 0.402 & 0.625 & 0.534 & 0.708 & 0.674 & 0.601 & 0.707 & \textbf{0.675} & 0.575 & 0.702 & \textbf{\color{red}0.685} & 0.609 & 0.703 \\
        %MTD & 0.065 & 0.024 & 0.289 & 0.193 & 0.112 & 0.381 & 0.179 & 0.120 & 0.346 & \textbf{0.390} & 0.314 & 0.490 & \textbf{\color{red}0.467} & 0.359 & 0.482 \\
        {\,\,}MVTec (object){\,\,} & 0.244 & 0.109 & 0.399 & 0.415 & 0.275 & 0.526 & 0.451 & 0.346 & 0.553 & \textbf{0.577} & 0.449 & 0.628 \\
        {\,\,}MVTec (texture){\,\,} & 0.273 & 0.123 & 0.402 & 0.625 & 0.534 & 0.708 & \textbf{0.674} & 0.601 & 0.707 & 0.669 & 0.570 & 0.698 \\
        MTD & 0.065 & 0.024 & 0.289 & 0.193 & 0.112 & 0.381 & 0.179 & 0.120 & 0.346 & \textbf{0.390} & 0.314 & 0.490 \\
        \midrule
        Overall & 0.251 & 0.112 & 0.394 & 0.532 & 0.427 & 0.631 & 0.573 & 0.491 & 0.636 & \textbf{0.623} & 0.516 & 0.663 \\ 
        \bottomrule
    \end{tabular}
    }
    %\vspace{-0.15in}
\end{table*}

%% file: table/table_03.tex
\iftrue
% DONE / DONE
\begin{table}[t]
    \centering
    \caption{Comparison to other clustering methods, including KMeans, KMedoids, GMM, spectral, and hierarchical clustering with various linkages, using maximum Hausdorff (maxH) or weighted average (WA) distances, and deep clustering methods, such as IIC~\cite{ji2019invariant}, GATCluster~\cite{niu2020gatcluster}, or SCAN~\cite{van2020scan}. For deep clustering methods, we provide in the parenthesis the performance of the best training epoch chosen by test set accuracy. We report NMIs, and complete results are in the Table~\ref{tab:supp_clusterer_comparison} of Appendix.}
    \label{tab:main_clusterer_comparison}
    \vspace{0.05in}
    \resizebox{0.96\linewidth}{!}{%
    \begin{tabular}{l|c|c|c|c|c|c}
        \toprule
        Dataset & \multicolumn{2}{c|}{{\,\,\,}MVTec (object){\,\,\,}} & \multicolumn{2}{c|}{{\,\,\,}MVTec (texture){\,\,\,}} & \multicolumn{2}{c}{MTD} \\ \cmidrule{1-7}
        Distance & {  }maxH{  } & WA & {  }maxH{  } & WA & {  }maxH{  } & {\,\,\,\,\,\,}WA{\,\,\,\,\,\,} \\
        \midrule
        %KMeans & --   & 0.429 & --   & 0.622 & --  & \textbf{0.204} \\
        %GMM & --   & 0.395 & --   & 0.578 & --  & \textbf{0.204} \\
        %Spectral & 0.419 & 0.428 & 0.609 & 0.573 & 0.143 & 0.150 \\
        %Single & 0.108 & 0.129 & 0.078 & 0.107 & 0.087 & 0.067 \\
        %Complete & 0.316 & 0.311 & 0.360 & 0.444 & 0.128 & 0.118 \\
        %Average & 0.245 & 0.238 & 0.223 & 0.380 & 0.080 & 0.094 \\
        %Ward & 0.415 & \textbf{0.451} & 0.625 & \textbf{0.649} & 0.193 & 0.179 \\
        %KMeans & --   & 0.429 & --   & 0.642 & --  & \textbf{0.204} \\
        %GMM & --   & 0.395 & --   & 0.578 & --  & \textbf{0.204} \\
        %Spectral & 0.419 & 0.428 & 0.609 & 0.606 & 0.143 & 0.150 \\
        %Single & 0.108 & 0.130 & 0.078 & 0.098 & 0.087 & 0.067 \\
        %Complete & 0.316 & 0.307 & 0.360 & 0.511 & 0.128 & 0.118 \\
        %Average & 0.245 & 0.276 & 0.223 & 0.380 & 0.080 & 0.094 \\
        %Ward & 0.415 & \textbf{0.451} & 0.625 & \textbf{0.674} & 0.193 & 0.179 \\
        KMeans & --   & 0.429 & --   & 0.642 & --  & \textbf{0.204} \\
        GMM & --   & 0.395 & --   & 0.578 & --  & \textbf{0.204} \\
        KMedoids & 0.140 & 0.235 & 0.274 & 0.430 & 0.050 & 0.076 \\
        Spectral & 0.419 & 0.428 & 0.609 & 0.606 & 0.143 & 0.150 \\
        Single & 0.108 & 0.133 & 0.078 & 0.108 & 0.087 & 0.065 \\
        Complete & 0.316 & 0.294 & 0.360 & 0.452 & 0.128 & 0.116 \\
        Average & 0.245 & 0.276 & 0.223 & 0.400 & 0.080 & 0.094 \\
        Ward & 0.415 & \textbf{0.451} & 0.625 & \textbf{0.674} & 0.193 & 0.179 \\
        \midrule
        {IIC} & \multicolumn{2}{c|}{0.086 (0.170)} & \multicolumn{2}{c|}{0.107 (0.188)} & \multicolumn{2}{c}{0.064 (0.034)} \\
        {GATCluster}{  } & \multicolumn{2}{c|}{0.119 (0.265)} & \multicolumn{2}{c|}{0.171 (0.298)} & \multicolumn{2}{c}{0.028 (0.113)} \\
        {SCAN} & \multicolumn{2}{c|}{0.176 (0.198)} & \multicolumn{2}{c|}{0.277 (0.314)} & \multicolumn{2}{c}{0.071 (0.087)} \\
        %{IIC} & 0.086 (0.170) & 0.107 (0.188) & \\
        %{GATCluster} & 0.119 (0.265) & 0.171 (0.298) & \\
        %{SCAN} & 0.176 (0.198) & 0.277 (0.314) & \\
        \bottomrule
    \end{tabular}
    }
    \vspace{-0.15in}
\end{table}
\fi

%% file: table/table_02.tex
\begin{table}[ht]
    \centering
    \resizebox{0.95\linewidth}{!}{%
    \begin{tabular}{c|c|c|c|c|c}
        \toprule
        \multicolumn{2}{c|}{{  }Dataset and Metric{  }} & {  }average{  } & {  }maxH{  } & {  }WA{  } & {  }WA (semi){  } \\ \midrule
        \multirow{4}{*}{object{  }} & mAUC ($\uparrow$) & 0.819 & 0.868 & 0.860 & \textbf{0.915} \\
        & R@0.9 ($\uparrow$) & 0.380 & 0.533 & 0.474 & \textbf{0.671} \\
        & R@0.95 ($\uparrow$) & 0.231 & 0.373 & 0.346 & \textbf{0.527} \\
        & R@0.99 ($\uparrow$) & 0.094 & 0.204 & 0.192 & \textbf{0.327} \\
        \midrule
        \multirow{4}{*}{texture{  }} & mAUC ($\uparrow$) & 0.807 & 0.926 & 0.907 & \textbf{0.940} \\
        & R@0.9 ($\uparrow$) & 0.378 & 0.769 & 0.760 & \textbf{0.824} \\
        & R@0.95 ($\uparrow$) & 0.243 & \textbf{0.702} & 0.629 & 0.666 \\
        & R@0.99 ($\uparrow$) & 0.083 & {0.346} & \textbf{0.393} & 0.366 \\
        \bottomrule
    \end{tabular}
    }
    \caption{Cluster purity in mAUC, R@0.9, 0.95, 0.99 on MVTec object and texture categories with various distances.}
    \label{tab:purity}
    \vspace{-0.15in}
\end{table}

%% file: table/table_10.tex
\begin{table}[ht]
    \centering
    \resizebox{0.9\linewidth}{!}{%
    \begin{tabular}{l|c|c|c|c}
        \toprule
         %& \multirow{2}{*}{holistic} & \multicolumn{3}{c}{patch-based} \\
         %\cmidrule{3-5}
        Datasets{\,\,\,\,} & {  }Holistic{  } & {  }Hausdorff{  } & {\,\,\,\,\,\,}WA{\,\,\,\,\,\,} & {  }WA (semi){  } \\
        \midrule
        Object & 0.256 & 0.281 & 0.320 & \textbf{0.381} \\
        Texture & 0.507 & 0.542 & 0.568 & \textbf{0.597} \\
        MTD & 0.205 & 0.250 & 0.227 & \textbf{0.280} \\
        \bottomrule
    \end{tabular}
    }
    \caption{NMI scores of holistic and patch representations.}
    \label{tab:abl_patch_vs_holistic}
    \vspace{-0.15in}
\end{table}

%% file: table/table_05-2.tex
\begin{figure}
    \centering
    \includegraphics[width=0.95\linewidth]{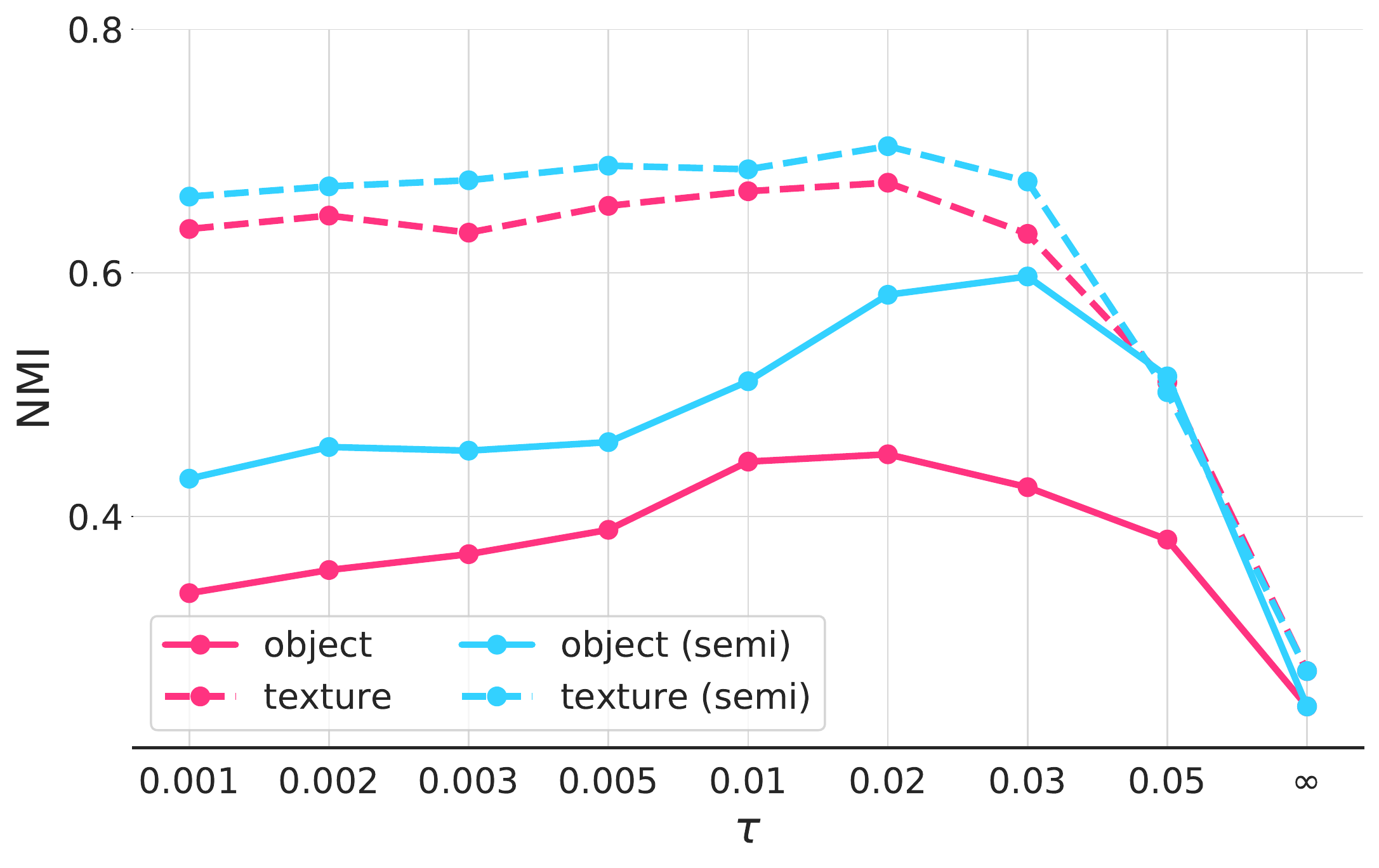}
    \caption{Sensitivity analysis of $\tau$ on MVTec dataset.}
    \label{fig:ablation_tau}
\end{figure}

\begin{table}[t]
    \centering
    \resizebox{0.92\linewidth}{!}{%
    \begin{tabular}{c|c|c|c|c|c}
        \toprule
       {  }Variants{  } & {  }Eq.~\eqref{eq:distance_hausdorff_single}{  } & {  }Eq.~\eqref{eq:distance_hausdorff}{  } & {  }Object{  } & {  }Texture{  } & {  }MTD{  } \\
        \midrule
        & {  }$\mathrm{mean}$ $\mathrm{mean}${  } & -- & 0.196 & 0.232 & 0.071 \\
        & $\max$ $\min$ & $\max$ & \textbf{0.415} & \textbf{0.625} & \textbf{0.193} \\
       {  }Hausdorff{  } & $\max$ $\min$ & $\mathrm{mean}$ & 0.372 & 0.562 & 0.160 \\
       {  }distance{  } & $\min$ $\min$ & -- & 0.126 & 0.187 & 0.130 \\
        & $\mathrm{mean}$ $\min$ & $\max$ & 0.220 & 0.400 & 0.141 \\
        & $\mathrm{mean}$ $\min$ & $\mathrm{mean}$ & 0.235 & 0.348 & 0.134 \\
        \midrule
        \midrule
        Variants & \multicolumn{2}{c|}{Eq.~\eqref{eq:weighted_distance_unsup}} & Object & Texture & MTD \\
        \midrule
        {  }Unsup.{  } & \multicolumn{2}{c|}{$\mathbb{E}_{j\neq i}$} & 0.451 & \textbf{0.674} & \textbf{0.179} \\
        {  }weights{  } &\multicolumn{2}{c|}{$\max_{j\neq i}$} & 0.252 & 0.614 & 0.138 \\
        &\multicolumn{2}{c|}{$\min_{j\neq i}$} & \textbf{0.472} & 0.625 & 0.052 \\
        \bottomrule
    \end{tabular}
    }
    \vspace{-0.03in}
    \caption{NMIs of anomaly clustering using variants of Hausdorff distance and unsupervised weights. 
    %We replace operators in Eq.~\eqref{eq:distance_hausdorff}, \eqref{eq:distance_hausdorff_single} and~\eqref{eq:weighted_distance_unsup} to define variants.
    }
    \label{tab:exp_variants_dist_measure}
    \vspace{-0.15in}
\end{table}

%% file: table/figure_08.tex
\begin{figure*}[t]
    \centering
    \includegraphics[height=1.7in]{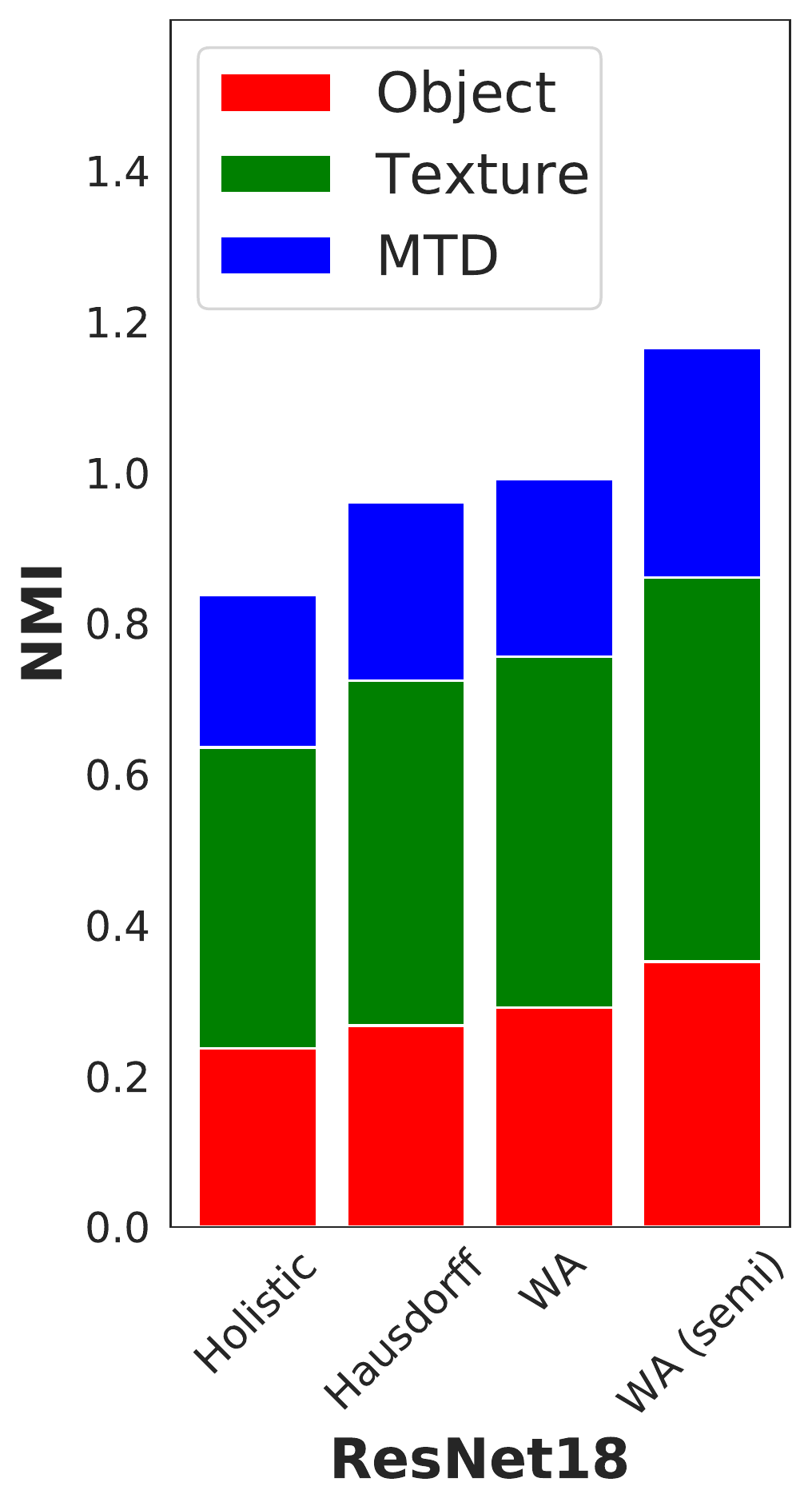}
    \includegraphics[height=1.7in]{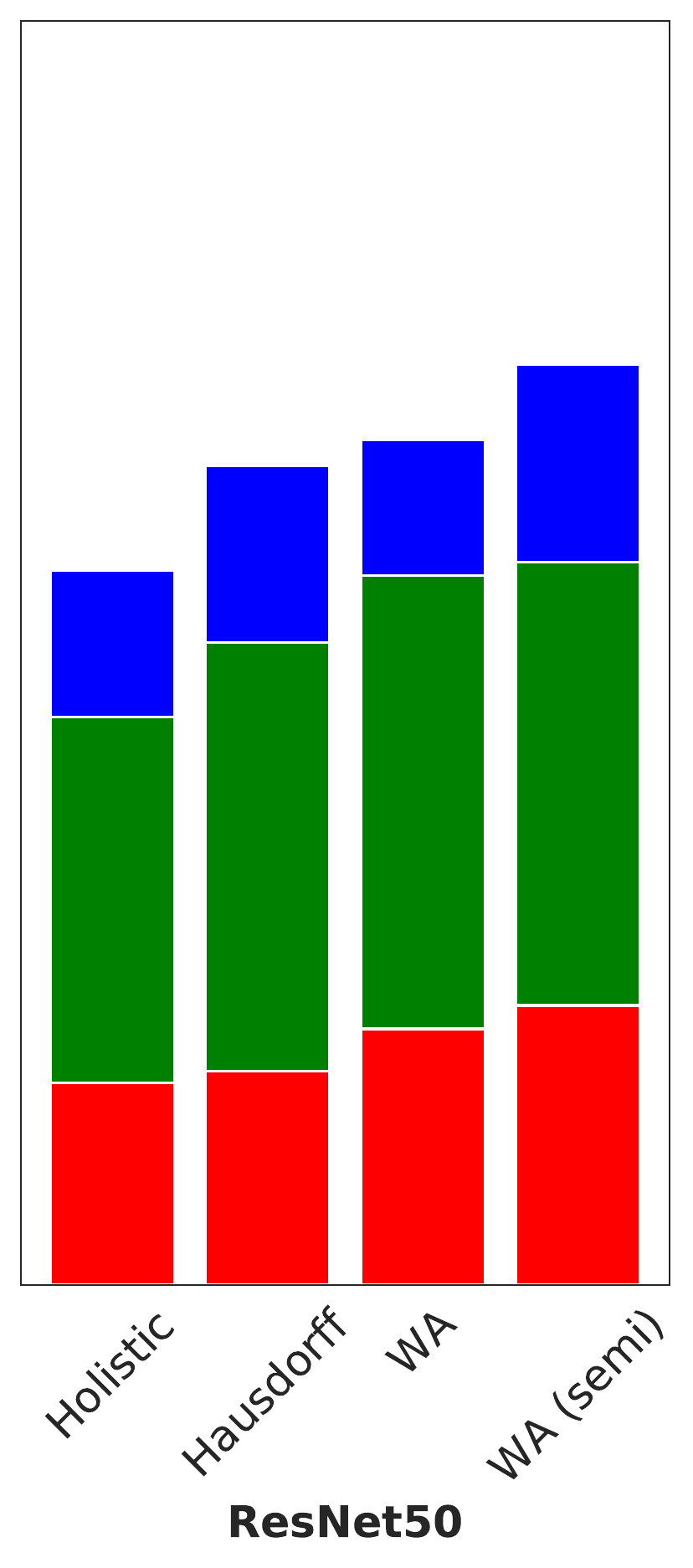}
    \includegraphics[height=1.7in]{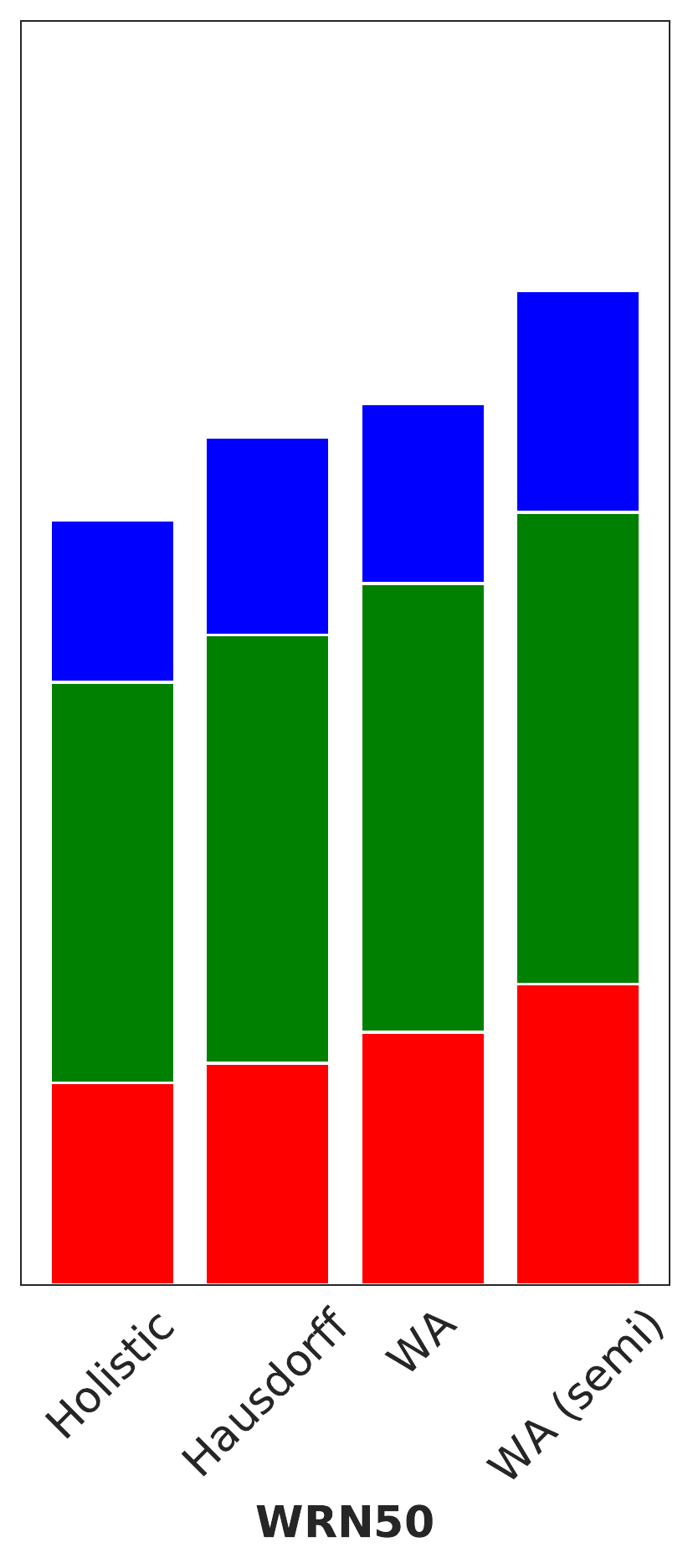}
    \includegraphics[height=1.7in]{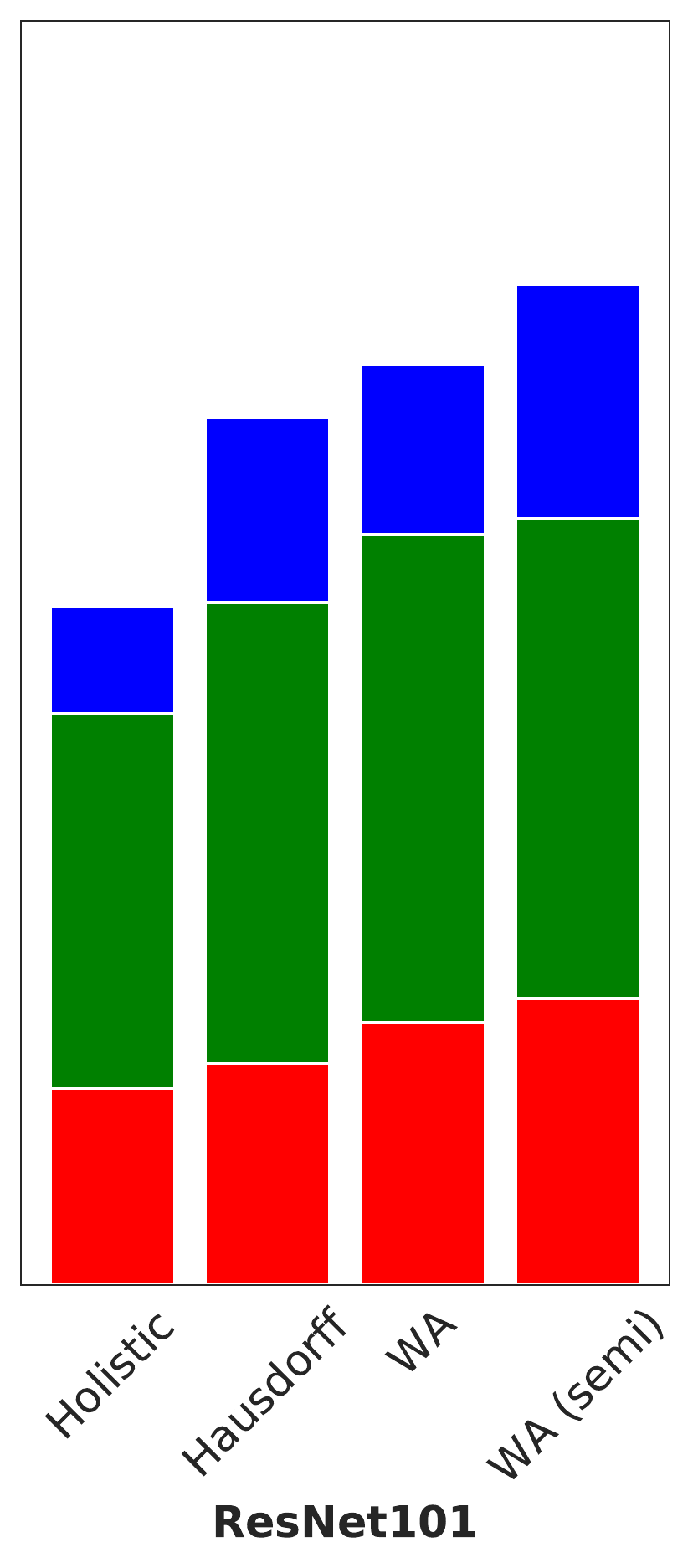}
    \includegraphics[height=1.7in]{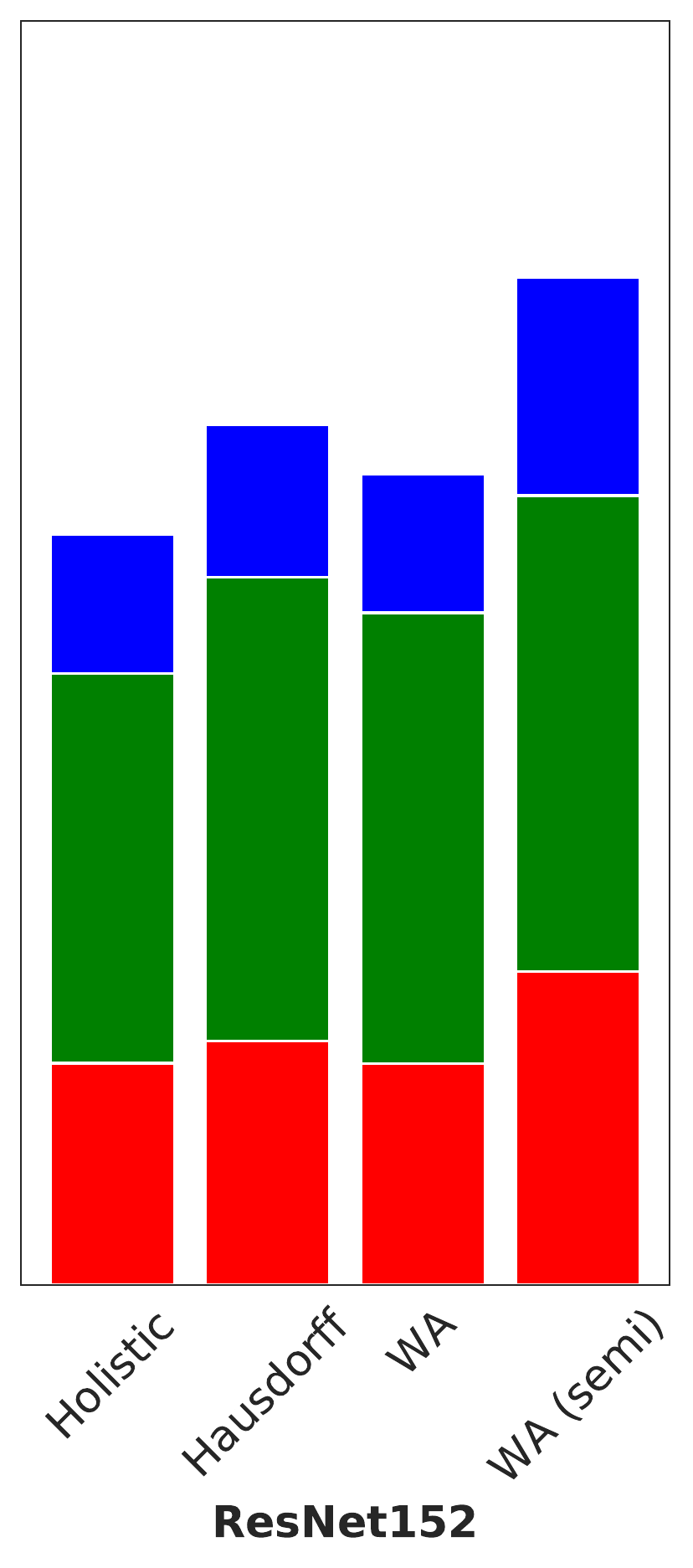}
    \includegraphics[height=1.7in]{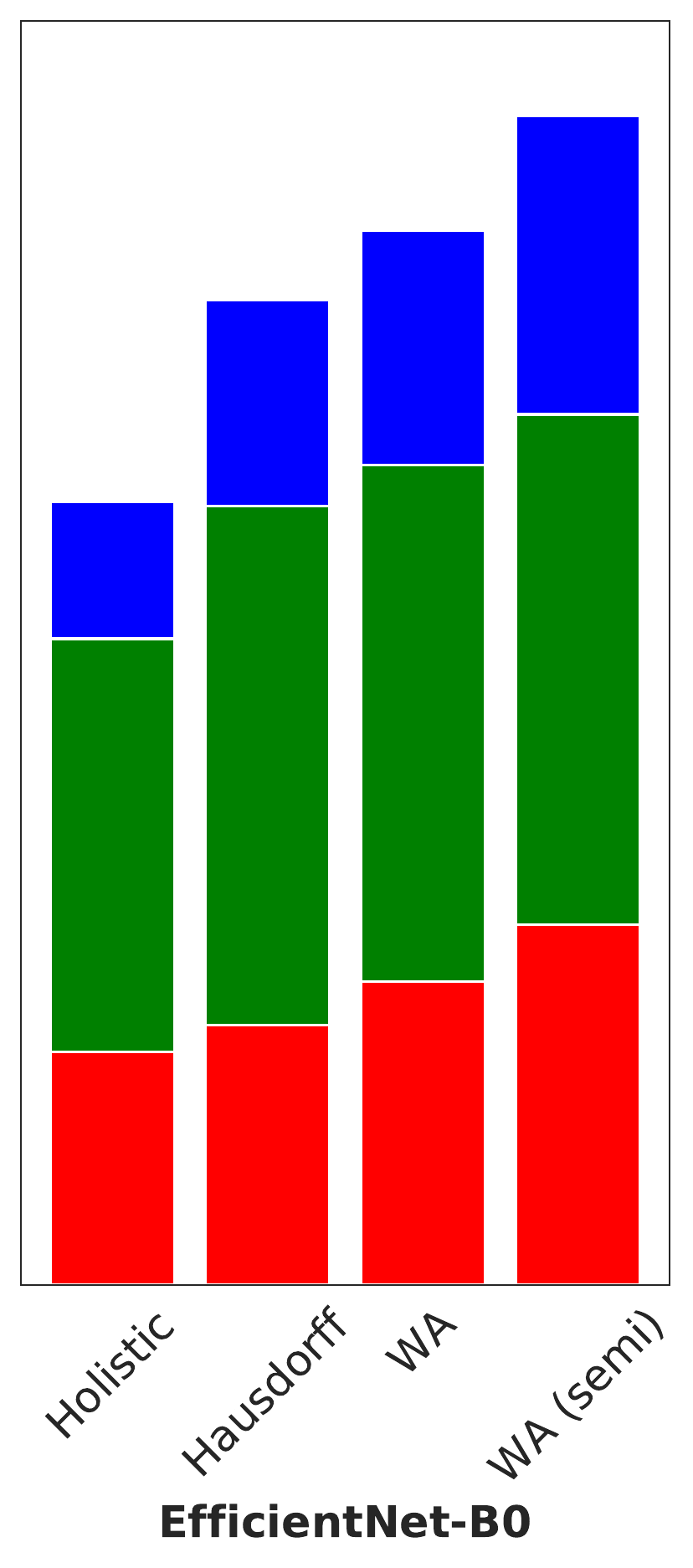}
    \includegraphics[height=1.7in]{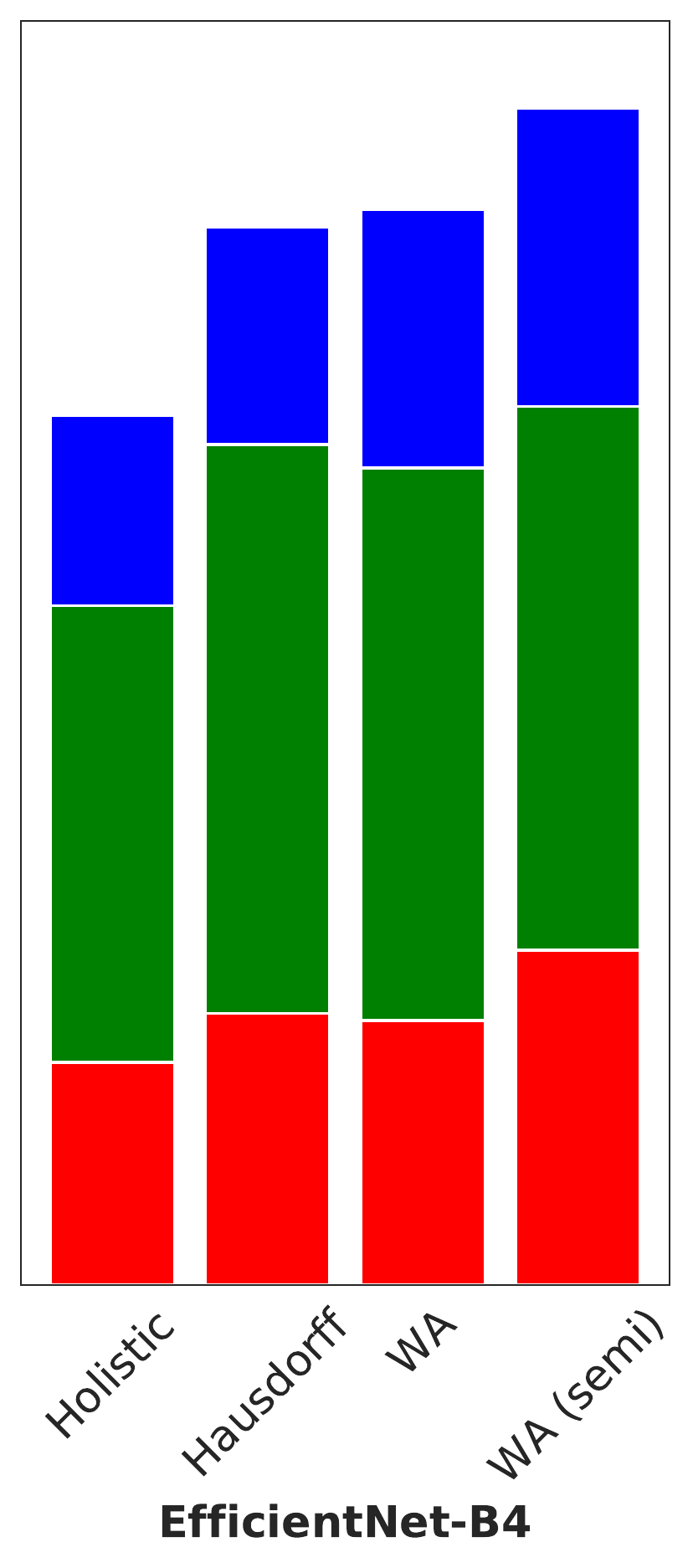}
    \includegraphics[height=1.7in]{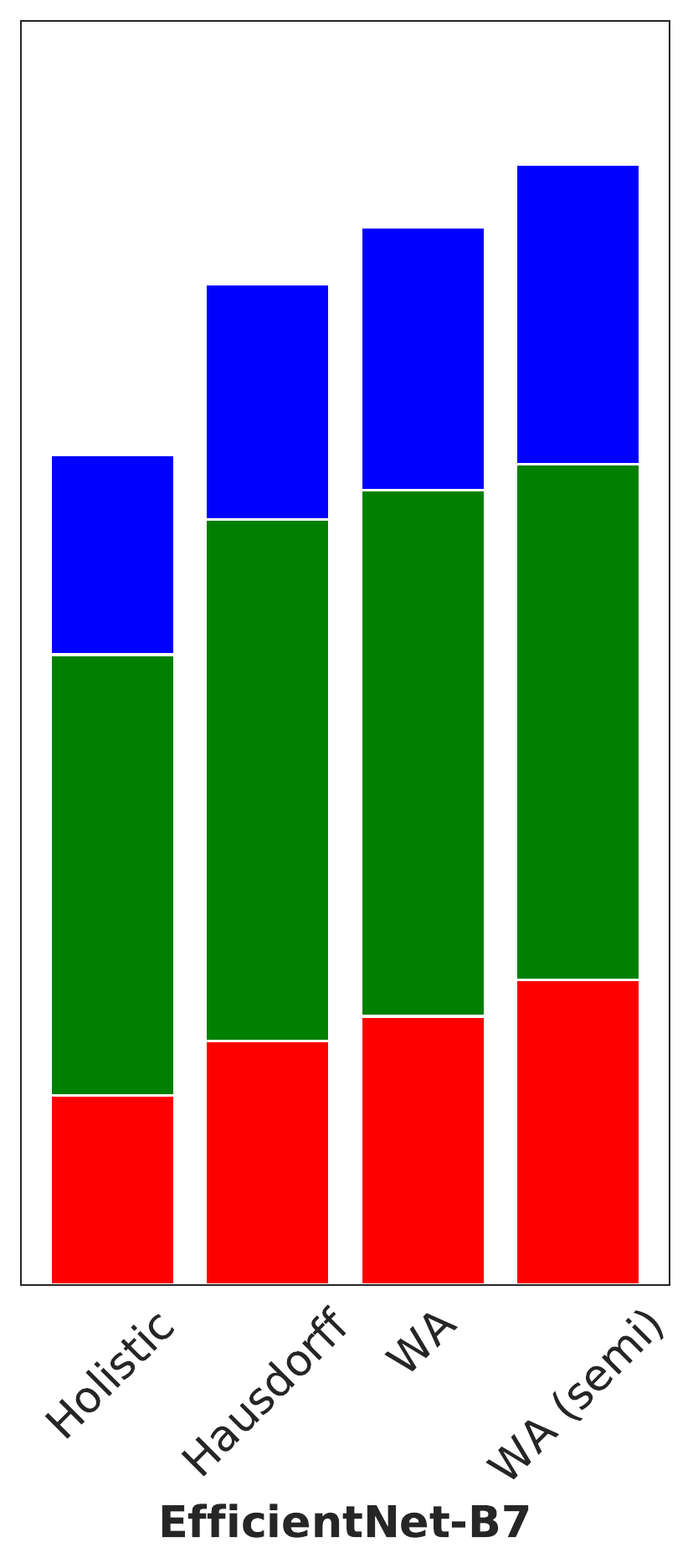}
    \caption{Bar plots with NMI scores over three datasets using various ResNet and EfficientNet models with last hidden layer. We show results for holistic and patch-based with maximum Hausdorff distance, weighted average distance, and its semi-supervised version.}
    \label{fig:app_patch_vs_holistic}
\end{figure*}

%% file: table/figure_06.tex
\iffalse
%\iftrue
\begin{figure*}[t]
    \centering
    \includegraphics[height=1.8in]{fig/nmi-vs-resnet.pdf}
    \includegraphics[height=1.8in]{fig/nmi-vs-effnet.pdf}
    \includegraphics[height=1.8in]{fig/nmi-vs-vit.pdf}
    \caption{NMI scores of various feature extractors on MVTec and MTD datasets. Soft attention with Ward clustering is used.}
    \label{tab:abl_feature_extractor}
\end{figure*}
\fi

% DONE / DONE
\iffalse
\begin{figure*}[t]
    \centering
    \includegraphics[height=1.8in]{fig/nmi-vs-resnet-v2.pdf}
    \includegraphics[height=1.8in]{fig/nmi-vs-effnet-v2.pdf}
    \includegraphics[height=1.8in]{fig/nmi-vs-vit-v2.pdf}
    \caption{NMI scores of various feature extractors on MVTec and MTD datasets. Soft attention with Ward clustering is used.}
    \label{tab:abl_feature_extractor}
\end{figure*}
\fi

\begin{figure*}[t]
    \centering
    \includegraphics[height=2.3in]{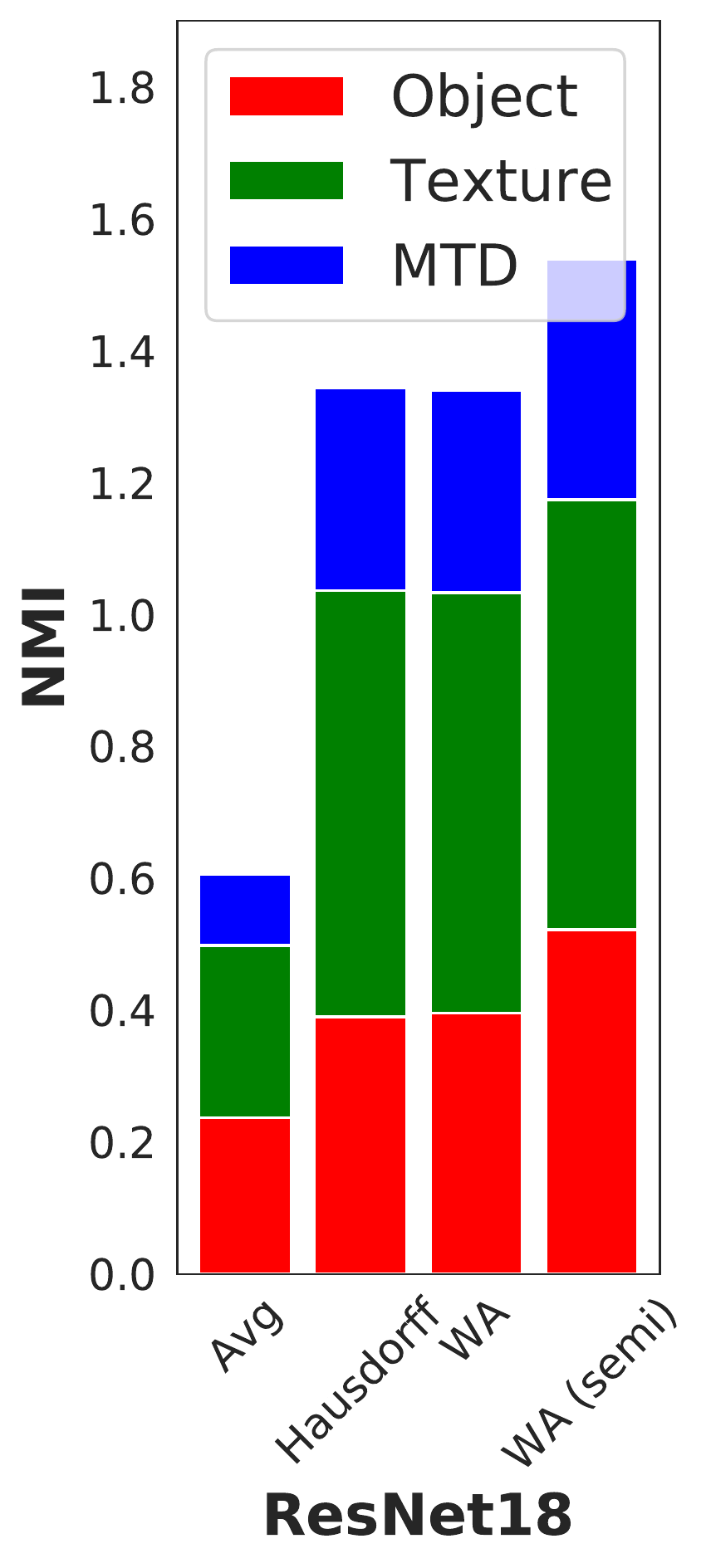}
    \includegraphics[height=2.3in]{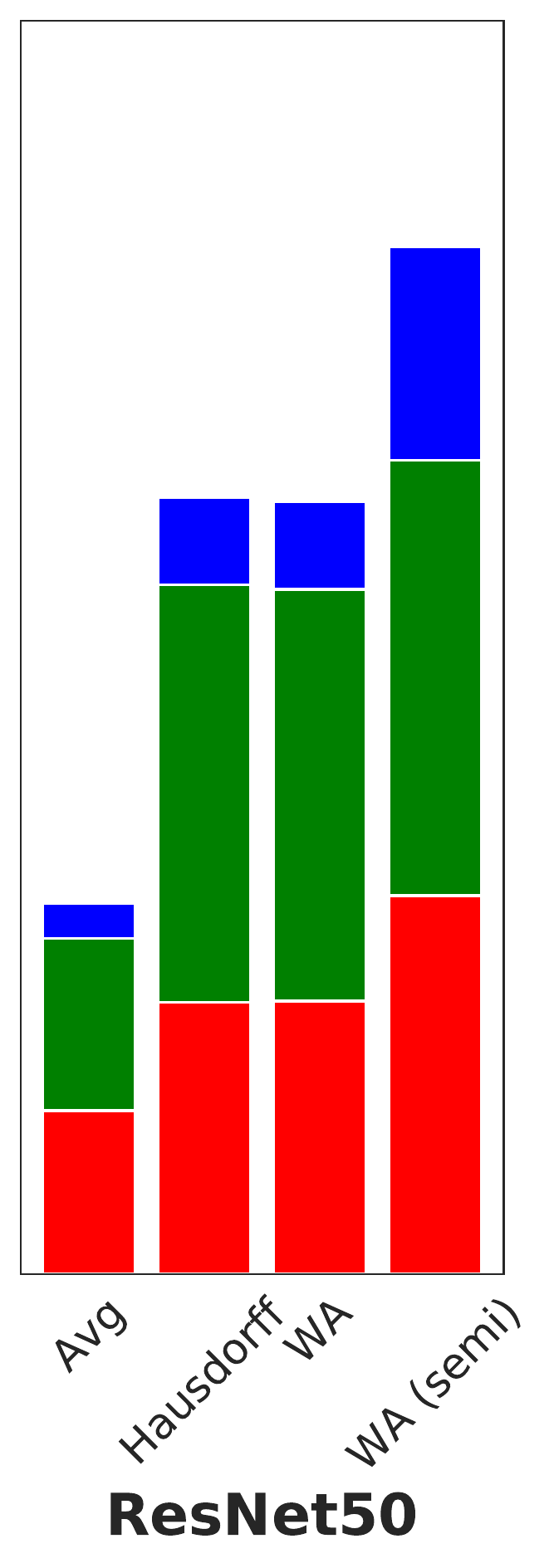}
    \includegraphics[height=2.3in]{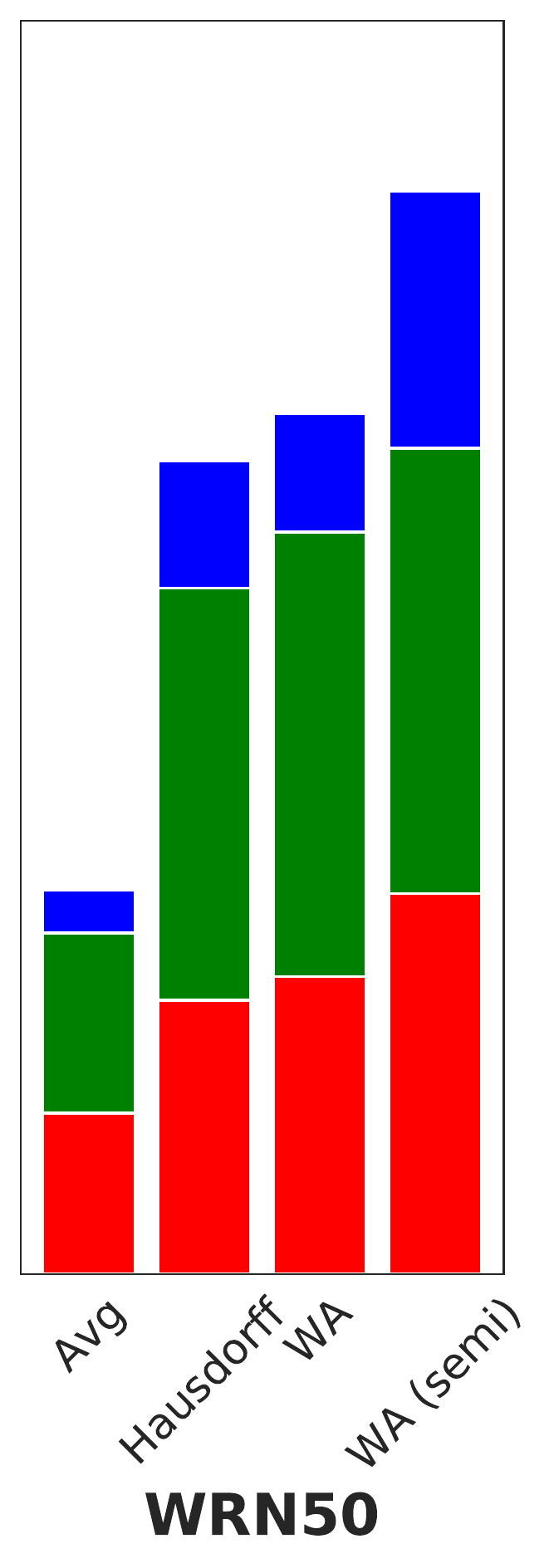}
    \includegraphics[height=2.3in]{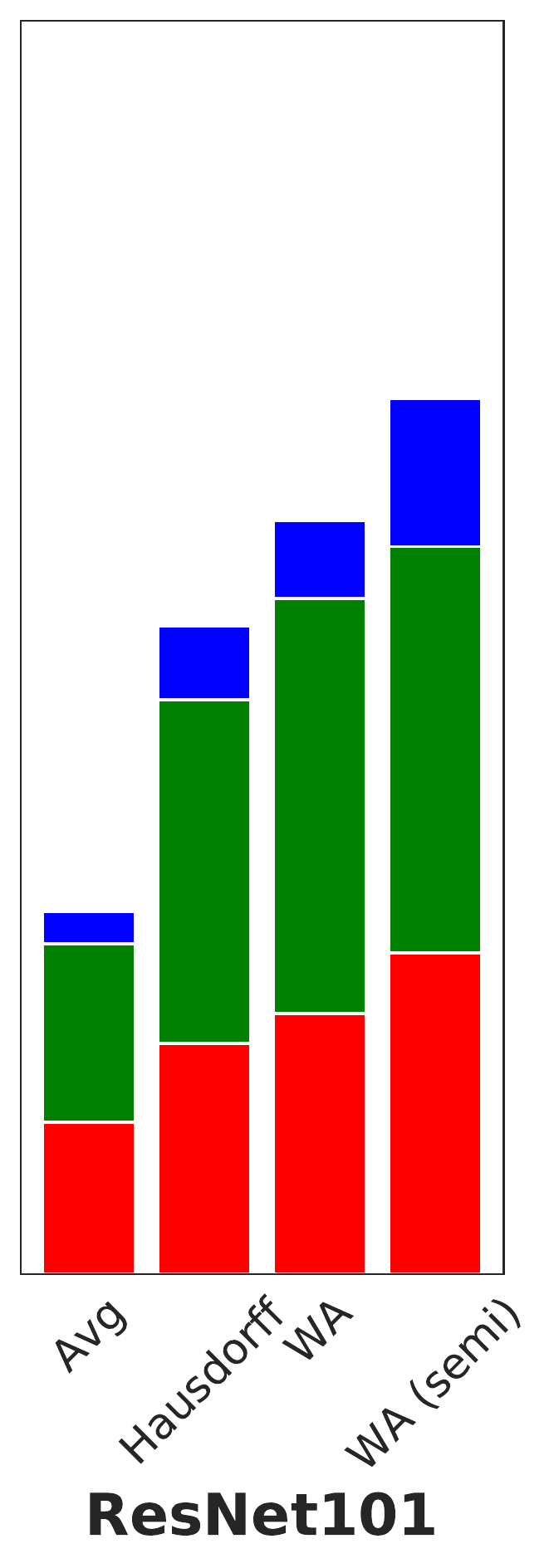}
    \includegraphics[height=2.3in]{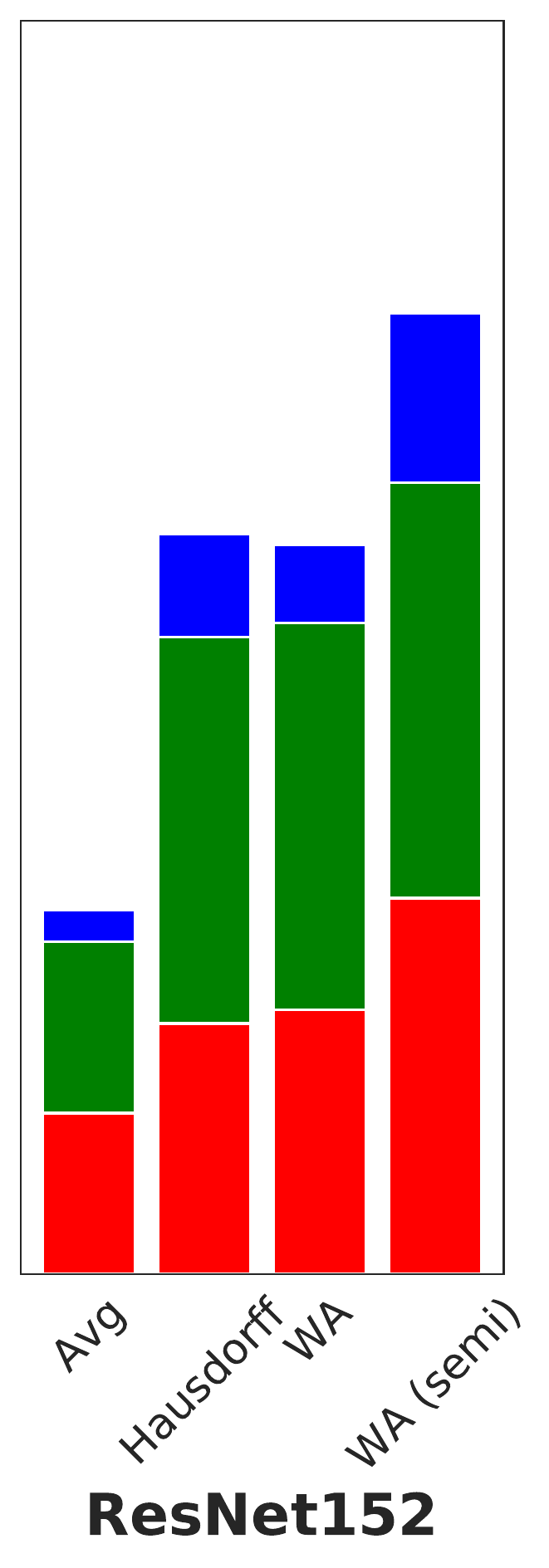}
    \caption{Bar plots with NMI scores over three datasets using various ResNet models with their intermediate layers as in Table~\ref{tab:app_network_config}. We show results for average, maximum Hausdorff distance, weighted average distance, and its semi-supervised version.}
    \label{fig:abl_feature_extractor_resnet}
\end{figure*}
\begin{figure*}[t]
    \centering
    \includegraphics[height=2.3in]{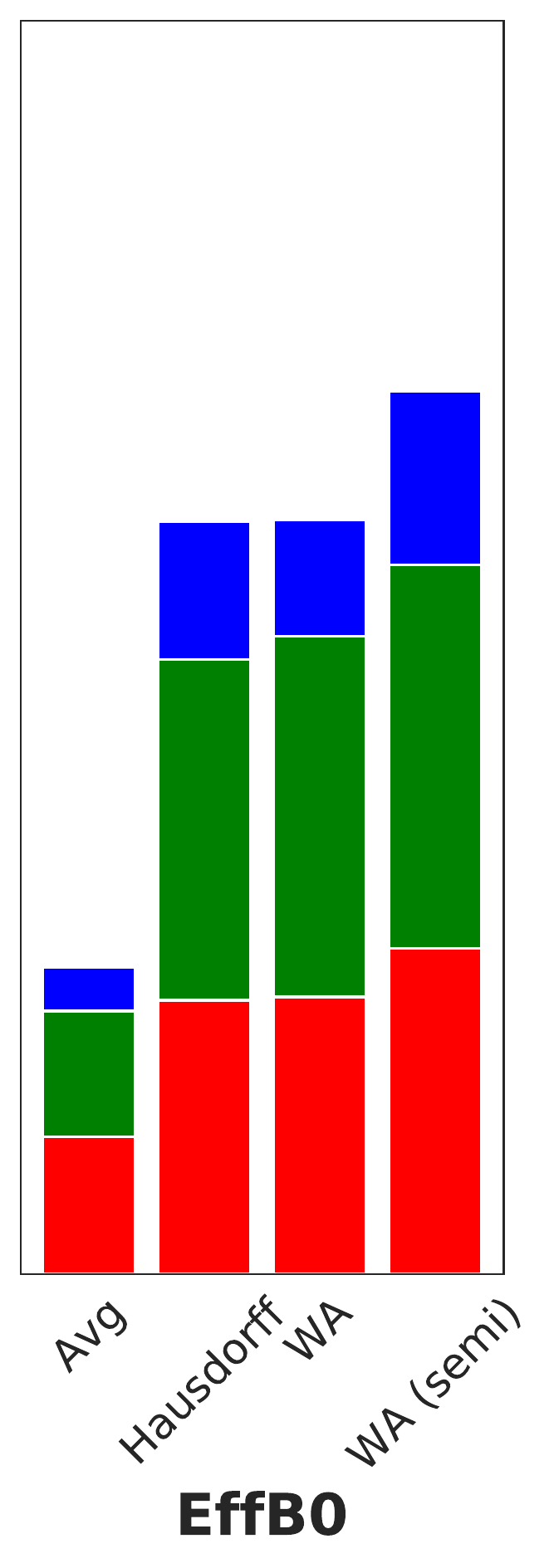}
    \includegraphics[height=2.3in]{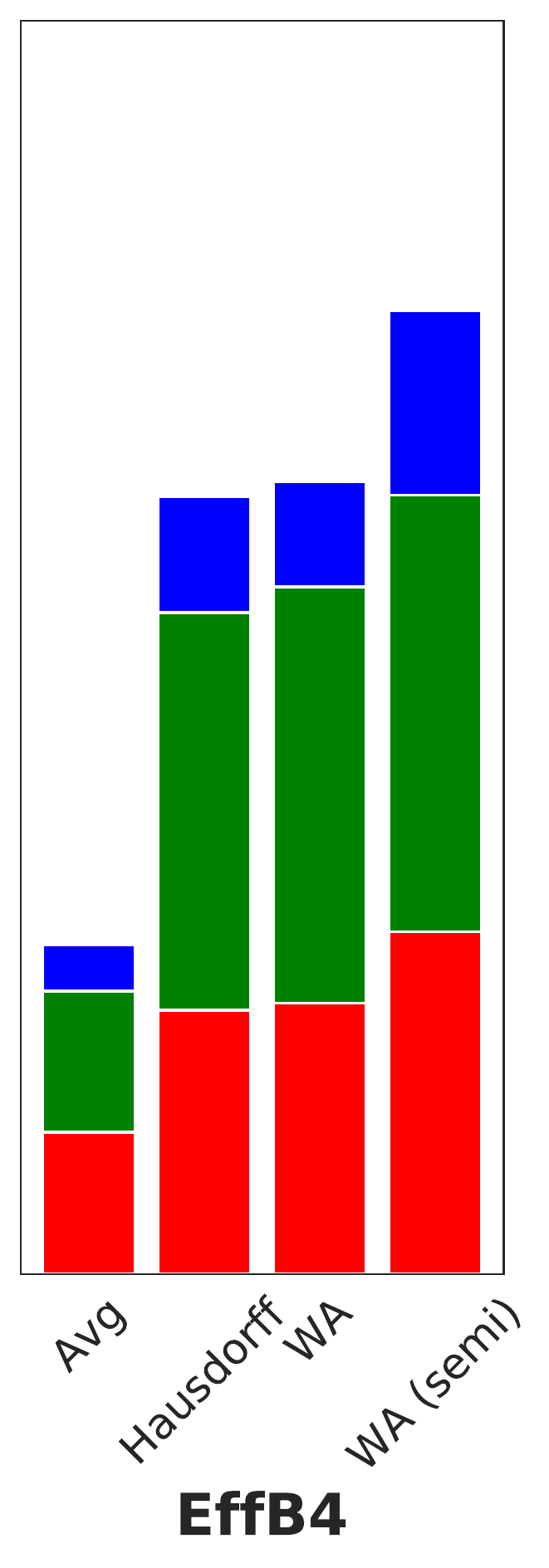}
    \includegraphics[height=2.3in]{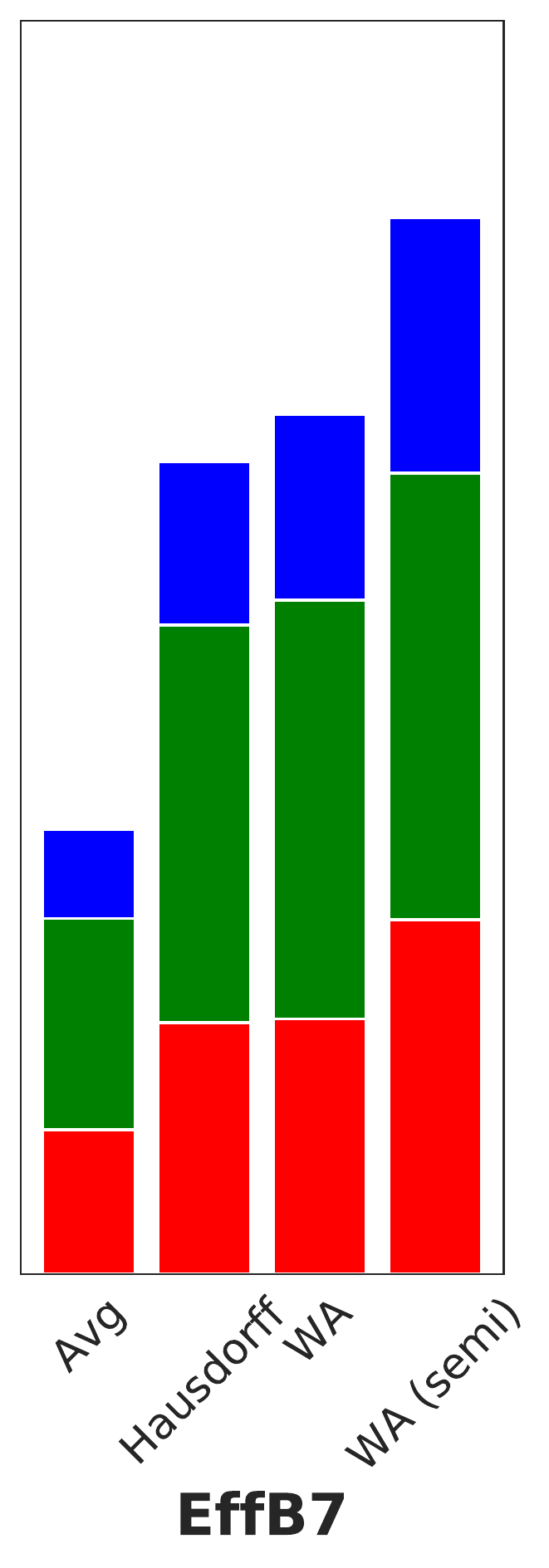}
    \includegraphics[height=2.3in]{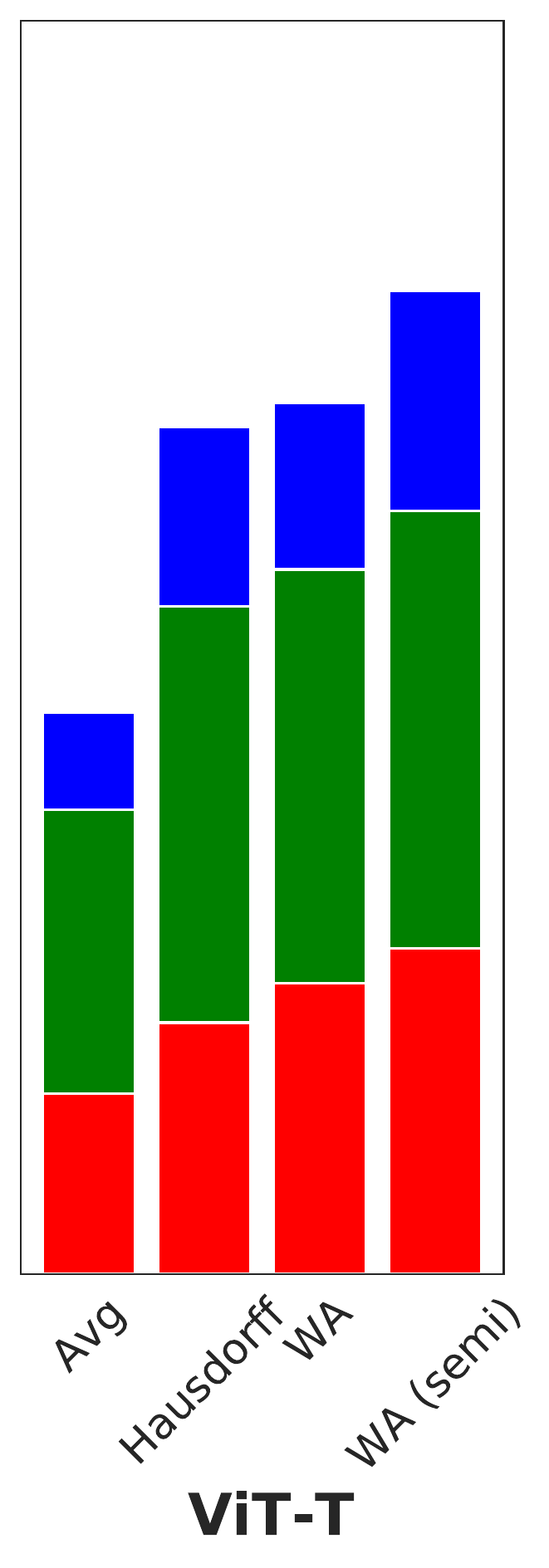}
    \includegraphics[height=2.3in]{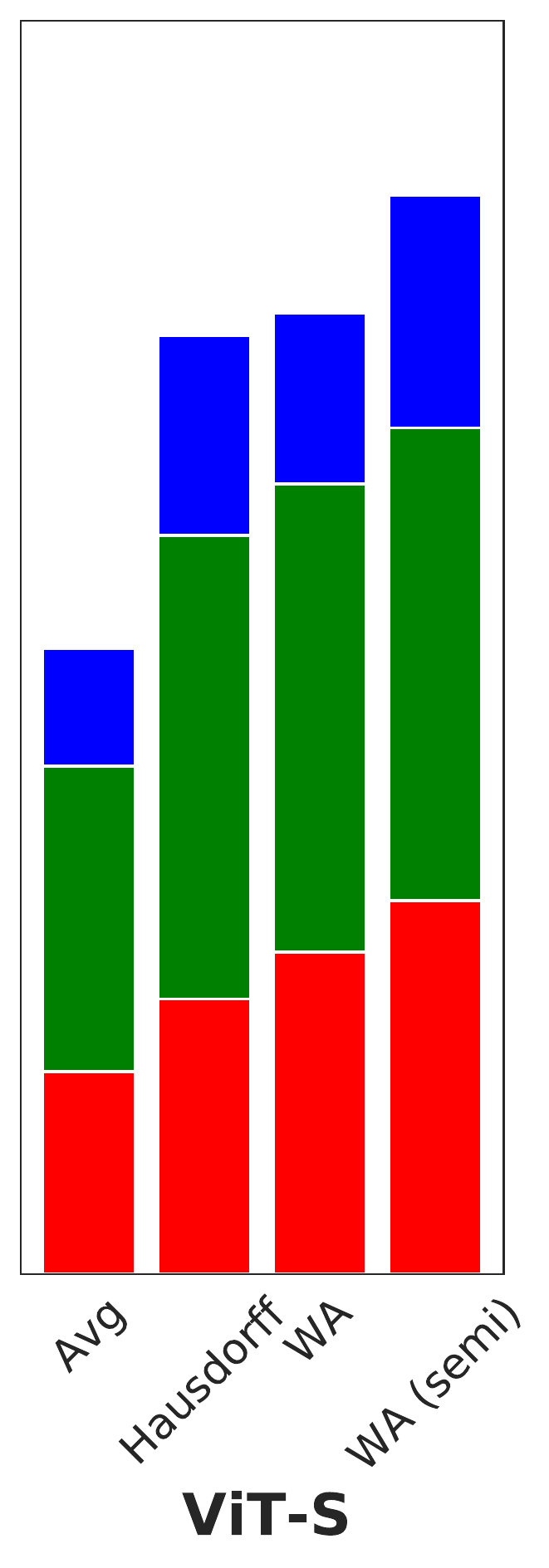}
    \includegraphics[height=2.3in]{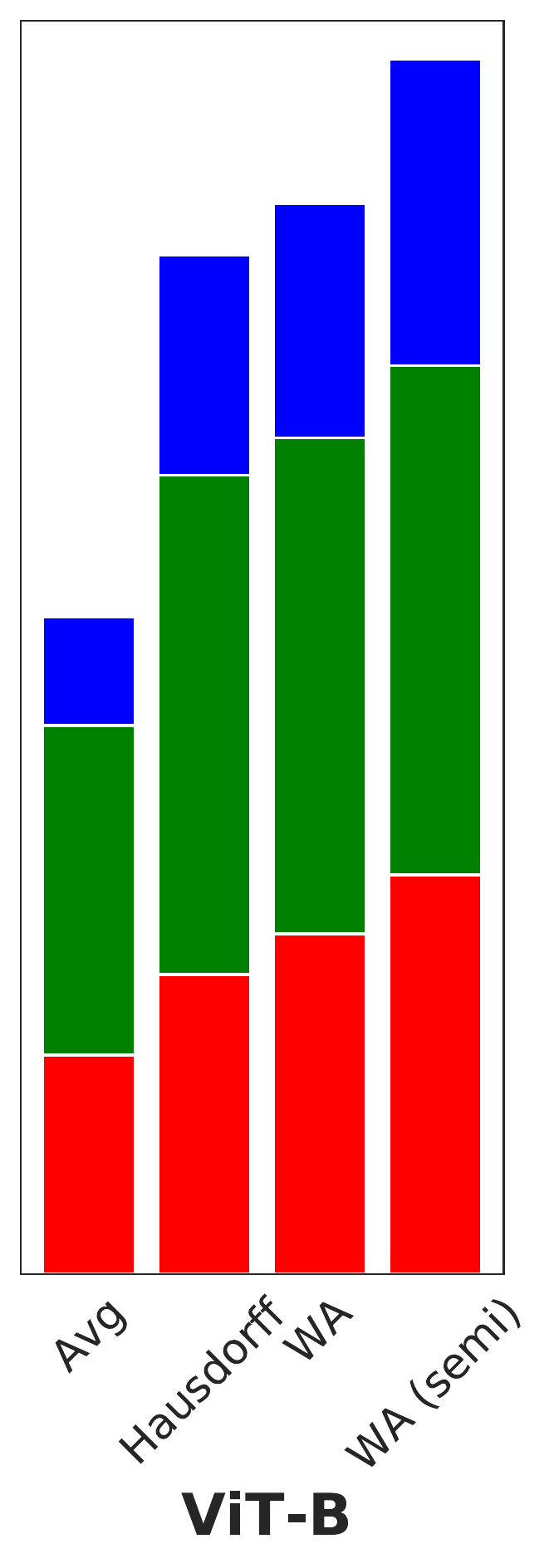}
    \includegraphics[height=2.3in]{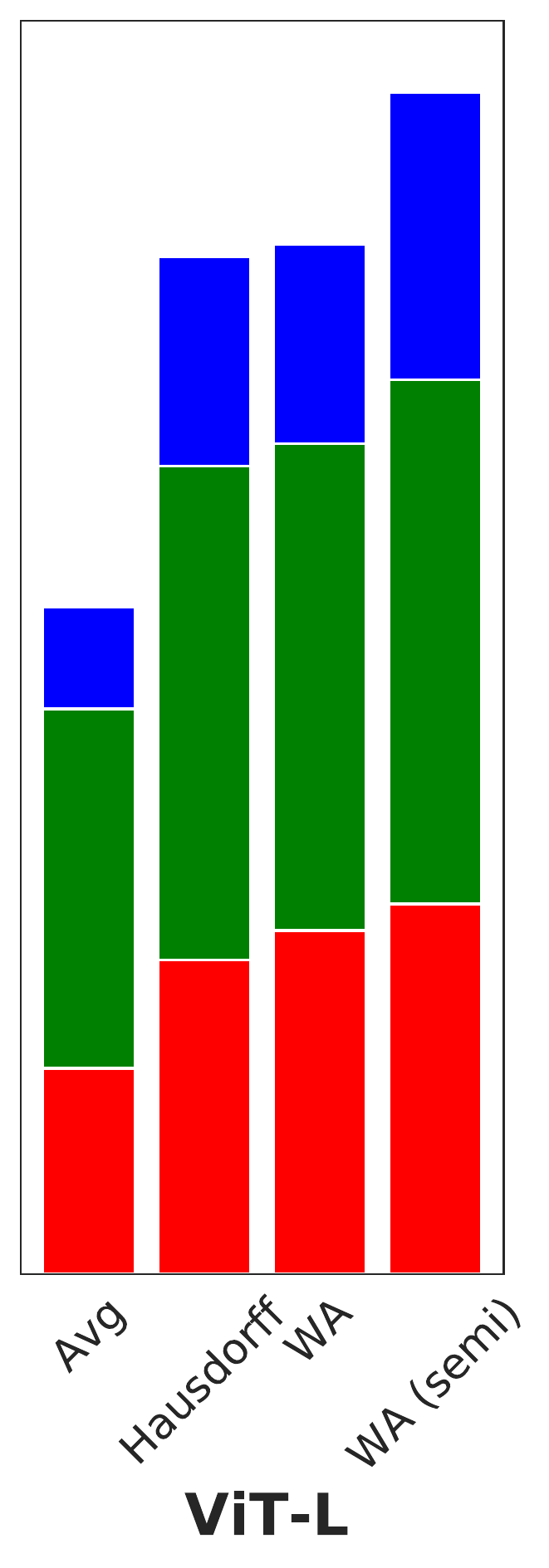}
    \caption{Bar plots with NMI scores over three datasets using various EfficientNet and ViT models with their intermediate layers as in Table~\ref{tab:app_network_config}. We show results for average, maximum Hausdorff distance, weighted average distance, and its semi-supervised version.}
    \label{fig:abl_feature_extractor_others}
\end{figure*}

%% file: table/table_07.tex
\begin{table*}[t]
    \centering
    \caption{Normalized mutual information (NMI), adjusted rand index (ARI) and F1 scores of unsupervised and semi-supervised clustering methods on MVTec (object, texture) and MTD datasets. Hierarchical Ward clustering is used for clustering, while various distance measures, such as average, maximum Hausdorff, or the proposed weighted average, are used to compute pairwise distances between data. We also report the performance of weighted average distance whose weights are generated from the ground-truth segmentation masks.}
    \label{tab:main_mvtec_per_cat}
    \vspace{0.05in}
    \resizebox{0.98\textwidth}{!}{%
    \begin{tabular}{c||c|c|c|c|c|c|c|c|c||c|c|c||c|c|c}
        \toprule
        %Setting & \multicolumn{9}{c||}{Unsupervised} & \multicolumn{3}{c||}{Semi-supervised} & \multicolumn{3}{c}{Segmentation mask} \\
        Supervision & \multicolumn{9}{c||}{Unsupervised} & \multicolumn{3}{c||}{Semi (labeled normal data)} & \multicolumn{3}{c}{Segmentation mask} \\
        \midrule
        Distance & \multicolumn{3}{c|}{Average} & \multicolumn{3}{c|}{Maximum Hausdorff} & \multicolumn{3}{c||}{Weighted Average} & \multicolumn{3}{c||}{Weighted Average} & \multicolumn{3}{c}{Weighted Average} \\
        \midrule
        Dataset & {   }NMI{   } & {   }ARI{   } & {   }F1{   } & {   }NMI{   } & {   }ARI{   } & {   }F1{   } & {   }NMI{   } & {   }ARI{   } & {   }F1{   } & {   }NMI{   } & {   }ARI{   } & {   }F1{   } & {   }NMI{   } & {   }ARI{   } & {   }F1{   } \\
        \midrule
        bottle & 0.426 & 0.186 & 0.448 & 0.585 & 0.510 & 0.764 & 0.495 & 0.421 & 0.567 & \textbf{0.607} & 0.461 & 0.639 & 0.531 & 0.438 & 0.584 \\
        cable & 0.439 & 0.209 & 0.421 & 0.636 & 0.348 & 0.579 & 0.730 & 0.673 & 0.770 & \textbf{0.889} & 0.903 & 0.935 & \textbf{\color{red}0.939} & 0.934 & 0.849 \\
        capsule & 0.172 & 0.060 & 0.339 & 0.156 & 0.045 & 0.276 & 0.185 & 0.070 & 0.380 & \textbf{0.334} & 0.191 & 0.466 & \textbf{\color{red}0.443} & 0.329 & 0.533 \\
        hazelnut & 0.063 & -0.003 & 0.314 & 0.552 & 0.327 & 0.500 & 0.568 & 0.430 & 0.610 & \textbf{0.868} & 0.889 & 0.936 & \textbf{\color{red}0.904} & 0.925 & 0.954 \\
        metal nut & 0.342 & 0.160 & 0.376 & 0.448 & 0.339 & 0.542 & 0.610 & 0.439 & 0.527 & \textbf{0.639} & 0.457 & 0.624 & 0.556 & 0.373 & 0.528 \\
        pill & 0.313 & 0.134 & 0.300 & 0.384 & 0.169 & 0.390 & 0.469 & 0.246 & 0.419 & \textbf{0.515} & 0.317 & 0.438 & \textbf{\color{red}0.653} & 0.484 & 0.576 \\
        screw & 0.049 & -0.000 & 0.264 & 0.031 & -0.006 & 0.239 & 0.038 & -0.007 & 0.251 & \textbf{0.376} & 0.267 & 0.418 & \textbf{\color{red}0.592} & 0.505 & 0.672 \\
        toothbrush & 0.000 & -0.018 & 0.581 & \textbf{0.251} & 0.050 & 0.652 & 0.214 & -0.008 & 0.599 & 0.214 & -0.008 & 0.599 & \textbf{\color{red}1.000} & 1.000 & 1.000 \\
        transistor & 0.282 & 0.110 & 0.497 & 0.499 & 0.478 & 0.703 & 0.573 & 0.674 & 0.755 & \textbf{0.651} & 0.462 & 0.594 & \textbf{\color{red}0.825} & 0.921 & 0.874 \\
        zipper & 0.353 & 0.255 & 0.454 & 0.606 & 0.491 & 0.615 & 0.628 & 0.521 & 0.648 & \textbf{0.677} & 0.552 & 0.635 & \textbf{\color{red}0.800} & 0.614 & 0.679 \\
        \midrule
        carpet & 0.287 & 0.138 & 0.392 & \textbf{0.660} & 0.586 & 0.795 & 0.656 & 0.576 & 0.647 & 0.550 & 0.430 & 0.553 & \textbf{\color{red}0.707} & 0.592 & 0.614 \\
        grid & 0.158 & 0.033 & 0.326 & 0.129 & 0.018 & 0.308 & 0.143 & 0.018 & 0.304 & \textbf{0.258} & 0.093 & 0.361 & 0.137 & 0.019 & 0.312 \\
        leather & 0.398 & 0.218 & 0.465 & 0.725 & 0.652 & 0.762 & 0.778 & 0.674 & 0.704 & \textbf{0.787} & 0.677 & 0.728 & 0.712 & 0.632 & 0.684 \\
        tile & 0.288 & 0.157 & 0.444 & 0.932 & 0.914 & 0.957 & \textbf{0.933} & 0.921 & 0.957 & 0.930 & 0.922 & 0.957 & \textbf{\color{red}1.000} & 1.000 & 1.000 \\
        wood & 0.231 & 0.066 & 0.384 & 0.678 & 0.500 & 0.716 & \textbf{0.860} & 0.815 & 0.921 & 0.823 & 0.725 & 0.893 & \textbf{\color{red}0.868} & 0.802 & 0.907 \\
        \midrule
        MTD & 0.065 & 0.024 & 0.289 & 0.193 & 0.112 & 0.381 & 0.179 & 0.120 & 0.346 & \textbf{0.390} & 0.314 & 0.490 & \textbf{\color{red}0.467} & 0.359 & 0.482 \\

        \bottomrule
    \end{tabular}
    }
\end{table*}

%% file: table/table_08.tex
% DONE
\begin{table*}[t]
    \centering
    \caption{Comparison to other clustering methods, including classic clustering methods such as KMeans, GMM, spectral, or hierarchical clustering with various linkages, using maximum Hausdorff (maxH) or weighted average (WA) distances, and deep clustering methods, such as IIC~\cite{ji2019invariant}, GATCluster~\cite{niu2020gatcluster}, or SCAN~\cite{van2020scan}. For deep clustering methods, we also provide in the parenthesis the performance of the best training epoch chosen by the test set accuracy.}
    \label{tab:supp_clusterer_comparison}
    \vspace{0.05in}
    \resizebox{0.98\linewidth}{!}{%
    \begin{tabular}{l|c|c|c|c|c|c|c|c|c|c|c|c|c|c|c|c|c|c}
        \toprule
        Dataset & \multicolumn{6}{c|}{MVTec Object} & \multicolumn{6}{c|}{MVTec Texture} & \multicolumn{6}{c}{Magnetic Tile Defect} \\ \cmidrule{1-19}
        Distance & \multicolumn{3}{c|}{maxH} & \multicolumn{3}{c|}{WA} & \multicolumn{3}{c|}{maxH} & \multicolumn{3}{c|}{WA} & \multicolumn{3}{c|}{maxH} & \multicolumn{3}{c}{WA} \\ \cmidrule{1-19}
        Metric & {   }NMI{   } & {   }ARI{   } & {   }F1{   } & {   }NMI{   } & {   }ARI{   } & {   }F1{   } & {   }NMI{   } & {   }ARI{   } & {   }F1{   } & {   }NMI{   } & {   }ARI{   } & {   }F1{   } & {   }NMI{   } & {   }ARI{   } & {   }F1{   } & {   }NMI{   } & {   }ARI{   } & {   }F1{   } \\
        \midrule
        \iffalse
        KMeans & \multicolumn{3}{c|}{--} & 0.390 & 0.237 & 0.509 & \multicolumn{3}{c|}{--} & 0.582 & 0.497 & 0.674 & \multicolumn{3}{c|}{--} & 0.180 & 0.116 & 0.359 \\
        GMM & \multicolumn{3}{c|}{--} & 0.365 & 0.228 & 0.475 & \multicolumn{3}{c|}{--} & 0.512 & 0.397 & 0.581 & \multicolumn{3}{c|}{--} & 0.177 & 0.116 & 0.371 \\
        Spectral & 0.401 & 0.275 & 0.539 & 0.413 & 0.285 & 0.546 & 0.628 & 0.563 & 0.734 & 0.540 & 0.433 & 0.622 & 0.133 & 0.081 & 0.345 & 0.148 & 0.097 & 0.346 \\
        Single & 0.108 & 0.028 & 0.241 & 0.145 & 0.044 & 0.264 & 0.075 & 0.008 & 0.173 & 0.109 & 0.016 & 0.178 & 0.087 & 0.019 & 0.202 & 0.062 & 0.011 & 0.196 \\
        Complete & 0.312 & 0.183 & 0.405 & 0.345 & 0.202 & 0.466 & 0.406 & 0.228 & 0.388 & 0.528 & 0.364 & 0.532 & 0.125 & 0.062 & 0.317 & 0.131 & 0.065 & 0.328 \\
        Average & 0.185 & 0.064 & 0.295 & 0.263 & 0.103 & 0.334 & 0.302 & 0.126 & 0.311 & 0.267 & 0.106 & 0.324 & 0.072 & 0.022 & 0.236 & 0.093 & 0.028 & 0.276 \\
        Ward & 0.415 & 0.274 & 0.507 & \textbf{0.448} & \textbf{0.309} & \textbf{0.528} & 0.644 & 0.527 & \textbf{0.699} & \textbf{0.666} & \textbf{0.582} & 0.696 & \textbf{0.164} & 0.086 & \textbf{0.355} & \textbf{0.164} & 0.095 & \textbf{0.349} \\
        \fi
        KMeans & \multicolumn{3}{c|}{--}   & 0.429 & 0.301 & 0.544 & \multicolumn{3}{c|}{--}   & 0.642 & 0.567 & 0.714 & \multicolumn{3}{c|}{--}   & \textbf{0.204} & 0.135 & 0.374 \\
        GMM & \multicolumn{3}{c|}{--}   & 0.395 & 0.264 & 0.498 & \multicolumn{3}{c|}{--}   & 0.578 & 0.469 & 0.635 & \multicolumn{3}{c|}{--}   & \textbf{0.204} & 0.141 & 0.377 \\
        Spectral & 0.419 & 0.287 & 0.546 & 0.428 & 0.305 & 0.555 & 0.609 & 0.525 & 0.702 & 0.606 & 0.516 & 0.681 & 0.143 & 0.089 & 0.354 & 0.150 & 0.098 & 0.341 \\
        Single & 0.108 & 0.025 & 0.238 & 0.133 & 0.041 & 0.261 & 0.078 & 0.008 & 0.173 & 0.108 & 0.005 & 0.186 & 0.087 & 0.019 & 0.202 & 0.065 & 0.012 & 0.200 \\
        Complete & 0.316 & 0.187 & 0.409 & 0.294 & 0.146 & 0.405 & 0.360 & 0.184 & 0.356 & 0.452 & 0.265 & 0.510 & 0.128 & 0.062 & 0.320 & 0.116 & 0.075 & 0.310 \\
        Average & 0.245 & 0.109 & 0.328 & 0.276 & 0.095 & 0.345 & 0.223 & 0.064 & 0.294 & 0.400 & 0.213 & 0.398 & 0.080 & 0.024 & 0.242 & 0.094 & 0.034 & 0.284 \\
        Ward & 0.415 & 0.275 & 0.526 & \textbf{0.451} & 0.346 & 0.553 & 0.625 & 0.534 & 0.708 & \textbf{0.674} & 0.601 & 0.707 & 0.193 & 0.112 & 0.381 & 0.179 & 0.120 & 0.346 \\
        \midrule\midrule
        & \multicolumn{2}{c|}{NMI} & \multicolumn{2}{c|}{ARI} & \multicolumn{2}{c|}{F1} & \multicolumn{2}{c|}{NMI} & \multicolumn{2}{c|}{ARI} & \multicolumn{2}{c|}{F1} & \multicolumn{2}{c|}{NMI} & \multicolumn{2}{c|}{ARI} & \multicolumn{2}{c}{F1} \\\midrule
        {IIC} & \multicolumn{2}{c|}{0.086 (0.170)} & \multicolumn{2}{c|}{0.019 (0.117)} & \multicolumn{2}{c|}{0.297 (0.366)} & \multicolumn{2}{c|}{0.107 (0.188)} & \multicolumn{2}{c|}{0.023 (0.096)} & \multicolumn{2}{c|}{0.261 (0.300)} & \multicolumn{2}{c|}{0.064 (0.034)} & \multicolumn{2}{c|}{0.020 (0.017)} & \multicolumn{2}{c}{0.252 (0.230)} \\
        {GATCluster} & \multicolumn{2}{c|}{0.119 (0.265)} & \multicolumn{2}{c|}{0.044 (0.202)} & \multicolumn{2}{c|}{0.320 (0.475)} & \multicolumn{2}{c|}{0.171 (0.298)} & \multicolumn{2}{c|}{0.072 (0.202)} & \multicolumn{2}{c|}{0.305 (0.442)} & \multicolumn{2}{c|}{0.028 (0.113)} & \multicolumn{2}{c|}{0.009 (0.064)} & \multicolumn{2}{c}{0.243 (0.333)} \\
        {SCAN} & \multicolumn{2}{c|}{0.176 (0.198)} & \multicolumn{2}{c|}{0.078 (0.123)} & \multicolumn{2}{c|}{0.335 (0.393)} & \multicolumn{2}{c|}{0.277 (0.314)} & \multicolumn{2}{c|}{0.153 (0.203)} & \multicolumn{2}{c|}{0.335 (0.393)} & \multicolumn{2}{c|}{0.071 (0.087)} & \multicolumn{2}{c|}{0.029 (0.053)} & \multicolumn{2}{c}{0.282 (0.309)} \\
        %{IIC} & 0.086 (0.170) & 0.107 (0.188) & \\
        %{GATCluster} & 0.119 (0.265) & 0.171 (0.298) & \\
        %{SCAN} & 0.176 (0.198) & 0.277 (0.314) & \\
        \bottomrule
    \end{tabular}
    }
\end{table*}

%% file: table/table_01_supp.tex
\begin{table*}[t]
    \centering
    \caption{NMI, ARI and F1 scores of unsupervised and semi-supervised clustering methods on MVTec (object, texture) datasets. Compared to the baseline method (``average'') that uses a holistic representation via average pooling of patch embeddings, the multiple instance clustering framework with various distance measures, such as maximum Hausdorff or the proposed weighted average distances, show huge improvement. We also report the performance of weighted average distance whose weights are computed using labeled normal data (``Semi''). 
    Furthermore, we include extended baselines using max pooling, generalized mean pooling (GeM), sum-pooled convolutional features (SPoC)~\cite{babenko2015aggregating}, and bag-of-words with the codebook size of 512.
    We test each method on the random subsets including 90\% images of the test set for 100 different random seeds to compute mean and standard errors.
    }
    \label{tab:supp_mvtec}
    \vspace{0.05in}
    \resizebox{0.9\textwidth}{!}{%
    \begin{tabular}{c||c|c|c|c|c|c|c|c|c||c|c|c}
        \toprule
        Supervision & \multicolumn{9}{c||}{Unsupervised} & \multicolumn{3}{c}{Semi} \\
        \midrule
        Metric & {\,\,}NMI{\,\,} & {\,\,}ARI{\,\,} & {\,\,}F1{\,\,} & {\,\,}NMI{\,\,} & {\,\,}ARI{\,\,} & {\,\,}F1{\,\,} & {\,\,}NMI{\,\,} & {\,\,}ARI{\,\,} & {\,\,}F1{\,\,} & {\,\,}NMI{\,\,} & {\,\,}ARI{\,\,} & {\,\,}F1{\,\,} \\
        \midrule
        Distance & \multicolumn{3}{c|}{Average} & \multicolumn{3}{c|}{{\,\,}Maximum Hausdorff{\,\,}} & \multicolumn{3}{c||}{{\,\,}Weighted Average{\,\,}} & \multicolumn{3}{c}{{\,\,}Weighted Average{\,\,}} \\
        \midrule
        MVTec (object) & 0.249 & 0.114 & 0.412 & 0.423 & 0.274 & 0.520 & 0.458 & 0.333 & 0.563 & 0.584 & 0.477 & 0.653\\
        std err. & (0.002) & (0.003) & (0.004) & (0.004) & (0.005) & (0.004) & (0.003) & (0.005) & (0.004) & (0.004) & (0.006) & (0.005)\\ \midrule
        MVTec (texture) & 0.288 & 0.122 & 0.405 & 0.650 & 0.560 & 0.722 & 0.665 & 0.582 & 0.709 & 0.702 & 0.616 & 0.743 \\
        std err. & (0.003) & (0.003) & (0.003) & (0.004) & (0.005) & (0.005) & (0.003) & (0.004) & (0.003) & (0.004) & (0.005) & (0.004) \\
        \bottomrule
    \end{tabular}
    }
    \resizebox{0.9\textwidth}{!}{%
    \begin{tabular}{c||c|c|c|c|c|c|c|c|c|c|c|c}
        \toprule
        Distance & \multicolumn{3}{c|}{Max} & \multicolumn{3}{c|}{GeM ($p\,{=}\,20$)} & \multicolumn{3}{c|}{SPoC ($\sigma\,{=}\,1000$)} & \multicolumn{3}{c}{Bag-of-Words} \\
        \midrule
        MVTec (object) & 0.336 & 0.204 & 0.488 & 0.338 & 0.209 & 0.486 & 0.249 & 0.114 & 0.412 & 0.226 & 0.102 & 0.396 \\
        std err. & (0.003) & (0.004) & (0.004) & (0.003) & (0.004) & (0.004) & (0.002) & (0.003) & (0.004) & (0.003) & (0.003) & (0.003) \\ \midrule
        MVTec (texture) & 0.598 & 0.478 & 0.658 & 0.602 & 0.482 & 0.660 & 0.288 & 0.122 & 0.405 & 0.312 & 0.126 & 0.359 \\
        std err. & (0.004) & (0.004) & (0.003) & (0.003) & (0.004) & (0.003) & (0.003) & (0.003) & (0.003) & (0.004) & (0.004) & (0.004) \\ \midrule
        \bottomrule
    \end{tabular}
    }
    %\vspace{-0.15in}
\end{table*}

%% file: table/table_03_supp.tex
\begin{table}[t]
    \centering
    \caption{Comparison to other clustering methods, including KMeans, KMedoids, GMM, spectral, and hierarchical clustering with various linkages, using maximum Hausdorff (maxH) or weighted average (WA) distances, and deep clustering methods, such as IIC~\cite{ji2019invariant}, GATCluster~\cite{niu2020gatcluster}, or SCAN~\cite{van2020scan}. For deep clustering methods, we provide in the parenthesis the performance of the best training epoch chosen by test set accuracy. We test each method on the random subsets including 90\% images of the test set for 100 different random seeds to compute mean and standard errors.}
    \label{tab:supp_clusterer_comparison_std}
    \vspace{0.05in}
    \resizebox{0.7\linewidth}{!}{%
    \begin{tabular}{l|c|c|c|c|c|c}
        \toprule
        Dataset & \multicolumn{2}{c|}{{\,\,\,}MVTec (object){\,\,\,}} & \multicolumn{2}{c|}{{\,\,\,}MVTec (texture){\,\,\,}} & \multicolumn{2}{c}{MTD} \\ \cmidrule{1-7}
        Distance & {  }maxH{  } & WA & {  }maxH{  } & WA & {  }maxH{  } & {\,\,\,\,\,\,}WA{\,\,\,\,\,\,} \\
        \midrule
        KMeans & --   & 0.429{\footnotesize{$\pm$0.002}} & --   & 0.637{\footnotesize{$\pm$0.002}} & --  & \textbf{0.204} \\
        GMM & --   & 0.397{\footnotesize{$\pm$0.002}} & --   & 0.583{\footnotesize{$\pm$0.003}} & --  & \textbf{0.204} \\
        KMedoids & 0.152{\footnotesize{$\pm$0.005}} & 0.250{\footnotesize{$\pm$0.004}} & 0.301{\footnotesize{$\pm$0.005}} & 0.391{\footnotesize{$\pm$0.006}} & 0.050 & 0.076 \\
        Spectral & 0.415{\footnotesize{$\pm$0.003}} & 0.422{\footnotesize{$\pm$0.002}} & 0.618{\footnotesize{$\pm$0.003}} & 0.606{\footnotesize{$\pm$0.003}} & 0.143 & 0.150 \\
        Single & 0.122{\footnotesize{$\pm$0.003}} & 0.141{\footnotesize{$\pm$0.003}} & 0.086{\footnotesize{$\pm$0.002}} & 0.116{\footnotesize{$\pm$0.002}} & 0.087 & 0.065 \\
        Complete & 0.321{\footnotesize{$\pm$0.005}} & 0.339{\footnotesize{$\pm$0.005}} & 0.404{\footnotesize{$\pm$0.007}} & 0.495{\footnotesize{$\pm$0.007}} & 0.128 & 0.116 \\
        Average & 0.225{\footnotesize{$\pm$0.005}} & 0.213{\footnotesize{$\pm$0.002}} & 0.272{\footnotesize{$\pm$0.007}} & 0.367{\footnotesize{$\pm$0.007}} & 0.080 & 0.094 \\
        Ward & 0.423{\footnotesize{$\pm$0.004}} & \textbf{0.458}{\footnotesize{$\pm$0.003}} & 0.650{\footnotesize{$\pm$0.004}} & \textbf{0.665}{\footnotesize{$\pm$0.003}} & 0.193 & 0.179 \\
        \midrule
        {IIC} & \multicolumn{2}{c|}{0.086 (0.170)} & \multicolumn{2}{c|}{0.107 (0.188)} & \multicolumn{2}{c}{0.064 (0.034)} \\
        {GATCluster}{  } & \multicolumn{2}{c|}{0.119 (0.265)} & \multicolumn{2}{c|}{0.171 (0.298)} & \multicolumn{2}{c}{0.028 (0.113)} \\
        {SCAN} & \multicolumn{2}{c|}{0.176 (0.198)} & \multicolumn{2}{c|}{0.277 (0.314)} & \multicolumn{2}{c}{0.071 (0.087)} \\
        %{IIC} & 0.086 (0.170) & 0.107 (0.188) & \\
        %{GATCluster} & 0.119 (0.265) & 0.171 (0.298) & \\
        %{SCAN} & 0.176 (0.198) & 0.277 (0.314) & \\
        \bottomrule
    \end{tabular}
    }
    \vspace{-0.15in}
\end{table}

%% file: table/figure_09.tex
\begin{figure}[t]
    \centering
    \begin{subfigure}{0.3\textwidth}
        \centering
        \includegraphics[width=\linewidth]{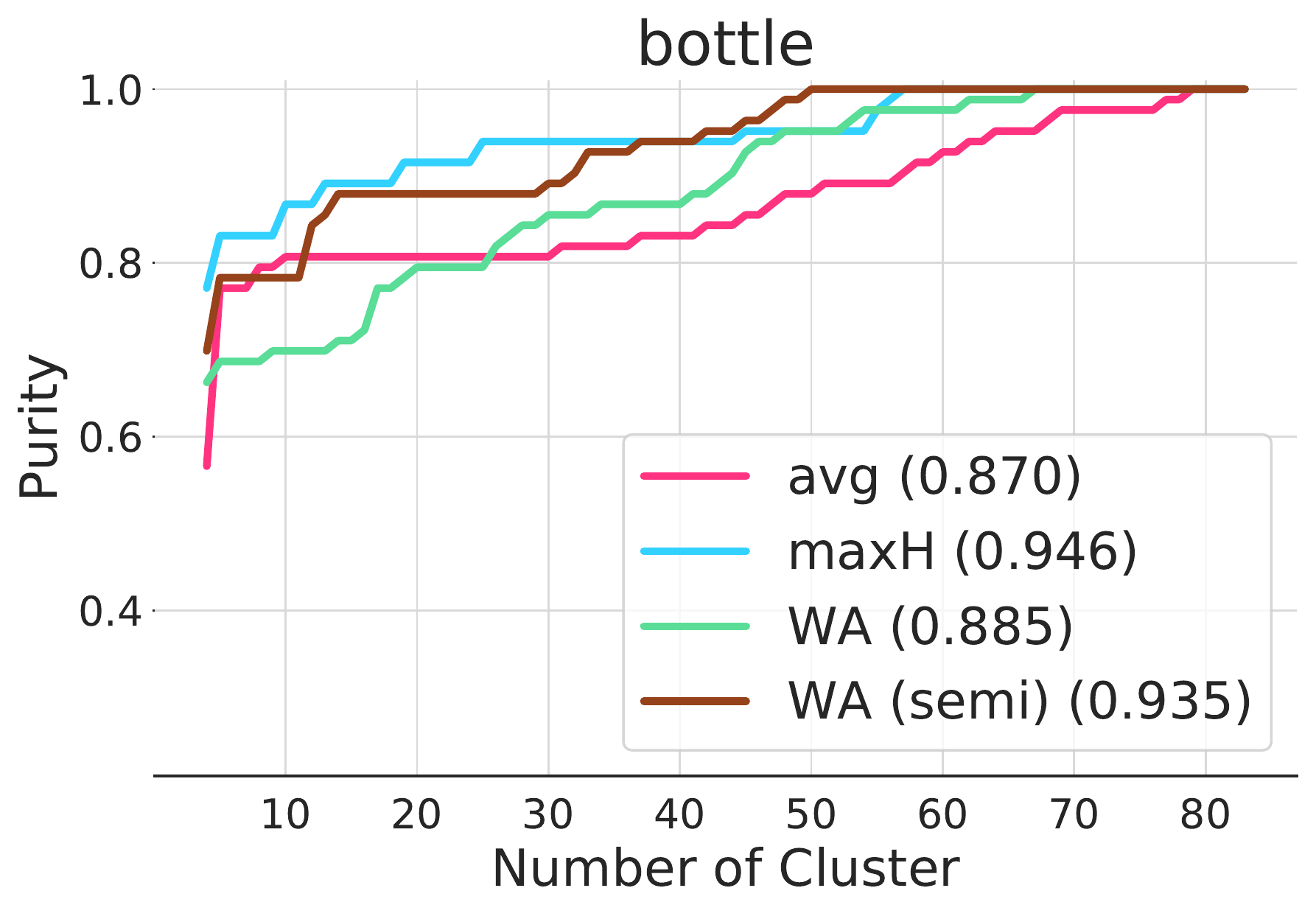}
    \end{subfigure}
    \begin{subfigure}{0.3\textwidth}
        \centering
        \includegraphics[width=\linewidth]{fig/purity/purity-cable.pdf}
    \end{subfigure}
    \begin{subfigure}{0.3\textwidth}
        \centering
        \includegraphics[width=\linewidth]{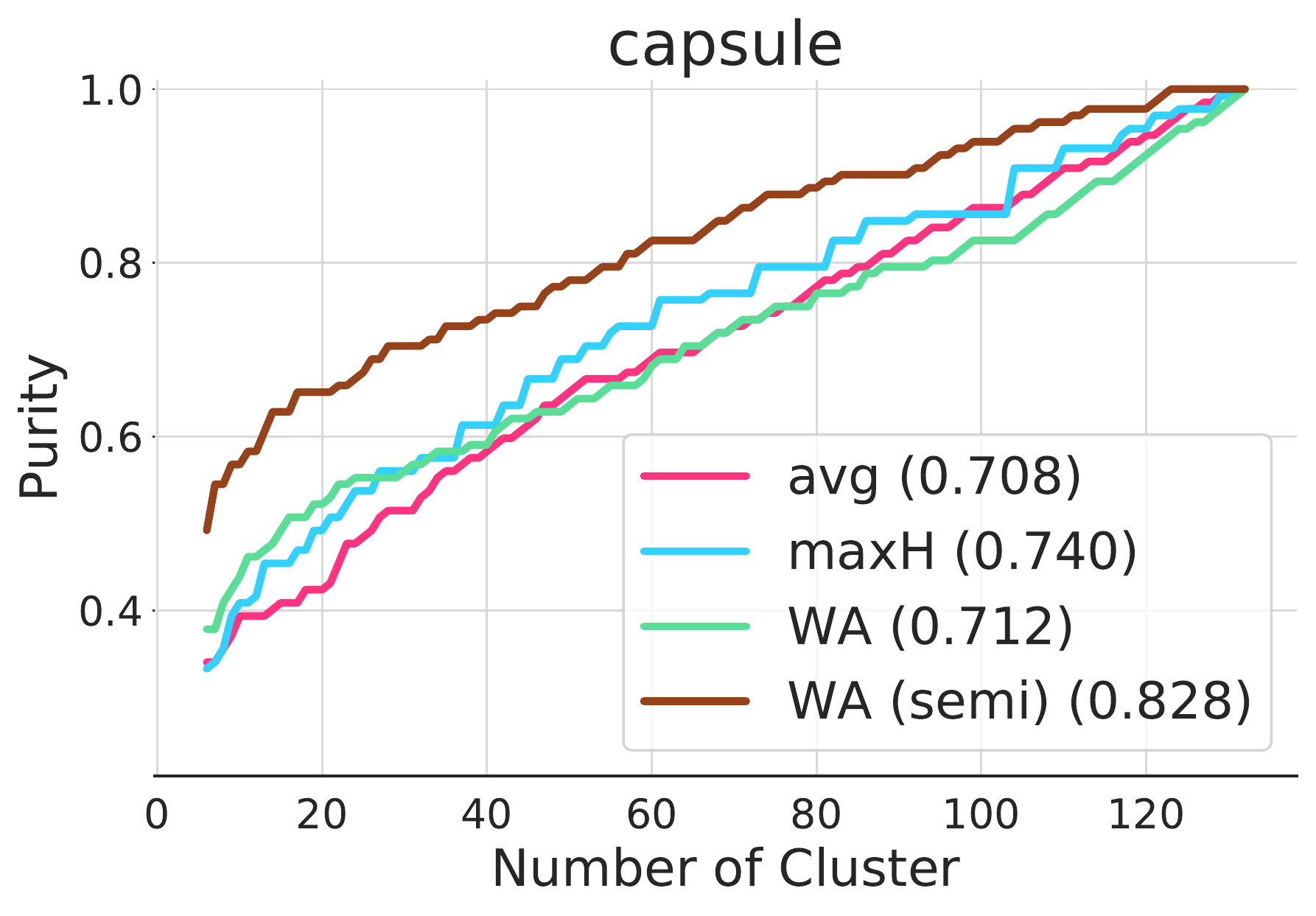}
    \end{subfigure}
    \begin{subfigure}{0.3\textwidth}
        \centering
        \includegraphics[width=\linewidth]{fig/purity/purity-hazelnut.pdf}
    \end{subfigure}
    \begin{subfigure}{0.3\textwidth}
        \centering
        \includegraphics[width=\linewidth]{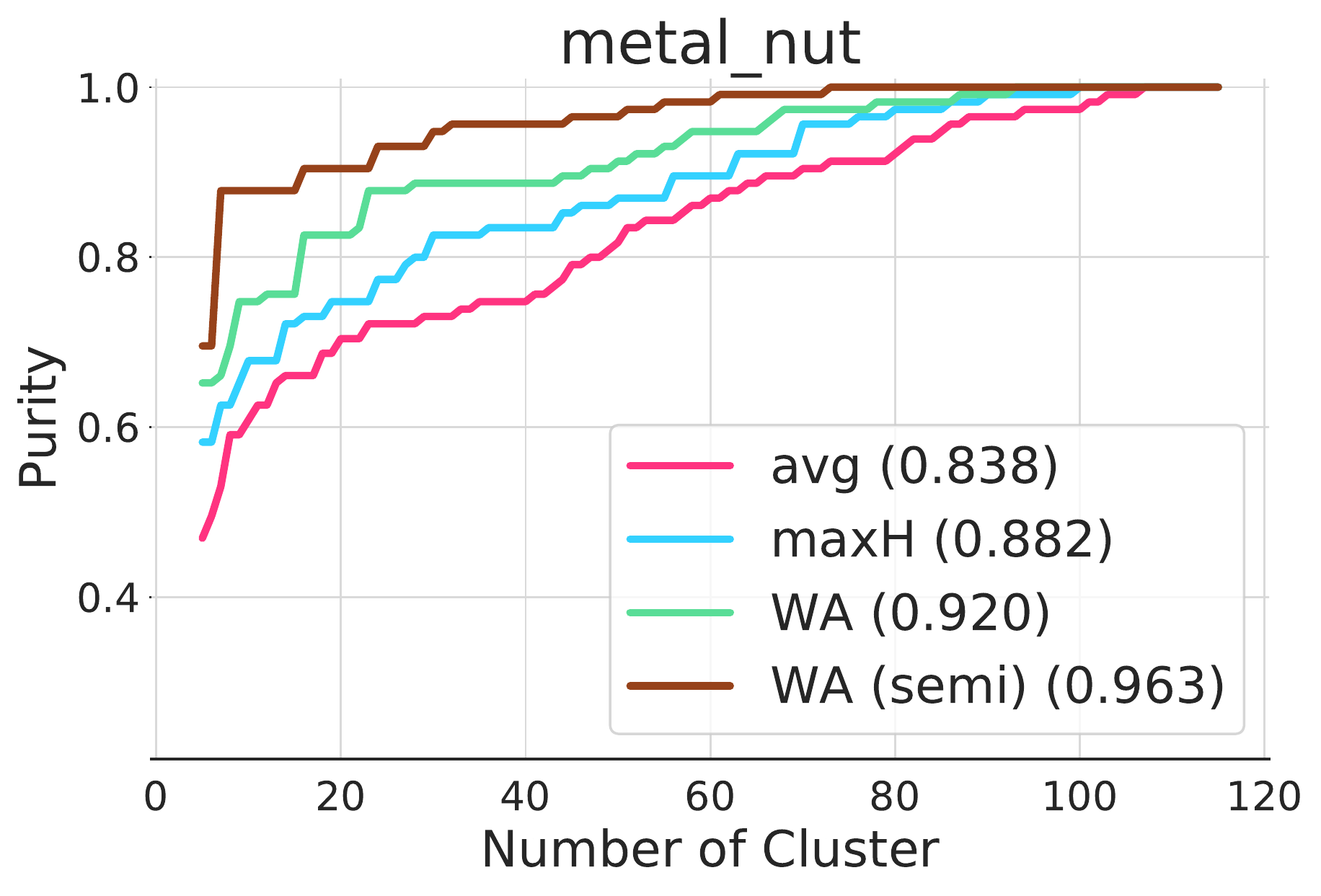}
    \end{subfigure}
    \begin{subfigure}{0.3\textwidth}
        \centering
        \includegraphics[width=\linewidth]{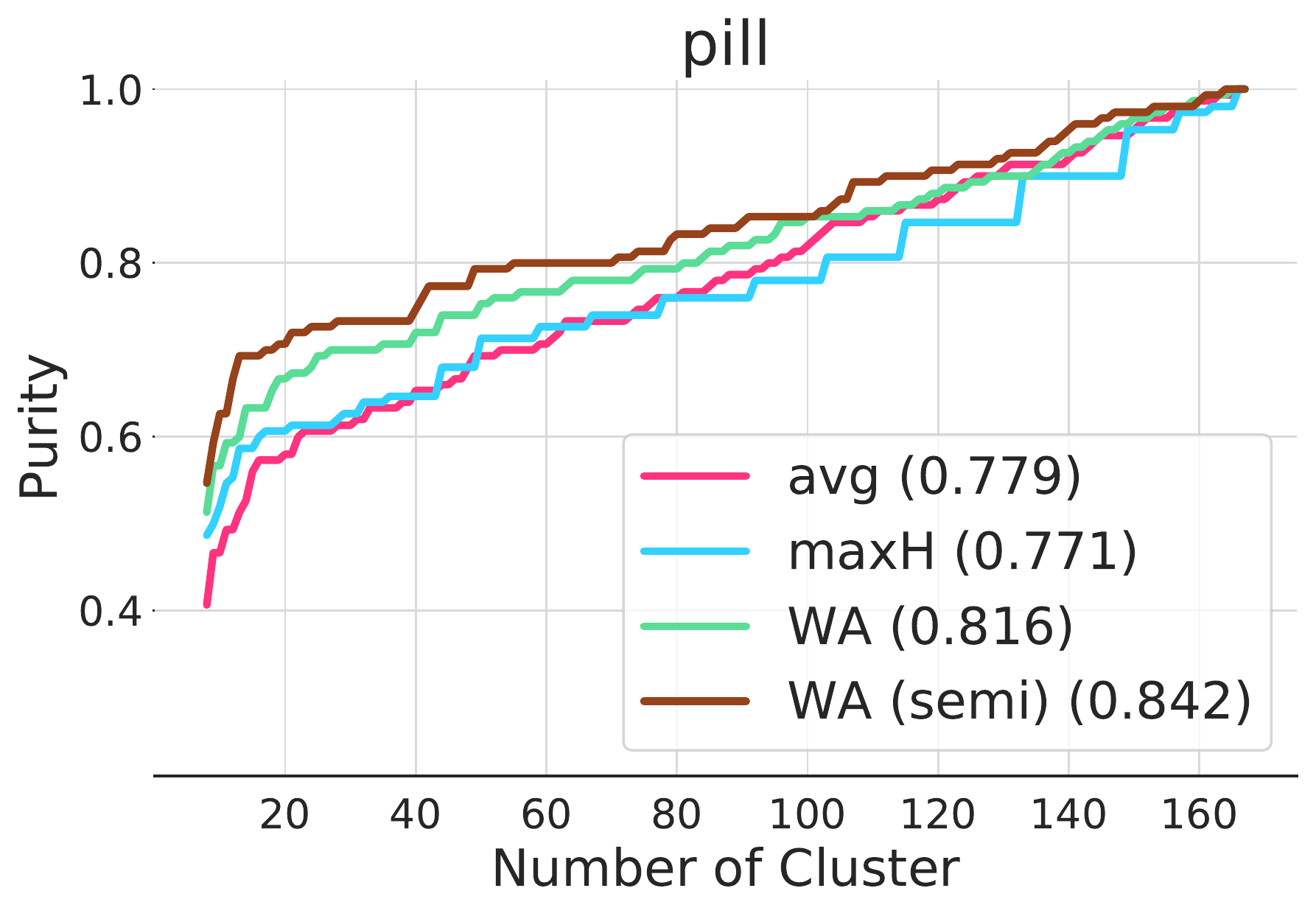}
    \end{subfigure}
    \begin{subfigure}{0.3\textwidth}
        \centering
        \includegraphics[width=\linewidth]{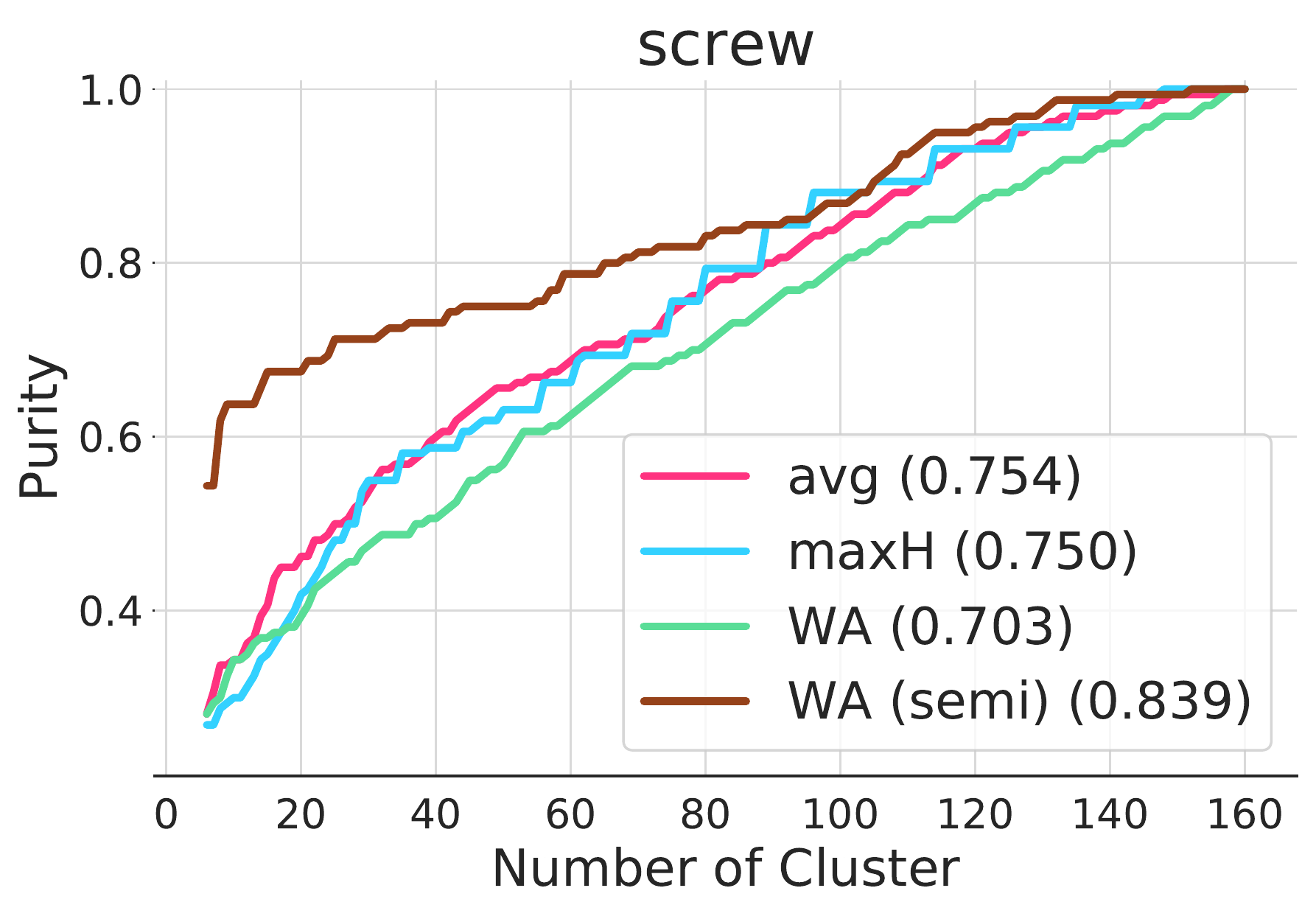}
    \end{subfigure}
    \begin{subfigure}{0.3\textwidth}
        \centering
        \includegraphics[width=\linewidth]{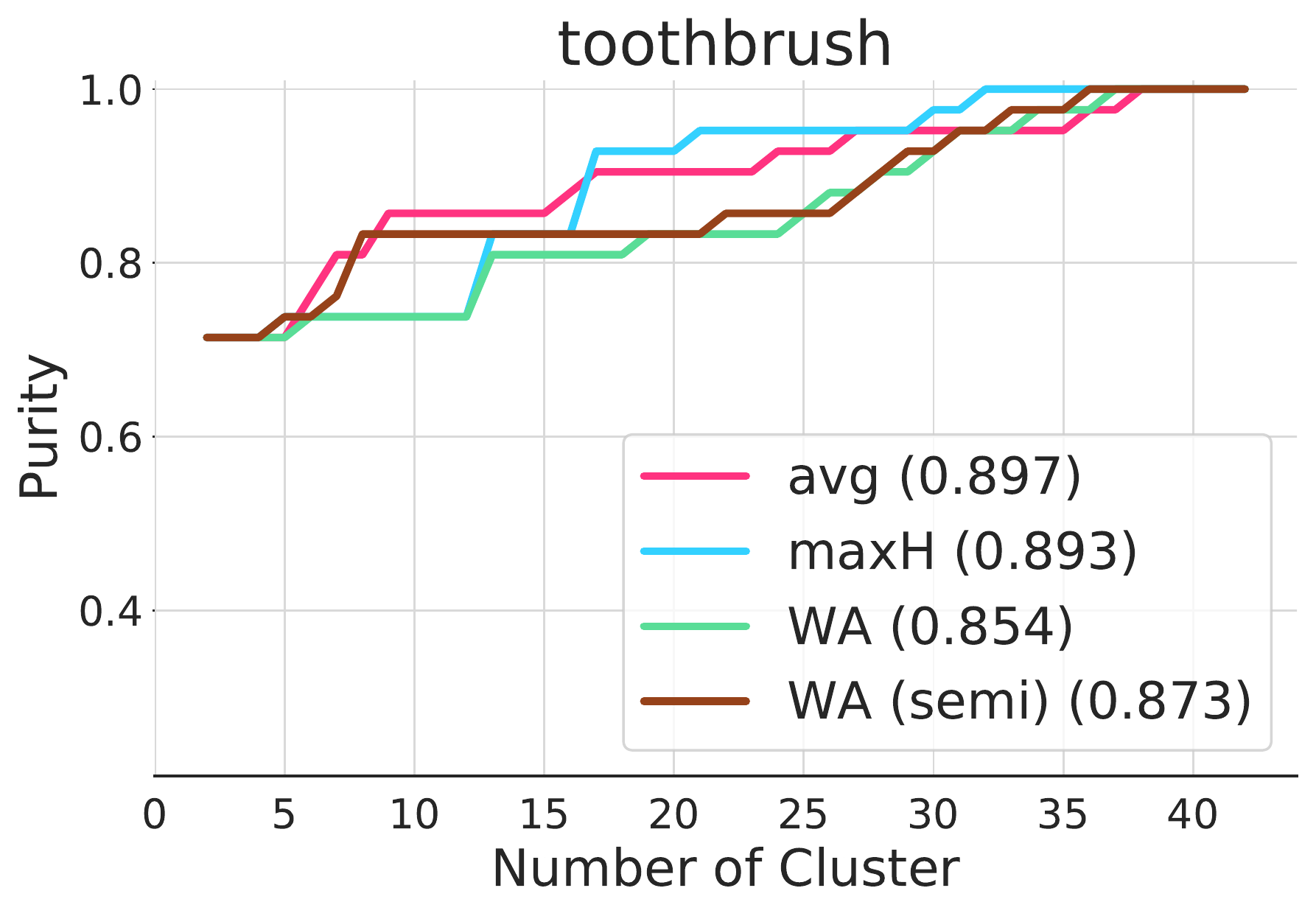}
    \end{subfigure}
    \begin{subfigure}{0.3\textwidth}
        \centering
        \includegraphics[width=\linewidth]{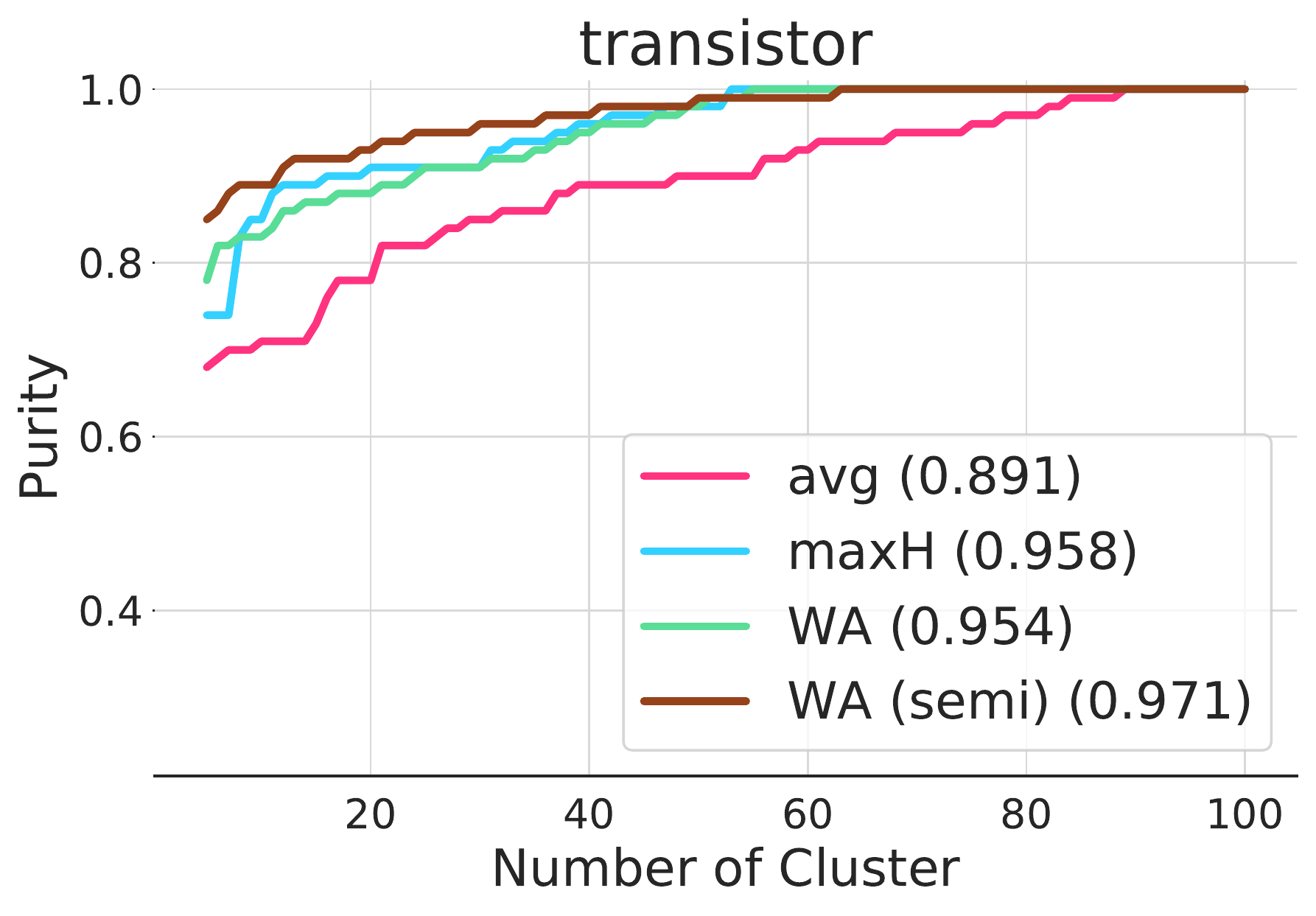}
    \end{subfigure}
    \begin{subfigure}{0.3\textwidth}
        \centering
        \includegraphics[width=\linewidth]{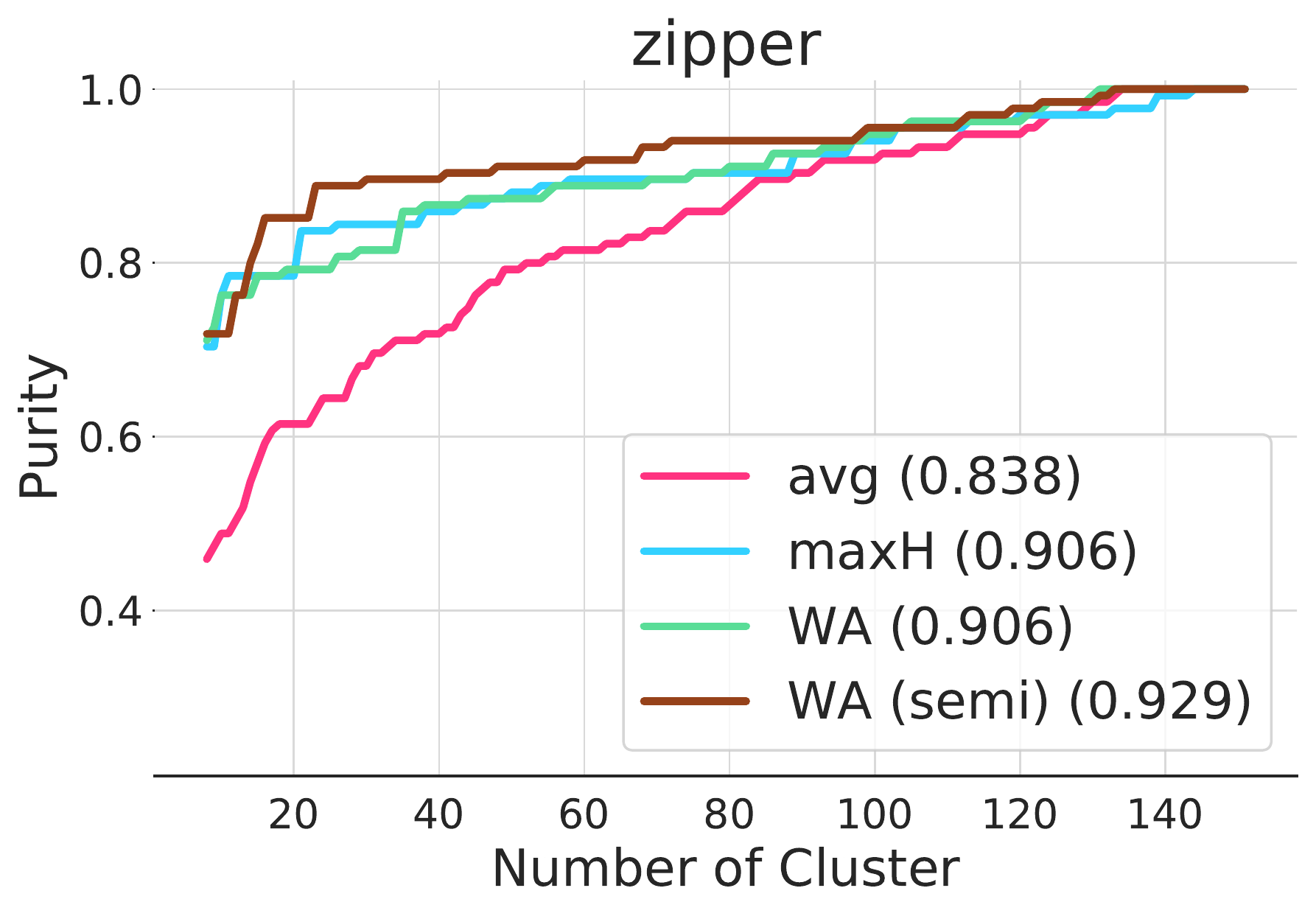}
    \end{subfigure}
    \begin{subfigure}{0.3\textwidth}
        \centering
        \includegraphics[width=\linewidth]{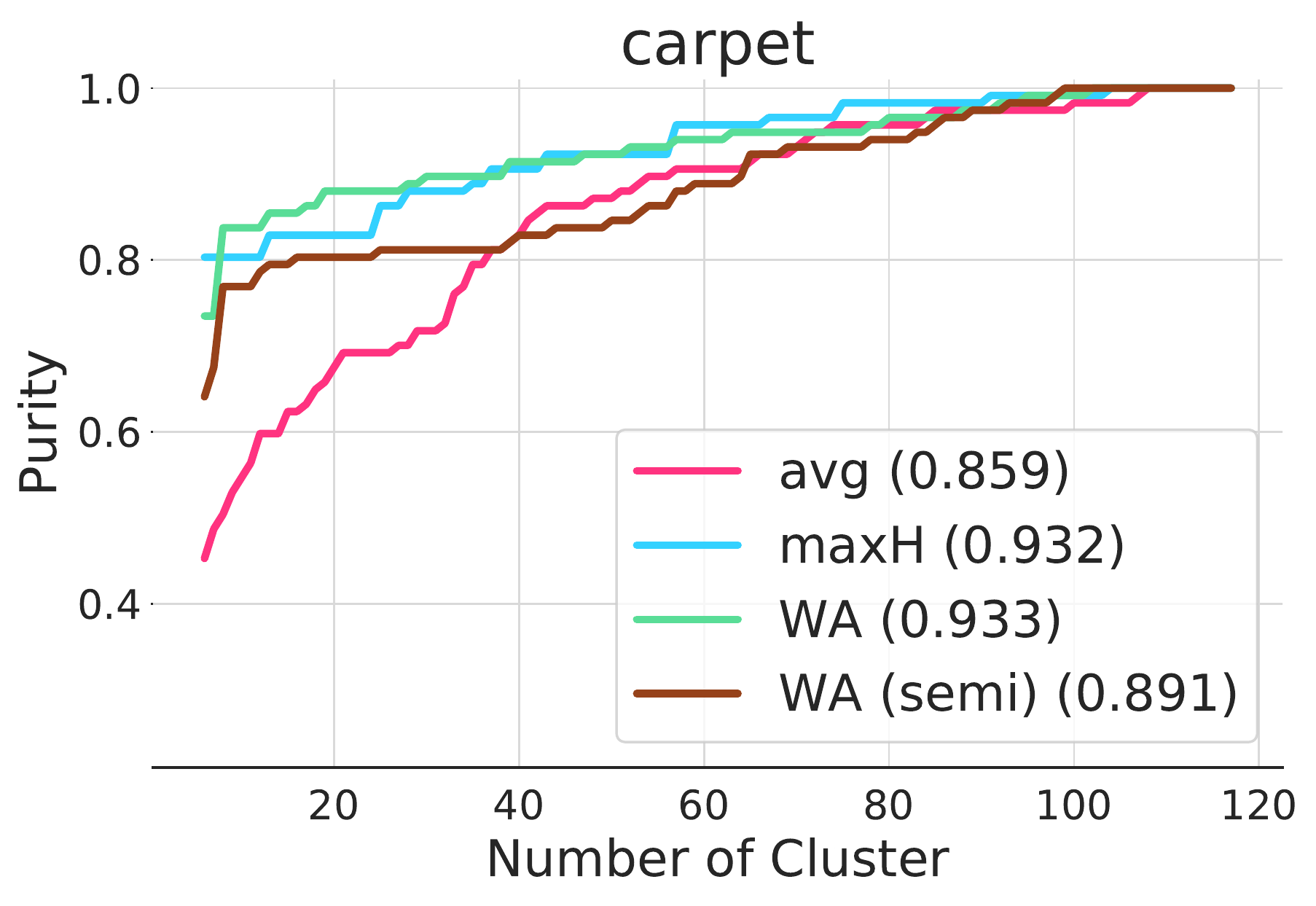}
    \end{subfigure}
    \begin{subfigure}{0.3\textwidth}
        \centering
        \includegraphics[width=\linewidth]{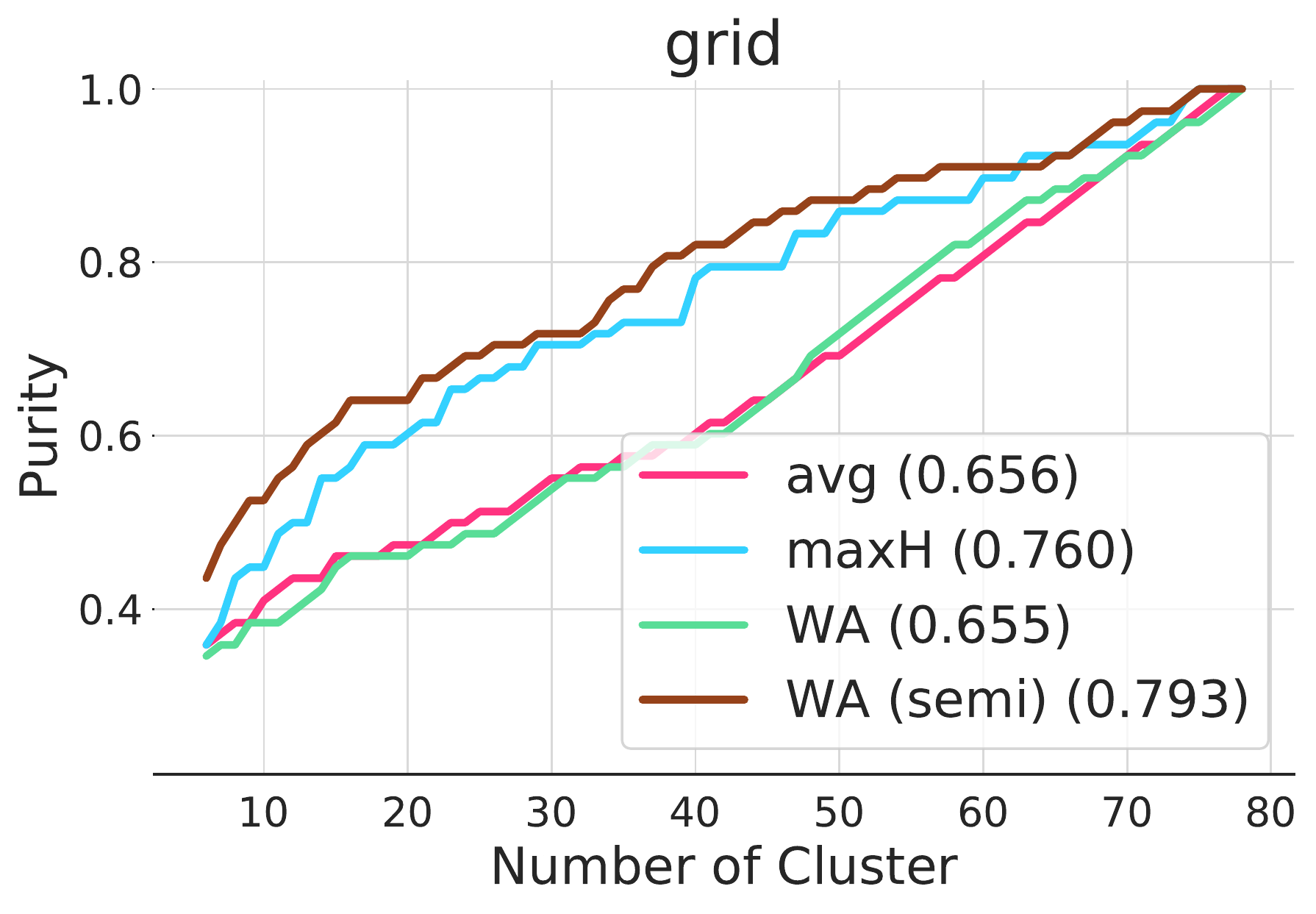}
    \end{subfigure}
    \begin{subfigure}{0.3\textwidth}
        \centering
        \includegraphics[width=\linewidth]{fig/purity/purity-leather.pdf}
    \end{subfigure}
    \begin{subfigure}{0.3\textwidth}
        \centering
        \includegraphics[width=\linewidth]{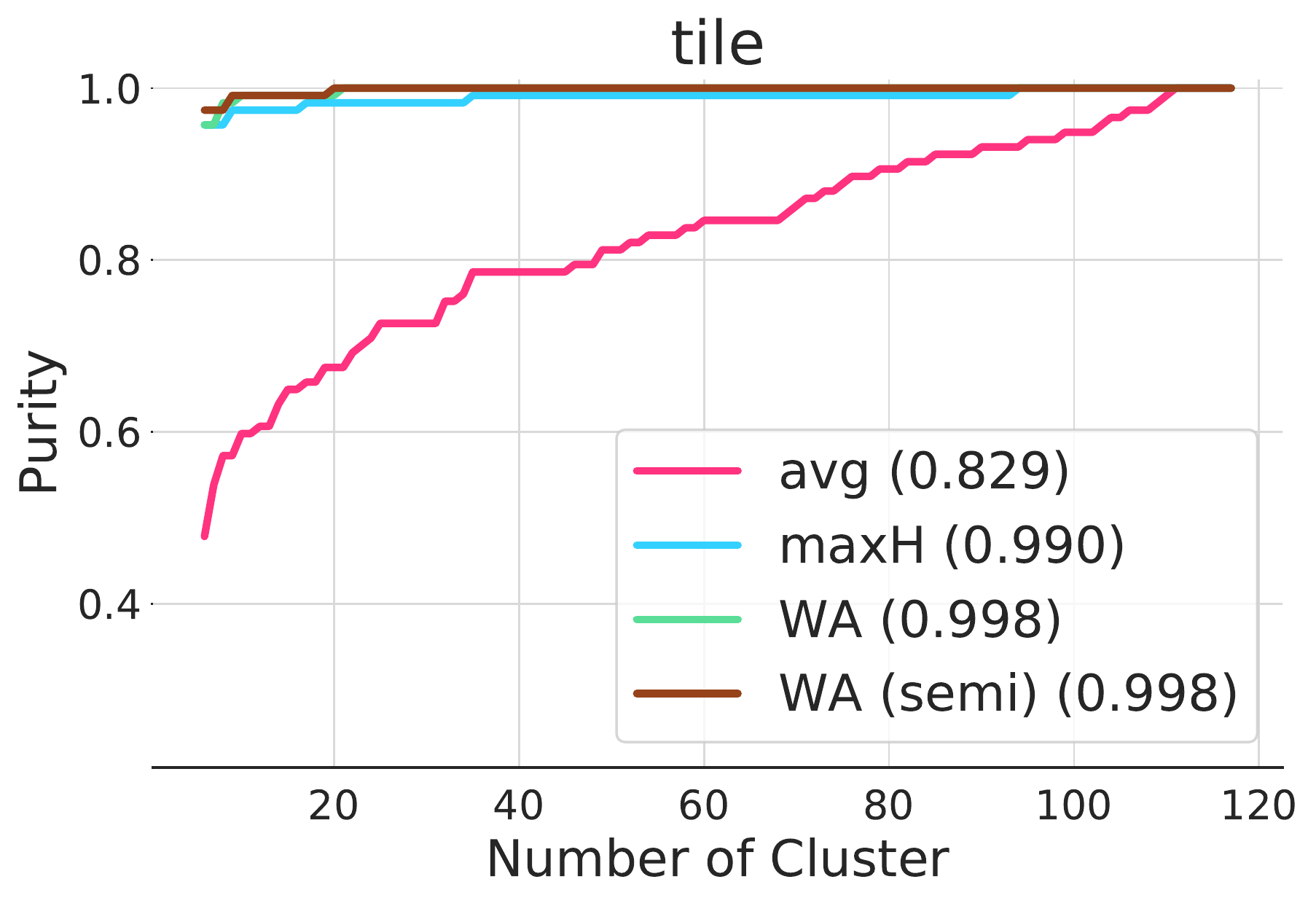}
    \end{subfigure}
    \begin{subfigure}{0.3\textwidth}
        \centering
        \includegraphics[width=\linewidth]{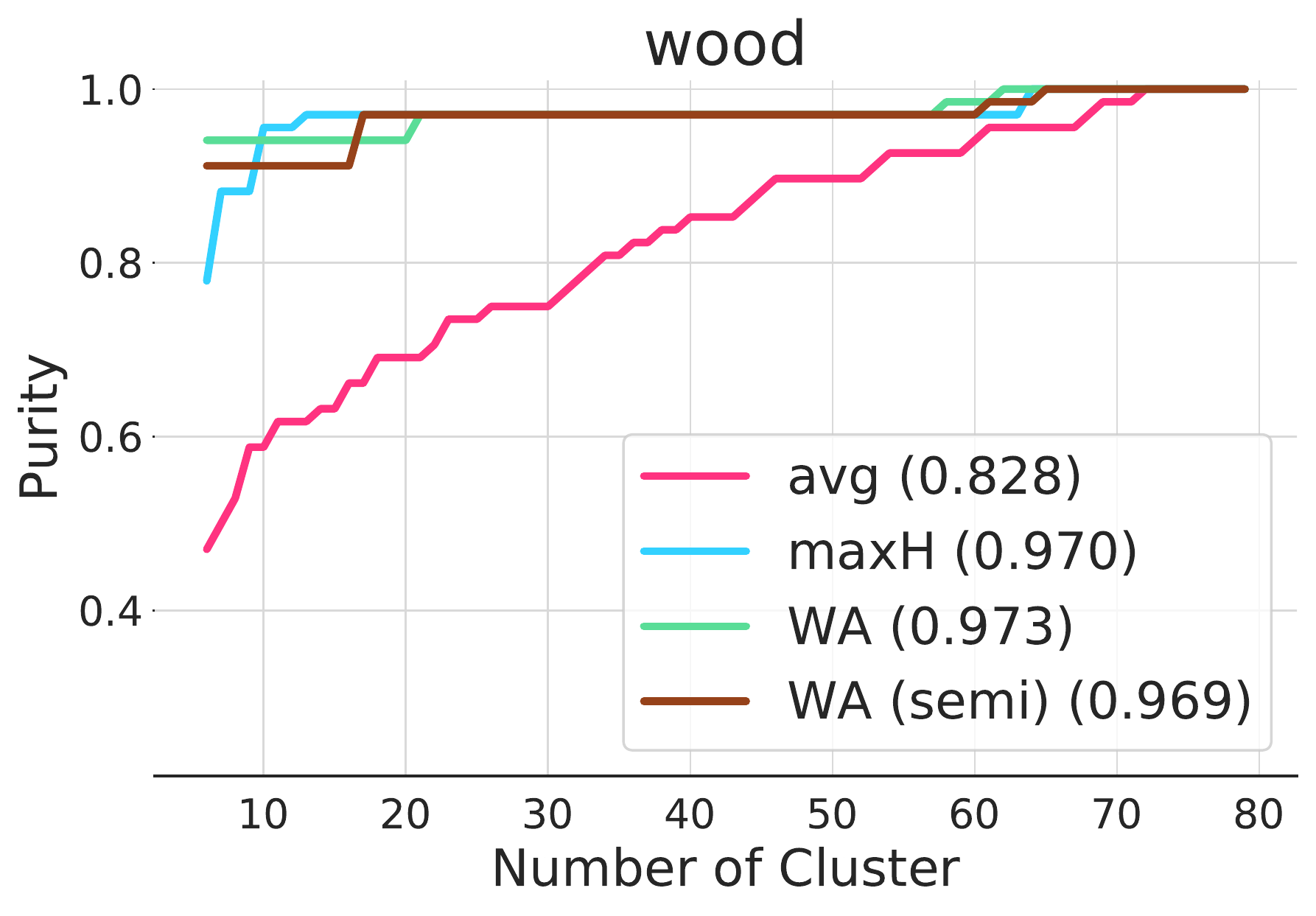}
    \end{subfigure}
    \caption{Purity of clusters with different number of clusters on MVTec dataset. Hierarchical Ward clustering is used for clustering method with different attention strategies including uniform, top-$k$, and soft. Numbers in the bracket represent the area under the curve divided by the total number of examples.}
    \label{fig:overclustering_auc}
\end{figure}